\documentclass[12pt,a4paper]{article}

\usepackage[utf8]{inputenc}
\usepackage[T1]{fontenc}

\usepackage[a4paper,left=2.2cm,right=2.2cm,top=2.2cm,bottom=2.6cm]{geometry}
\DeclareUnicodeCharacter{2006}{\,}% six-per-em space
\DeclareUnicodeCharacter{2009}{\,}% thin space
\DeclareUnicodeCharacter{200B}{} % zer-width space


\usepackage[numbers,sort&compress]{natbib} % \citep, \citet, styl numeryczny

\usepackage{microtype}
\usepackage{amsmath,amssymb,mathtools}
\usepackage{bm}
\usepackage{physics} % używasz makr z tego pakietu

\usepackage{graphicx}
\usepackage{subcaption}
\usepackage{booktabs}
\usepackage{multirow}
\usepackage{array}
\usepackage{tabularx}
\usepackage[table]{xcolor}

\usepackage{algorithm}
\usepackage[noend]{algpseudocode}

\usepackage{tikz}
\usetikzlibrary{calc,matrix,fit,shapes.geometric,shapes.misc,arrows.meta,%
                positioning,backgrounds,plotmarks,decorations.pathmorphing,%
                decorations.pathreplacing}
\usepackage{pgfplots}
\pgfplotsset{compat=1.13}

\usepackage{siunitx}
\usepackage{acronym}

\usepackage[hidelinks]{hyperref}
\usepackage[nameinlink,noabbrev]{cleveref}

\usepackage{enumitem}
\usepackage{placeins}

\newcommand{\sep}{;\ } % separator używany wewnątrz środowiska keyword
\newenvironment{keyword}{%
  \par\noindent\textbf{Keywords: }\ignorespaces}{\par}

\newcommand{\R}{\mathbb{R}}
\newcommand{\C}{\mathbb{C}}

\newcommand{\Supp}{\operatorname{supp}}
\newcommand{\hot}{\mathrm{h.o.t.}}

\title{Newton-Puiseux Analysis for Interpretability and Calibration of Complex-Valued Neural Networks} 

\author{Piotr Migus \\ \texttt{migus.piotr@gmail.com}}
\date{}

\begin{document}

\maketitle

\begin{abstract}
Complex-valued neural networks (CVNNs) are particularly suitable for handling phase-sensitive signals, including electrocardiography (ECG), radar/sonar, and wireless in-phase/quadrature (I/Q) streams. Nevertheless, their \emph{interpretability} and \emph{probability calibration} remain insufficiently investigated. In this work, we present a Newton--Puiseux framework that examines the \emph{local decision geometry} of a trained CVNN by (i) fitting a small, kink-aware polynomial surrogate to the \emph{logit difference} in the vicinity of uncertain inputs, and (ii) factorizing this surrogate using Newton--Puiseux expansions to derive analytic branch descriptors, including exponents, multiplicities, and orientations. These descriptors provide phase-aligned directions that induce class flips in the original network and allow for a straightforward, \emph{multiplicity-guided} temperature adjustment for improved calibration. We outline assumptions and diagnostic measures under which the surrogate proves informative and characterize potential failure modes arising from piecewise-holomorphic activations (e.g., modReLU). Our phase-aware analysis identifies sensitive directions and enhances Expected Calibration Error in two case studies beyond a controlled $\C^2$ synthetic benchmark---namely, the MIT--BIH arrhythmia (ECG) dataset and RadioML 2016.10a (wireless modulation)---when compared to uncalibrated softmax and standard post-hoc baselines. We also present confidence intervals, non-parametric tests, and quantify sensitivity to inaccuracies in estimating branch multiplicity. Crucially, this method requires no modifications to the architecture and applies to any CVNN with complex logits transformed to real moduli.
\end{abstract}


%========================================================
\section{Introduction}\label{sec:intro}
%========================================================

Deep learning serves as the foundation for contemporary pattern recognition across various domains, including vision, language, and biomedical signals \citep{lecun2015deep}. While the majority of architectures are \emph{real-valued}, there is a growing argument that \emph{complex-valued neural networks} (CVNN) present a more fitting inductive bias when \emph{phase} is of significance (e.g., radar/sonar, \emph{Short-Time Fourier Transform} (STFT)-based audio, \emph{magnetic resonance imaging} (MRI), \emph{electroencephalography} (EEG)/\emph{electrocardiography} (ECG), coherent optics, wireless, and quantum machine learning). By modeling amplitude and phase, CVNNs honor rotational symmetries and utilize Wirtinger calculus, enabling gradient-based optimization with advantageous landscapes \citep{trabelsi2017deep, barrachina2023theory, abdalla2023complex, basterretxea2021survey}.

\textbf{Practical challenge: Interpretability and calibration}
Despite advancements in architecture (e.g., complex batch normalization, modReLU/zReLU) \citep{trabelsi2017deep}, a gap continues to exist between the enhancement of \emph{predictive} performance and our capability to \emph{interpret} and \emph{calibrate} CVNNs \citep{abdalla2023complex}. Post-hoc explanation methods---such as saliency, \emph{Local Interpretable Model-agnostic Explanations} (LIME), \emph{SHapley Additive exPlanations} (SHAP), and \emph{Layer-wise Relevance Propagation} (LRP)---are primarily designed for real-valued activations \citep{montavon2018methods,samek2017explainable}. Variants aimed at the complex domain often reduce to $\R^n$ or focus solely on visualizing magnitudes, implicitly assuming simple, single-valued decision boundaries. Similarly, most calibration strategies (including temperature/vector scaling, isotonic regression, Dirichlet/Beta distributions) are adapted from those used for real-valued networks \citep{naeini2015bayesian}, overlooking the fact that \emph{multi-branched (multi-sheeted)} decision functions in $\C$ can lead to abrupt confidence discontinuities with minimal phase alterations.

\textbf{An algebraic-geometry lens}
Branching behavior is intrinsic to algebraic curves.  

According to the Newton--Puiseux theorem, in the vicinity of an isolated singularity, \emph{an} algebraic curve can be decomposed into a finite number of fractional-power series, with initial exponents and directions captured by a Newton polygon \citep{wall2004singular,brieskorn1986plane,walker1950algebraic}. Related concepts provide insights into singular learning theory and piecewise-linear decision regions \citep{watanabe2009book,Maragos_2021,suzuki2023waic}, yet these ideas have not been employed to elucidate learned \emph{complex}-valued classifiers. 

\textbf{One-sentence summary.}
We present a Newton--Puiseux framework that (i) fits a \emph{local} polynomial surrogate to the \emph{logit difference} near high-uncertainty inputs and (ii) decomposes this surrogate into Puiseux series, resulting in analytic descriptors of branching, curvature, and phase alignment that facilitate robustness analysis and calibration.

\paragraph{Why Newton--Puiseux beyond attribution}
Attributions (saliency/LIME/SHAP) provide descriptive insights but \emph{do not} offer an actionable \emph{direction} that can be proven to cross the decision boundary, nor do they assess how quickly higher-order terms will dominate.
Our Newton--Puiseux analysis addresses this gap:
 (i) a local polynomial approximation of the \emph{logit difference} presents analytic branch descriptors (exponents, multiplicities, orientations); (ii) these descriptors generate \emph{phase-aligned rays} that can change the class on the original network within a small radius around most uncertain anchors; and (iii) the same multiplicity signal informs a phase-aware temperature that minimizes \emph{Expected Calibration Error} (ECE).
We ensure that the surrogate remains compact and well-conditioned (with degree~$d{\le}4$, kink-aware filtering, and column scaling) and consistently validate the Newton--Puiseux predictions \emph{on the original network} through a radial sweep; potential failure modes and safeguards are detailed in \ref{app:impl}.

\textbf{Contributions.}
\begin{enumerate}
\item \textbf{Local surrogate of decision geometry.}
We create a robust, kink-aware bivariate polynomial surrogate of the logit difference $f:\C^{2}\!\to\!\R$ around an uncertain input and factor it to extract Puiseux branches (exponents, multiplicities, orientations). The fitting process specifically manages neighborhoods with modReLU/zReLU through outlier rejection and distance weighting.

\item \textbf{Phase-aware robustness descriptors.}
The coefficients from the Puiseux series provide closed-form, phase-aligned curvature indicators that forecast adversarial flip radii and pinpoint fragile directions in $\C^{2}$---information that is not accessible through gradients, LIME, or SHAP.

\item \textbf{Branch-guided calibration.}
We introduce phase-aware temperature scaling where the estimated branch multiplicity influences the effective temperature. This approach integrates with standard post-hoc calibration methods and enhances ECE compared to the uncalibrated softmax. We present 95\% confidence intervals and utilize Wilcoxon signed-rank tests to examine robustness to potential multiplicity mis-estimation (refer to the comparisons and ablations sections).

\item \textbf{Generality across domains.}
In addition to a controlled synthetic $\C^{2}$ benchmark, we assess performance on MIT--BIH arrhythmia (ECG) and RadioML~2016.10a (RadioML) \citep{o2016convolutional} for modulation recognition. The same $\C^{2}$ pipeline is applicable without any architectural modifications, suggesting broader transferability beyond biomedicine.

\item \textbf{Design guidance and ablations.}
We provide: (a) an analysis of the sensitivity of uncertainty mining to $(\tau, \delta)$ (anchor counts, dispersion, explanatory value), (b) a resource benchmark comparing the per-anchor wall-clock and memory usage of Puiseux analysis against gradient saliency, and (c) insights into how calibration is sensitive to fractional mis-estimation of branch multiplicity.
\end{enumerate}

\textbf{Scope and assumptions.}
Our analysis specifically targets \emph{local} decision geometry in $\C^{2}$. We do not assume global holomorphicity; instead, we quantify how frequently uncertain anchors fall into nonholomorphic regions caused by piecewise activations, reporting the effects on surrogate fitting and bias.

\textbf{Roadmap.}
Section~\ref{sec:related} reviews relevant literature.
Section~\ref{sec:notation} establishes notation and calibration preliminaries.
Section~\ref{sec:method} describes the surrogate and the Newton--Puiseux solver in detail. 
Section~\ref{sec:theory} outlines the assumptions, bounds, and failure modes.
Section~\ref{sec:setup} discusses datasets, models, calibration baselines, and anchor selection.
Results for MIT--BIH are covered in Section~\ref{sec:results-mit}, with ablation studies in Section~\ref{sec:ablations-mit} and direct comparisons (including statistical tests) in Section~\ref{sec:comparisons-mit}.
Findings and analyses for RadioML are presented in Sections~\ref{sec:results-radio} through \ref{sec:comparisons}.
Section~\ref{sec:discussion} addresses generalization and limitations, while Section~\ref{sec:conclusion} provides a summary.
A synthetic sanity check is located in \ref{app:synthetic_sanity}; additional algorithmic, implementation, computational, sensitivity, calibration reporting, and dataset/model specifics can be found in \ref{app:newton} through \ref{app:fragileRadio}.

\paragraph{Reproducibility}
All source code, pre-trained weights, and reproduction scripts are publicly accessible at \url{https://github.com/piotrmgs/puiseux-cvnn}.
%========================================================
\section{Related Work}\label{sec:related}
%========================================================

We examine four key areas: (i) architectures and training methodologies for CVNNs, (ii) interpretability approaches for complex-domain models, (iii) methods for probability calibration, and (iv) algebraic--geometric analyses of learning systems. We emphasize recent advancements and highlight the gaps that our method aims to fill.

\subsection{Architectures and training for CVNNs}
Early studies established the theoretical foundation for holomorphic activations and Wirtinger--gradient learning in $\C$ \citep{hirose2006complex,nitta2004orthogonality,kreutz2009complex,abdalla2023complex,barrachina2023theory}. The development of practical deep training was facilitated by activation functions like \emph{zReLU} and \emph{modReLU}, which address the issue of vanishing gradients while maintaining phase information \citep{guberman2016complex}. The resurgence of deep learning led to the introduction of systematic complex architectures, including Complex CNNs that utilize appropriate normalization and initialization techniques \citep{trabelsi2017deep} as well as unitary/complex recurrent models \citep{arjovsky2016unitary}.

Recently, the scope and sophistication of CVNNs have expanded further: function classes such as \emph{complex-valued Kolmogorov--Arnold networks} (CVKAN) extend structured models into $\C$ \citep{cvkan2025}. Concurrently, real-domain Kolmogorov--Arnold Networks (KAN) and their hierarchical, backpropagation-free variant (HKAN) prioritize intrinsic interpretability through learnable univariate mappings and layer-wise least-squares training. Although these approaches are orthogonal to our post-hoc, model-agnostic analysis, they are complementary in nature \citep{liu2024kan,dudek2025hkan}. Complex-valued training is also being utilized in distributed or federated settings for phase-sensitive signals \citep{cvnn_federated_2024}, while complex architectures increasingly contribute to end-to-end perception tasks \citep{cvyolo_2025}. These trends are supported by domain-specific applications reported in areas such as audio (e.g., STFT-based separation and acoustic scene analysis), MRI, and wireless/sonar/radar technologies, as well as quantum-state modeling \citep{basterretxea2021survey,zhang2018complex}. Beyond the fields of vision and biomedicine, wave-based non-destructive evaluation also benefits from phase-aware learning techniques, such as ultrasonic characterization \citep{xu2023ultrasonic}. Additionally, complex-valued pipelines have been explored for Magnetic Particle Imaging (MPI) system calibration, extending CVNN utility to medical imaging \citep{Li_2025}.

\noindent\textit{Gap.} Despite these advancements, there remains a paucity of research that leverages the intrinsic $\C$-geometry for explanation: many visualizations simplify activations to magnitudes, thereby implicitly assuming single-valued decision surfaces. While recent works on CVNNs have primarily focused on evaluating predictive metrics, few have investigated the \emph{local decision geometry} or established a connection between phase-branching phenomena and uncertainty and calibration---precisely the interface that our approach addresses.

\subsection{Interpretability for complex-domain networks}
Mainstream explainable Artificial Intelligence (XAI) methods (such as saliency, LRP, LIME, and SHAP \citep{ribeiro2016should,lundberg2017unified}) have been designed for real activations, often reducing complex activations to magnitudes or real--imaginary components \citep{montavon2018methods,samek2017explainable}. For complex-domain vision tasks, saliency modeling has also been studied with deep complex networks in transformed domains \citep{jiang2019image}. In safety--critical contexts, the essential criteria extend beyond merely achieving accuracy to include \emph{interpretability} and \emph{probability calibration} \citep{arrieta2020explainable,guo2017calibration}. A recent advancement towards \emph{complex-native} explanations is \textsc{DeepCShap}, which adapts SHAP for complex-valued features and networks \citep{deepcshap_2024}. In the realm of phase-aware signal processing (such as in speech enhancement), efforts have been made to partially reclaim lost phase information \citep{choi2018phase}. Nevertheless, existing complex-domain XAI and calibration techniques typically inherit assumptions rooted in real values---examples include temperature scaling, isotonic regression, and Bayesian binning \citep{naeini2015bayesian}---which treat each class score as if it were single-valued. However, these methods still regard each score as single-valued and fail to reveal the \emph{branching} of decision surfaces in $\C^2$.

\noindent\textit{Gap.} We propose to provide analytic, phase-aware descriptors through Newton--Puiseux factorization of a local surrogate model, effectively capturing the multi-sheeted behavior that gradient-based techniques, LIME, and SHAP do not account for.

\subsection{ Confidence calibration (notes for CVNNs)}
The topic of calibration has garnered renewed focus, highlighted by recent surveys and extensive benchmarks of post-hoc methods (such as temperature/vector scaling, Dirichlet/Beta calibration, isotonic/spline families) \citep{calibration_survey_2023,benchmark_calib_2023,manyclass_calibration_2024, guo2017calibration,kull2019dirichlet}. While these developments are mostly architecture-agnostic, there is limited understanding of how \emph{complex-phase geometry} (e.g., branch multiplicity) influences calibration in CVNNs. Additionally, we evaluate the robustness of these calibration improvements with respect to unitary re-parameterizations of $(\Re,\Im)$ (including global phase rotations and channel permutations) \citep{minderer2021revisiting}; the findings are presented in Sections~\ref{sec:ablations-mit} and \ref{sec:rml-ablations}. Orthogonally to post-hoc calibration, predictive uncertainty can also be improved via deep ensembles \citep{lakshminarayanan2017simple}, while coverage guarantees can be provided by conformal prediction frameworks \citep{messoudi2022conformal}.

\noindent\textit{Gap.} We correlate Puiseux branch multiplicity with confidence distortion and utilize it to inform the process of phase-aware temperature scaling; we present confidence intervals and non-parametric tests to quantify these effects.

\subsection{Algebraic/tropical analyses and singular learning theory}
There is an ongoing discussion regarding the application of algebraic and tropical geometry to analyze decision regions and piecewise-linear abstractions \citep{maclagan2015tropical,montufar2014number}, which includes recent tutorials on machine learning and optimization \citep{tropical_tutorial_icassp_2024,Maragos_2021}. A broader strand connects algebraic geometry and modern ML more directly, from expository bridges to concrete applications \citep{douglas2021mumford,bao2022mlag,coates2023terminal}. Concurrently, singular learning theory is advancing PAC--Bayesian and asymptotic tools designed for models characterized by singular parameterizations \citep{suzuki2023waic, slt_upper_bound_2024, degeneracy_mechinterp_iclr_2024}. Our contribution is not simply a novel Newton--Puiseux theorem; it marks the first application of this theorem to derive local analytic descriptors of intricate decision geometry, which directly relate to robustness radii and phase-aware calibration---an aspect not addressed by tropical analyses or singular learning theory. 

\ \noindent\textit{Positioning.} To the best of our knowledge, the Newton--Puiseux decompositions have yet to be utilized for deriving \emph{local, analytic} descriptors of complex decision geometry in learned classifiers. Our framework enhances tropical and singular analyses through (i) the fitting of a locally polynomial surrogate that accounts for kinks, and (ii) the extraction of Puiseux branches that provide insights into robustness directions and calibration.

%========================================================
\section{Notation and Problem Setup}\label{sec:notation}
%========================================================

\paragraph{Complex domain and input layout}
Write $z=(z_1,\ldots,z_D)\in\C^{D}$ with $z_j = x_j + i\,y_j$, $x_j,y_j\in\R$.
We can equivalently represent $\C^{D}\cong\R^{2D}$ following the block convention
\[
  [\Re_1,\ldots,\Re_D \mid \Im_1,\ldots,\Im_D]\in\R^{2D},
\]
where each pair $(\Re_j,\Im_j)$ corresponds to a single complex coordinate $z_j$. In all experiments, we set $D=2$, meaning $z=(z_1,z_2)\in\C^2$.

\paragraph{Complex-valued neural network (CVNN)}
Let $g:\R^{2D}\to\R^{2K}$ represent a CVNN, where outputs consist of real and imaginary components concatenated as $g(z)=(u_1,v_1,\dots,u_K,v_K)$, resulting in complex scores $c_k=u_k+i\,v_k$. The real \emph{logits} are defined as the magnitudes of these complex scores,
\begin{equation}\label{eq:modlogits}
  \ell_k(z)=|c_k(z)|=\sqrt{u_k(z)^2+v_k(z)^2+\varepsilon},
\end{equation}
with a numerical stabilizer $\varepsilon>0$ (we utilize $\varepsilon=10^{-9}$).
The class probabilities are determined using a softmax function with an optional temperature $T>0$:
\begin{equation}\label{eq:softmax}
  p_k(z;T)=\frac{\exp(\ell_k(z)/T)}{\sum_{j=1}^{K}\exp(\ell_j(z)/T)},
  \qquad \sum_{k=1}^{K}p_k=1.
\end{equation}
We exclusively use the symbol $T$ for the softmax temperature throughout this document.

\paragraph{Binary margin and decision function}
For the case of $K=2$, we define the absolute \emph{margin} as
\begin{equation}\label{eq:margin}
  m(z)=\bigl|p_1(z;T)-p_2(z;T)\bigr|.
\end{equation}
In our subsequent analysis, we \emph{reserve}
\begin{equation}\label{eq:logitdiff}
  f(z) = \ell_1(z)-\ell_2(z), \qquad f:\C^{D}\to\R,
\end{equation}
for the \emph{logit difference}. 
Since $p_1(z;T)$ is a monotonic function of $f(z)/T$, the zero level set $f(z)=0$ coincides with the binary decision boundary defined as
\begin{equation}\label{eq:boundary}
  \mathcal{B} = \bigl\{z\in\C^D : f(z)=0\bigr\}.
\end{equation}
For scenarios where $K>2$, we utilize the top‑2 variant $f(z)=\ell_{(1)}(z)-\ell_{(2)}(z)$; while our experiments focus on binary classification, we include this extension for thoroughness.

\paragraph{Expected calibration error}
Given $B$ confidence bins denoted as $B_1,\ldots,B_B$ with respective sizes $|B_b|$, the \emph{expected calibration error} is calculated as
\[
  \mathrm{ECE} = \sum_{b=1}^{B}\frac{|B_b|}{n}
  \bigl|\mathrm{acc}(B_b)-\mathrm{conf}(B_b)\bigr|,
\]
where $\mathrm{acc}$ and $\mathrm{conf}$ represent the empirical accuracy and average confidence in bin $b$. Our probability and loss notation follows standard pattern-recognition conventions \citep{bishop2006pattern}.

\paragraph{Uncertain inputs (anchor selection)}
We mine \emph{anchors} from the union of two standard uncertainty tests:
\begin{align}
  \mathcal{U}_{\mathrm{prob}}
    &= \bigl\{z : \max_k p_k(z;T) < \tau \bigr\}, & \tau\in(0,1), \label{eq:uncertain_prob}\\
  \mathcal{U}_{\mathrm{margin}}
    &= \bigl\{z : m(z) < \delta \bigr\},            & \delta\in(0,1). \label{eq:uncertain_margin}
\end{align}
We set $\mathcal{U}_\star := \mathcal{U}_{\mathrm{prob}}\cup\mathcal{U}_{\mathrm{margin}}$.
Candidate anchors $z^\star \in \mathcal{U}_\star$ are selected for local analysis and postprocessing. When a post-hoc calibrator is developed using validation data (MIT--BIH: temperature or isotonic; RadioML: Platt or isotonic), the \emph{calibrated} validation probabilities guide the selection of $(\tau,\delta)$, and the same calibrator is subsequently applied unchanged to the test set to ensure fair selection and evaluation (refer to \ref{app:impl}).

\paragraph{Local neighborhood and surrogate coordinates}
Fix an anchor $z^\star=(z_1^\star,z_2^\star)\in\C^2$ and
\[
  B_\Delta(z^\star):=\{z\in\C^2:\|z-z^\star\|_\infty \le \Delta\}\cong[-\Delta,\Delta]^4\subset\R^4.
\]
We recenter the variables and denote $(\xi,\eta)=(z_1-z_1^\star,\; z_2-z_2^\star)\in\C^2$ (resulting in four real coordinates in~$\R^4$).

\paragraph{Local surrogate (overview)}
We approximate $f$ in $B_\Delta(z^\star)$ using a degree‑$d$ bivariate polynomial given by  
\begin{equation}\label{eq:poly}
  \widehat f_d(\xi,\eta)=\sum_{i+j\le d} c_{ij}\,\xi^i\eta^j,\qquad \widehat f_d(0,0)=0.
\end{equation}

This polynomial is fitted via weighted least squares on $N$ random perturbations within the range
$[-\Delta,\Delta]^4$.
To highlight the non‑linear structure that is significant for branching, we exclude constant and linear terms (setting all monomials with $i{+}j<2$ to zero). To enhance robustness near piecewise‑holomorphic kinks (such as modReLU/zReLU), we implement (i) a down‑weighting of samples based on their distance and (ii) the exclusion of a small band around detected kinks (which we refer to as “kink‑aware” filtering). Unless specified otherwise, our defaults include $d=4$, $\Delta\approx 0.05$ (feature scale), $N\approx 600$, and a kink threshold of $10^{-6}$; comprehensive ablations can be found in Sections~\ref{sec:ablations-mit} and \ref{sec:rml-ablations}.

\paragraph{Branch (singular) points and Wirtinger notation}
Utilizing the Wirtinger operators
\(
 \partial_{z}=\tfrac12(\partial_{x}-i\,\partial_{y}),\;
 \partial_{\bar z}=\tfrac12(\partial_{x}+i\,\partial_{y})
\)
component-wise for each complex coordinate, we define a point $z_\star\in\mathcal{B}$ as \emph{singular} if $f(z_\star)=0$ and both $\partial_z f(z_\star)=0$ and $\partial_{\bar z} f(z_\star)=0$ hold true (noting that our function $f$ is real‑valued as a result of moduli) \citep{brandwood1983complex,kreutz2009complex}. In the vicinity of such points, the zero-set of $\widehat f_d$ may split into multiple Puiseux branches; this scenario is addressed in the Newton--Puiseux stage (see Section~\ref{sec:method}).  

%========================================================
\section{Method}
\label{sec:method}
%========================================================

We investigate the \emph{local} decision geometry of the logit difference  
\(f(z)=\ell_1(z)-\ell_2(z)\) (refer to \eqref{eq:logitdiff}) in \(\C^2\) around an uncertain input \(z^\star\). Our methodology comprises two interconnected stages: (i) the creation of a robust polynomial surrogate 
\(\widehat f_d\) fitted within a small neighborhood \(B_\Delta(z^\star)\), and (ii) a Newton--Puiseux factorization of \(\widehat f_d=0\) that reveals branch exponents, multiplicities, and orientations. From this analysis, we derive (a) phase-aware robustness directions and (b) a multiplicity-guided temperature scaling rule for calibration.  

%--------------------------------------------------------
\subsection{Local polynomial surrogate in \texorpdfstring{$\C^2$}{C2}}
\label{sec:method:surrogate}
%--------------------------------------------------------

With a determined anchor \(z^\star=(x^\star,y^\star)\in\C^2\), we transition to local coordinates  
\((\xi,\eta)=(x-x^\star,y-y^\star)\) and estimate the logit difference using a low-degree bivariate polynomial expressed as
\[
  \widehat f_d(\xi,\eta)=\sum_{i+j\le d} c_{ij}\,\xi^i\eta^j,\qquad \widehat f_d(0,0)=0.
\]
(see also \eqref{eq:poly} for the overview definition). 
To better capture the non‑linear geometry pertinent to branching, we exclude constant and linear terms (removing all monomials with \(i{+}j<2\)), fit \(\widehat f_d\) on random perturbations originating from a compact hypercube around the anchor, and reduce emphasis on samples that are distant or located near piecewise‑holomorphic kinks (e.g., modReLU/zReLU sectors). Fidelity on held-out perturbations is assessed via metrics such as RMSE, MAE, Pearson~\(\rho\), and sign agreement (SA); low fidelity prompts a reduction in the neighborhood size and/or a decrease in the polynomial degree.

Unless otherwise indicated, we typically employ small neighborhoods and low polynomial degrees (usually \(d{=}4\)) to ensure a well-conditioned design matrix while adequately capturing curvature and branching characteristics. \emph{As outlined in our released code}, the specific processes for sampling, weighting, kink management, and numerical safeguards (including scaling, ridge, and rank checks) are detailed in \ref{app:impl} (“Implementation details and controls”).

%--------------------------------------------------------
\subsection{Newton--Puiseux factorization and descriptors}
\label{sec:method:np}
%--------------------------------------------------------

We apply a Newton--Puiseux decomposition to factor the surrogate equation \(\widehat f_d(\xi,\eta)=0\). Each compact edge of the Newton polygon defines quasi‑homogeneous factors, the univariate reductions of which produce leading terms of the form \(a\,\xi^{\theta}\) or \(a\,\eta^{\theta}\). From these terms, we extract: (i) the count of branches and their \emph{multiplicities}; (ii) rational exponents \(\theta>0\); and (iii) phase orientations, represented by the arguments of the leading complex coefficients. These descriptors serve as the foundation for the subsequent robustness and calibration steps. Detailed algorithmic processes and worked examples are provided in \ref{app:newton}; implementation choices and computational details are summarized in \ref{app:compute}.

%--------------------------------------------------------
\subsection{ Phase-aware temperature scaling guided by multiplicity }
\label{sec:method:temp}
%--------------------------------------------------------

Let \(m\) represent the estimated branch multiplicity at an anchor, which is the sum of the multiplicities of branches that locally intersect the decision sheet. From the surrogate model, we derive a \emph{phase-aware} temperature as follows:
\begin{equation}
  T’(m) \;=\; T_{\text{base}}\;\cdot\; m^{-\gamma},\qquad
  \gamma\in[0,1],\ \ \text{default } \gamma=\tfrac12,
  \label{eq:temp-map}
\end{equation}
Here, \(T_{\text{base}}\) refers to the global temperature (or \(1\) if it is absent). The intuition behind this is that a larger \(m\) signifies sharper confidence discontinuities; thus, a lower effective \(T’\) adjusts the logits to soften these discontinuities. In Sections~\ref{sec:comparisons-mit} and \ref{sec:comparisons}, we present calibration metrics such as expected calibration error (ECE), negative log likelihood (NLL), Brier score, and confidence intervals. To examine robustness against surrogate noise, we explore relative mis-estimation as \(m_{\text{est}} = (1+\varepsilon)\,m_{\text{true}}\) and apply the resulting multiplier \(T_{\text{mult}}=(m_{\text{true}}/m_{\text{est}})^\gamma=(1+\varepsilon)^{-\gamma}\) (with the default \(\gamma=\tfrac12\)); the results of this exploration are included in our ablations.

\paragraph{Rationale for multiplicity as an indicator of calibration}
An increased branch multiplicity results in a steeper local profile of the logit difference surface. By decreasing the effective temperature according to $m^{-\gamma}$, we counterbalance this steepness, thereby softening the class likelihoods precisely in regions where multi-sheeted behavior would typically lead to inflated confidence levels. We quantify this effect in our fixed-encoder MIT--BIH study, detailed in Section~\ref{sec:comparisons-mit}.

%--------------------------------------------------------
\subsection{ Transitioning from descriptors to robustness directions}
\label{sec:method:robust}
%--------------------------------------------------------

The principal Puiseux terms delineate phase-aligned directions where the sign of \(\widehat f_d\) changes most rapidly. We identify candidate directions by examining the phase indicated by the dominant coefficients, then validate these directions on the original network by traversing along rays from the anchor and documenting the smallest radius that leads to a change in the predicted class. This generates robustness curves and flip radii per anchor, which can be compared with gradient-based benchmarks. The exact probing protocol, along with default radii and step counts, is specified in \ref{app:impl}.

%--------------------------------------------------------
\subsection{Practical control parameters (design choices)}
\label{sec:controls}
%--------------------------------------------------------

To achieve a balance between conditioning and geometric fidelity, we maintain small neighborhoods, low degrees, and moderate sample counts (typical values include \(d{=}4\), \(\Delta\!\approx\!0.05\), and hundreds of perturbations). Kink-aware filtering is employed to prevent the mixing of holomorphic sheets near modReLU/zReLU transitions, while light ridge regularization is utilized to stabilize the fitting process when necessary. A comprehensive list of default parameters \emph{as implemented in the released code}, along with guidance for tuning, can be found in \ref{app:impl}, Table~\ref{tab:defaults}.

%========================================================
\section{Theoretical Insights and Potential Failure Modes}
\label{sec:theory}
%========================================================

We formalize the conditions under which a local polynomial surrogate provides meaningful information, how Newton--Puiseux descriptors constrain curvature and the emergence of higher-order dominance, and how piecewise-holomorphic activations lead to identifiable failure modes. Throughout this section, we denote $f(z)=\ell_1(z)-\ell_2(z)$ as the logit difference (refer to Section~\ref{sec:notation}). Anchors are derived using the union rule as described in Section~\ref{sec:uncertainty}; the surrogate$\widehat f_d$ is fitted with constant and linear terms excluded (all monomials where $i{+}j<2$ are fixed to zero) using weighted least squares, while applying a small Tikhonov regularization term \citep{golub2013matrix,hansen1998rank,demmel1997applied}. The diagnostics and numerical safeguards align with \ref{app:impl} and the released code.

%--------------------------------------------------------
\subsection{Conditions for Informative Local Surrogates}
\label{sec:theory:assumptions}
%--------------------------------------------------------
We operate in $\C^2\!\cong\!\R^4$ using local coordinates $(\xi,\eta)$ centered around an anchor $z^\star$; we denote the Euclidean norm on $\R^4$ as $\|\cdot\|$ and define$\mathcal{Q}_\Delta=\{(\xi,\eta):\|(\xi,\eta)\|_\infty\le\Delta\}$.

\paragraph{Local regularity (away from kinks)}
We assume that within $\mathcal{Q}_\Delta$ outside of a narrow \emph{kink band}$\mathcal{K}_\varepsilon$ (which is defined below), the logit difference $f$
satisfies two conditions: (i) it is locally Lipschitz with a constant $L$, and (ii) it is twice differentiable with a remainder of the third order that is bounded by
$\kappa\|(\xi,\eta)\|^3$.
These assumptions reflect the piecewise-holomorphic nature of typical CVNN activations (modReLU/zReLU) while excluding a vanishing neighborhood surrounding sector boundaries.

\paragraph{Kink band and sample selection}
Let $a_h$ represent the complex pre-activation of the $h$-th hidden unit within the first complex layer, while $b_h$ denotes its modReLU bias. We define the per-unit margin as $s_h=|a_h|+b_h$ and introduce the \emph{ANY-unit} band $\mathcal{K}_\varepsilon=\{(\xi,\eta):\min_h s_h\le\varepsilon\}$. In our implementation (\ref{app:impl}), we exclude samples from $\mathcal{K}_\varepsilon$ (with a default setting of \texttt{exclude\_kink\_eps}$=10^{-6}$) and weight the retained samples in proportion to $\max(\min_h s_h,0)+10^{-9}$, optionally applying distance decay to reduce the emphasis on points situated near sector boundaries.

\paragraph{A practical $L^2$ error bound}
For a degree $d\in\{3,4\}$, using a column-normalised design along with a weighted ridge (with a default $\lambda{=}10^{-8}$), the fit error satisfies the following inequality:
\[
  \|f-\widehat f_d\|_{L^2(\mathcal{Q}_\Delta\setminus\mathcal{K}_\varepsilon)}
  \;\le\;
  C_1\,\kappa\,\Delta^3 \;+\; C_2\,\sigma\,\sqrt{\tfrac{M}{N_{\mathrm{kept}}}},
\]
where $M$ indicates the number of active monomials (with $i{+}j\!\ge\!2$), $N_{\mathrm{kept}}$ represents the count of perturbations remaining after filtering, $\sigma$ signifies an effective noise scale, and constants $C_{1,2}$ solely depend on the basis and weights. Thus, smaller values of $\Delta$ and a moderate $M$ (indicating lower $d$) lead to improved fidelity, motivating our preference for default $d{=}4$ and compact neighborhoods.

\paragraph{Technical note: approximation error near activation boundaries (informal)}
Let $f(z)=\ell_1(z)-\ell_2(z)$ denote the (real) logit difference and let
$\widehat f_d$ be the degree-$d$ local polynomial surrogate fitted in a cube
$\mathcal{Q}_\Delta=\{(\xi,\eta)\in\R^{4}:\|(\xi,\eta)\|_\infty\le \Delta\}$
around an anchor. Assume that, \emph{away from a thin activation-band}
$\mathcal{K}_\varepsilon$ (e.g., modReLU/zReLU sector boundaries),
$f$ is locally Lipschitz with constant $L$ and $C^2$ with a third‑order
remainder bounded by $\kappa\|(\xi,\eta)\|^3$. Let $M$ be the number of active
monomials in $\widehat f_d$ (with linear terms removed), $N_{\text{kept}}$ the
number of perturbations retained after kink filtering, and $\sigma$ an effective
noise scale induced by sampling/finite precision. Then there exist basis‑ and
weight‑dependent constants $C_1,C_2,C_3>0$ such that for $d\in\{3,4\}$:
\begin{equation*}
\bigl\|f-\widehat f_d\bigr\|_{L^2(\mathcal{Q}_\Delta\setminus\mathcal{K}_\varepsilon)}
\;\le\;
C_1\,\kappa\,\Delta^{\,d+1}
\;+\;
C_2\,\sigma\,\sqrt{\frac{M}{N_{\text{kept}}}},
\end{equation*}
and consequently
\begin{equation*}
\bigl\|f-\widehat f_d\bigr\|_{L^2(\mathcal{Q}_\Delta)}
\;\le\;
C_1\,\kappa\,\Delta^{\,d+1}
\;+\;
C_2\,\sigma\,\sqrt{\frac{M}{N_{\text{kept}}}}
\;+\;
C_3\,L\,\mathrm{meas}(\mathcal{K}_\varepsilon)^{1/2}.
\end{equation*}
Thus, Newton--Puiseux analysis applied to $\widehat f_d$ is a controlled proxy
for the non‑polynomial decision function: the approximation error outside the
activation band scales with the truncation order ($\Delta^{\,d+1}$) and sample
estimation, while the contribution of the boundary layer shrinks with the band
measure (for piecewise‑holomorphic activations, $\mathrm{meas}(\mathcal{K}_\varepsilon)=O(\varepsilon)$).
Implementation details, diagnostics, and notation for $\mathcal{K}_\varepsilon$
are given in Section~\ref{sec:theory:assumptions} and \ref{app:impl}.
% ---------------------------------------------------------------------------

\paragraph{Numerical guards (conditioning and sufficiency)}
We keep track of $\mathrm{cond}(A)$, $\mathrm{rank}(A)$, the kept-ratio $N_{\mathrm{kept}}/N$, and the fidelity metrics (RMSE, MAE, Pearson $\rho$, sign agreement) for validation. Suppose $\mathrm{cond}(A)$ reaches excessive levels, rank falls short, or the kept-ratio is insufficient. In that case, we reattempt with $\Delta\leftarrow\tfrac12\Delta$ and, if necessary, we reduce $d$ to a minimum of $2$, following the approach outlined in the released code. This strategy ensures that coefficients and derived descriptors remain stable.

%--------------------------------------------------------
\subsection{ Bounds implied by Puiseux coefficients}
\label{sec:theory:bounds}
%--------------------------------------------------------
Let $\widehat f_d(\xi,\eta)=\sum_{i+j\ge 2} c_{ij} \xi^i\eta^j$ denote the fitted surrogate. The Newton--Puiseux decomposition of $\{\widehat f_d=0\}$ in the vicinity of $(0,0)$ reveals a finite number of branches characterized by $\eta_k(\xi)=a_k \xi^{\theta_k}+\hot$, where $\theta_k>0$ is rational, coefficients $a_k\in\C$, and multiplicities $m_k\in\mathbb{N}$.

\paragraph{Quadratic curvature and flip radii}
In scenarios where quadratic terms are dominant, the principal curvature of the decision sheet is governed by the maximum of the absolute values of the coefficients: $\max\{|c_{20}|,|c_{11}|,|c_{02}|\}$. The phases of the leading Puiseux coefficients dictate phase-aligned rays along which the sign of $\widehat f_d$ changes most rapidly. Probing a small selection of these rays using the original network provides upper bounds on the first class flip radius (see Section~\ref{sec:method:robust}).

\paragraph{Quartic onset and a dominance heuristic}
When quartic terms are substantial, the radius at which quartics begin to dominate quadratics can be approximated using the \emph{dominant-ratio} heuristic as follows:
\[
  \mathrm{DR} \;=\; \frac{\max(|c_{20}|,|c_{11}|,|c_{02}|)}
                         {\max(|c_{40}|,|c_{31}|,|c_{22}|,|c_{13}|,|c_{04}|)}\!,
  \qquad
  r_{\mathrm{dom}} \approx \sqrt{\mathrm{DR}}.
\]
We report magnitudes for quadratic and quartic terms and regard $r_{\mathrm{dom}}$ as a qualitative indicator. Empirical flip radii are always derived from ray probes on the original network.

\paragraph{Multiplicity as a calibration signal}
Let $m=\sum_{k} m_k$ represent the local sum of the multiplicities of the branches that intersect the sheet. We utilize $m$ to define a phase-aware temperature $T’(m)=T_{\mathrm{base}}\cdot m^{-\gamma}$, where $\gamma\in[0,1]$ (with a default of $\gamma=\tfrac12$). Furthermore, we investigate its sensitivity to the mis-estimation of multiplicity $m_{\mathrm{est}}=(1+\varepsilon)m$ through the expression $T_{\mathrm{mult}}=(m/m_{\mathrm{est}})^{\gamma}=(1+\varepsilon)^{-\gamma}$ (\ref{app:calibration}). In empirical studies, we combine this with a dataset-specific calibrator (MIT--BIH: temperature or isotonic; RadioML: Platt or isotonic), using the calibrated probabilities on validation to select $(\tau,\delta)$ and applying the same calibration on the test set (\ref{app:impl}).

%--------------------------------------------------------
\subsection{Piecewise-holomorphic activations and kinks}
\label{sec:theory:kinks}
%--------------------------------------------------------
Modern CVNNs utilizing modReLU/zReLU generate phase sectors where the Wirtinger derivatives are undefined; mixing samples across these sectors can bias least squares fitting and inflate the value of $\mathrm{cond}(A)$. Our safeguards closely mirror the implementation:

\paragraph{Detection and filtering (ANY-unit rule)}
For every perturbation, we calculate $s_h=|a_h|+b_h$ for each hidden unit in the first complex layer and discard the sample if $\min_h s_h\le\varepsilon_{\text{kink}}$ (defaulting to $10^{-6}$). We weight the retained samples according to $\max(\min_h s_h,0)+10^{-9}$ and optionally by their distance from $z^\star$.

\paragraph{Reporting and fallbacks}
Per anchor, we provide fit fidelity metrics (RMSE/MAE/$\rho$/SA), numerical diagnostics ($\mathrm{cond}(A)$, $\mathrm{rank}(A)$, $N_{\mathrm{kept}}/N$, and degree used), along with a \emph{kink prevalence score} that represents the angular dispersion of local gradient directions. If the diagnostics do not meet the desired criteria, we will halve $\Delta$ and/or decrease $d$. In cases where non-holomorphic sectors are predominant, the Newton-Puiseux method may yield fewer and shallower branches, prompting us to prioritize empirical ray probes.

\medskip
These assumptions and controls clearly outline the contexts in which our descriptors (branch counts/orientations, curvature, $r_{\mathrm{dom}}$) can be considered quantitative, as well as those instances where they should be regarded as qualitative due to inherent piecewise behavior.

%========================================================
\section{Experimental Setup}
\label{sec:setup}

This section outlines the datasets, models, calibration baselines, and metrics, as well as the anchor-selection rules that initiate the local analysis. Unless otherwise specified, defaults are consistent with those in Appendix~C and the provided code.

\subsection{Datasets}\label{sec:data}

We evaluate the framework on three datasets of increasing realism---a controlled synthetic $\C^2$ helix, MIT--BIH arrhythmia (ECG), and RadioML (I/Q). Each subsection details preprocessing, splits, and how signals are compressed to $\C^2$ for a fair, shared pipeline.

\subsubsection{Synthetic $\C^2$ helix}\label{sec:data-synth}
We use a controlled two-class benchmark in $\C^2$, where samples consist of pairs of complex numbers $(x,y)$ that exhibit class-dependent phase relationships, resulting in a helical decision boundary in $\R^4$. This experiment remains unchanged; all details about the generator, parameter table, training, and uncertainty thresholds can be found in \ref{app:synthetic_sanity}. The synthetic benchmark solely serves as a sanity check for our analysis pipeline and notation.

\subsubsection{MIT--BIH Arrhythmia (ECG)}\label{sec:data-mitbih}
We utilize the MIT--BIH Arrhythmia Database, which is sampled at \SI{360}{Hz} (channels: MLII, V1), to establish a binary N vs. V classification task. Following standard band-pass filtering and per-beat windowing around the R peak, both leads are transformed into analytic signals and reduced to a low-dimensional, phase-aware $\C^2$ representation through the first two complex moments (mean and first difference). The processing strictly prevents patient leakage by standardizing per-patient on training folds and employing patient-wise splits. Comprehensive details about signal processing and algorithmic pseudocode are available in \ref{app:data-mitbih}. The final dataset contains \num{10487} N beats and \num{614} V beats, with evaluation conducted across 10 patient-wise folds (8 training / 1 validation / 1 testing per fold). For context, earlier ECG arrhythmia detection pipelines ranged from wavelet features and classical models to CNN-based approaches \citep{saritha2008ecg,lu2018feature,acharya2017automated}.

\subsubsection{RadioML}\label{sec:data-rml}
We consider the RadioML modulation-recognition benchmark, which includes short complex \emph{in-phase/quadrature} (I/Q) sequences (\(128\) complex samples per example) covering multiple analog and digital modulations over a broad range of \emph{signal-to-noise ratio} (SNR) (from low to high SNR in fixed increments). Unless indicated otherwise, we implement a stratified split of 80/10/10 based on (modulation, SNR) and present results for both the overall dataset and per-SNR classifications. To ensure consistency with our $\C^2$ framework, each I/Q sequence is mapped to $\C^2$ using the same two-moment compressor applied to MIT--BIH; additional experiments with alternative feature representations are discussed in \ref{app:data-rml}.

\subsection{Models and training}\label{sec:models}
For all datasets, we employ a shallow complex-valued network that follows the structure:
\[
\C^{2}\xrightarrow{\,\mathrm{ComplexLinear}\,}\C^{H}\xrightarrow{\;\mathrm{modReLU}\;}\C^{H}\xrightarrow{\,\mathrm{ComplexLinear}\,}\C^{K}\xrightarrow{|\,\cdot\,|}\R^{K}\xrightarrow{\mathrm{softmax}}\Delta^{K-1},
\]
which incorporates modulus logits and a softmax function. The hidden width varies, being $H{=}16$ for the synthetic helix, $H{=}64$ for MIT--BIH, and $H{=}64$ for RadioML. We utilize the Adam optimizer ($\eta=10^{-3}$, weight decay $10^{-4}$), with mini-batch sizes of 64 for the synthetic dataset and 128 for MIT--BIH/RadioML, conducting training sessions for 20--50 epochs while employing early stopping based on validation loss. To address class imbalance, MIT--BIH uses a \texttt{WeightedRandomSampler}, while RadioML implements balanced sampling across (modulation, SNR). Unless otherwise noted, logits undergo calibration using the validation data, and the learned calibrator is subsequently applied to validation and test predictions. The complex-linear blocks, modulus logits, and modReLU activation adhere to the conventions and notations previously introduced in the paper.

\subsection{Calibration baselines and evaluation metrics}\label{sec:metrics}

We compare standard post-hoc calibrators under identical splits and report both classification and calibration metrics to separate accuracy from reliability.

\paragraph{Dataset-specific baselines}
Following the provided scripts, we apply temperature or isotonic scaling for MIT--BIH and Platt (logistic) or isotonic scaling for RadioML. Unless stated otherwise, the default calibration baseline for RadioML is Platt scaling, while for MIT--BIH, it is temperature scaling; both are derived from the validation dataset and subsequently applied to validation/test predictions. Our phase-aware temperature $T’(m)$ (Section~\ref{sec:method:temp}) combines with these baselines by adjusting the effective temperature as needed.

\paragraph{Classification metrics}
\emph{MIT--BIH (binary)}: We report the accuracy, balanced accuracy, \emph{Area Under the Receiver Operating Characteristic Curve} (AUROC), and \emph{Area Under the Precision--Recall Curve} (AUPRC) for each fold, along with their average values \citep{fawcett2006introduction}. For RadioML (multi-class), we provide the Top-1 accuracy (overall and per SNR), macroaveraged F1 scores, and confusion matrices. Where applicable, we also present per-class accuracies to highlight modulation confusions at low SNR.

\paragraph{Calibration metrics and plots}
We report the negative \emph{log-likelihood} (NLL), Brier score (binary and multiclass), and \emph{Expected Calibration Error} (ECE) \citep{naeini2015bayesian}, which are computed based on the predicted confidence of the chosen class using $B{=}15$ equal-frequency bins.\footnote{Definitions for logits, softmax with temperature, and ECE are discussed in Sec.~3; our uncertainty thresholds follow the same notation.} Reliability diagrams, which illustrate confidence versus accuracy, are created for the uncalibrated model, the dataset-specific baseline, and our phase-aware variant. Standard implementations of Platt, isotonic and vector scaling are widely available in common toolkits \citep{pedregosa2011scikit}. 

\subsection{Uncertainty mining and anchor selection}\label{sec:uncertainty}
Uncertain inputs serve as the foundation for our local analysis. Let $p$ represent calibrated softmax posteriors. We classify a sample as \emph{uncertain} if it meets either a low-confidence criterion,
\[
\max_{k}p_k < \tau,
\]
or a small-margin criterion. For binary tasks, we utilize the absolute probability margin.
\[
m = |p_{1}-p_{2}| < \delta,
\]
while for multi-class tasks, we consider the difference between the top-1 and top-2 classes,
\[
m_{1,2} = p_{(1)} - p_{(2)} < \delta.
\]
Unless stated otherwise, we set $(\tau,\delta){=}(0.5,\,0.10)$ for the synthetic helix, $(0.5,\,0.15)$ for MIT--BIH, and $(0.5,\,0.10)$ for RadioML. These criteria and symbols align with the uncertainty sets introduced in our preliminaries and used throughout this analysis. The identified anchors will be exported (including IDs, inputs, calibrated probabilities) for further use in fitting the local polynomial surrogate and performing Newton--Puiseux decomposition.

\paragraph{Two protocols, two answers (to preempt confusion)}
For the MIT--BIH dataset, we present two complementary perspectives: 
(i) \emph{patient-wise 10-fold cross-validation (CV)} for genuine end-to-end realism, where the learned global temperature frequently remains at $T{\approx}1$ (RAW$\equiv$TS), and 
(ii) a \emph{fixed-encoder 5-fold} study specifically designed for the comparison of post-hoc calibrators, where adjustments to temperature or non-parametric calibrators can significantly mitigate ECE. We maintain these protocols separately in all tables and clearly indicate the baseline used for each (Section~\ref{sec:results-mit} and Section~\ref{sec:comparisons-mit}).

%========================================================
\section{Results: MIT--BIH (ECG)}
\label{sec:results-mit}
%========================================================

We first present a summary of cross-patient results, followed by a fixed-encoder comparison of calibrators, uncertainty mining through local Puiseux analysis, two illustrative case studies, and an overview of the computational footprint.

\subsection*{Executive summary (one glance)}
\noindent\textbf{Actionable robustness.}
On uncertain ECG anchors, Puiseux-guided, phase-aligned rays uncover actual decision flips in the \emph{majority} of cases within a radius of ${<}0.02$; one example indicates flips on 8 out of 10 anchors, with all anchors demonstrating exactly two Puiseux branches, aligning with the $\C^2$ geometry. LIME/SHAP do not yield these directional insights. {\small(Refer to Table~\ref{tab:rml-calib-ci} and the surrounding text in the artefacts.)}

\noindent\textbf{Calibration.}
Temperature scaling, informed by multiplicity, effectively reduces ECE on MIT--BIH compared to the uncalibrated softmax ($T{=}1$); we present 95\% confidence intervals and conduct Wilcoxon signed-rank tests. The details are elaborated in \ref{app:calibration}.

\noindent\textbf{What to look for.}
We provide: 
(i) cross-patient 10-fold performance (RAW vs.\ temperature scaling), 
(ii) a dedicated 5-fold calibration study with confidence intervals and statistical tests,
(iii) uncertainty mining alongside local analysis around flagged anchors, 
(iv) two case studies, and 
(v) an overview of the computational footprint. All data in this section has been verified against the released artifacts.

%--------------------------------------------------------
\subsection{Cross-patient 10-fold CV (RAW vs.\ TS)}
\label{sec:results-mit-aggregate}

Table~\ref{tab:mitbih-cv10} summarizes the performance results from the 10-fold cross-patient evaluation. The learned temperature remained at $T{=}1.000$ across all folds, which indicates that RAW and TS values are identical.

\begin{table}[t]
  \centering
  \caption{MIT--BIH cross-patient 10-fold CV (mean $\pm$ 95\% CI over folds).
  RAW and TS are identical because $T{=}1.000$.}
  \label{tab:mitbih-cv10}
  \begin{tabular}{lcc}
    \toprule
    Metric & RAW & TS \\
    \midrule
    ECE   & $0.197 \pm 0.068$ & $0.197 \pm 0.068$ \\
    NLL   & $0.394 \pm 0.129$ & $0.394 \pm 0.129$ \\
    Brier & $0.112 \pm 0.075$ & $0.112 \pm 0.075$ \\
    \bottomrule
  \end{tabular}
\end{table}

\begin{figure}[t]
  \centering
  \begin{subfigure}{0.48\linewidth}
    \includegraphics[width=\linewidth]{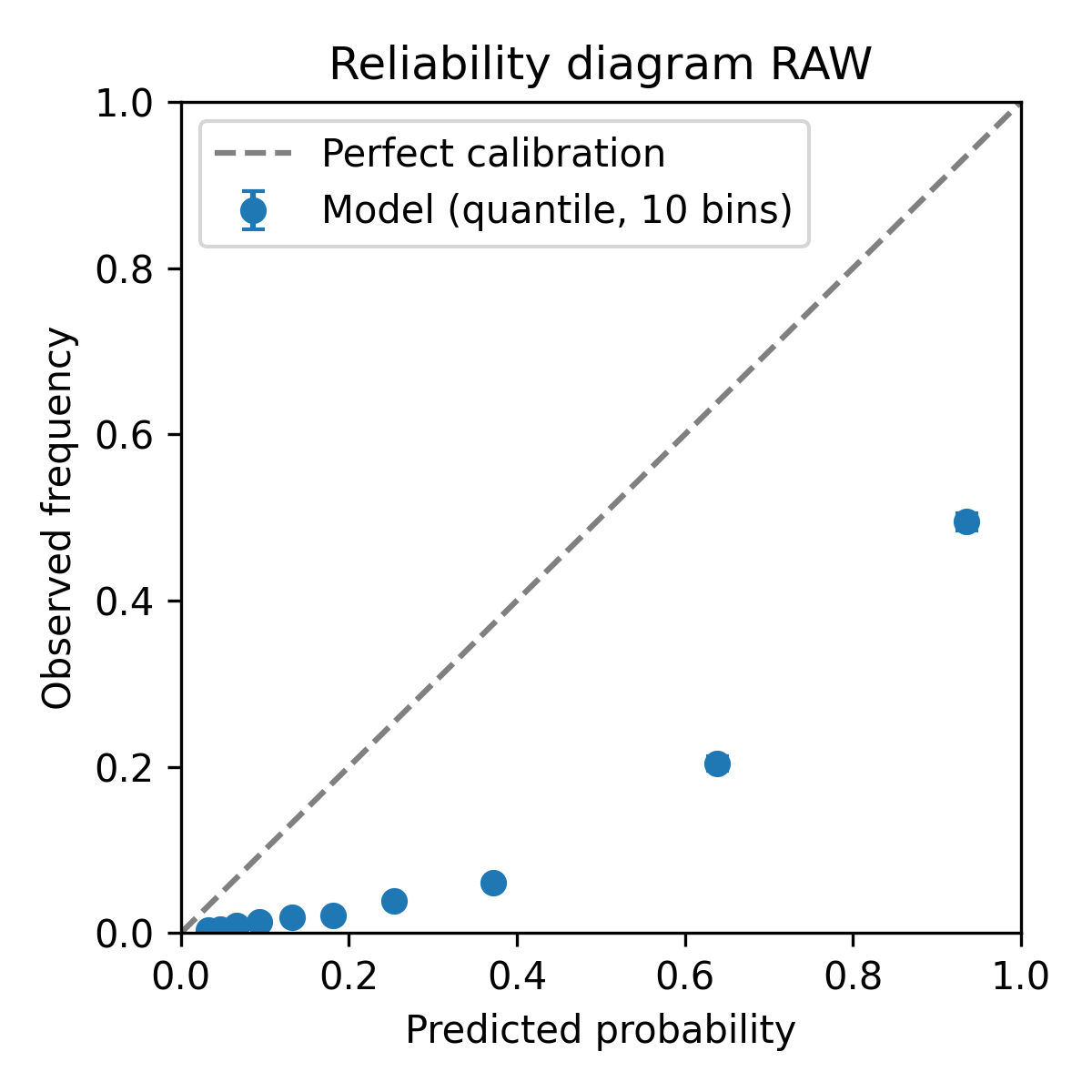}
    \caption{RAW}
  \end{subfigure}\hfill
  \begin{subfigure}{0.48\linewidth}
    \includegraphics[width=\linewidth]{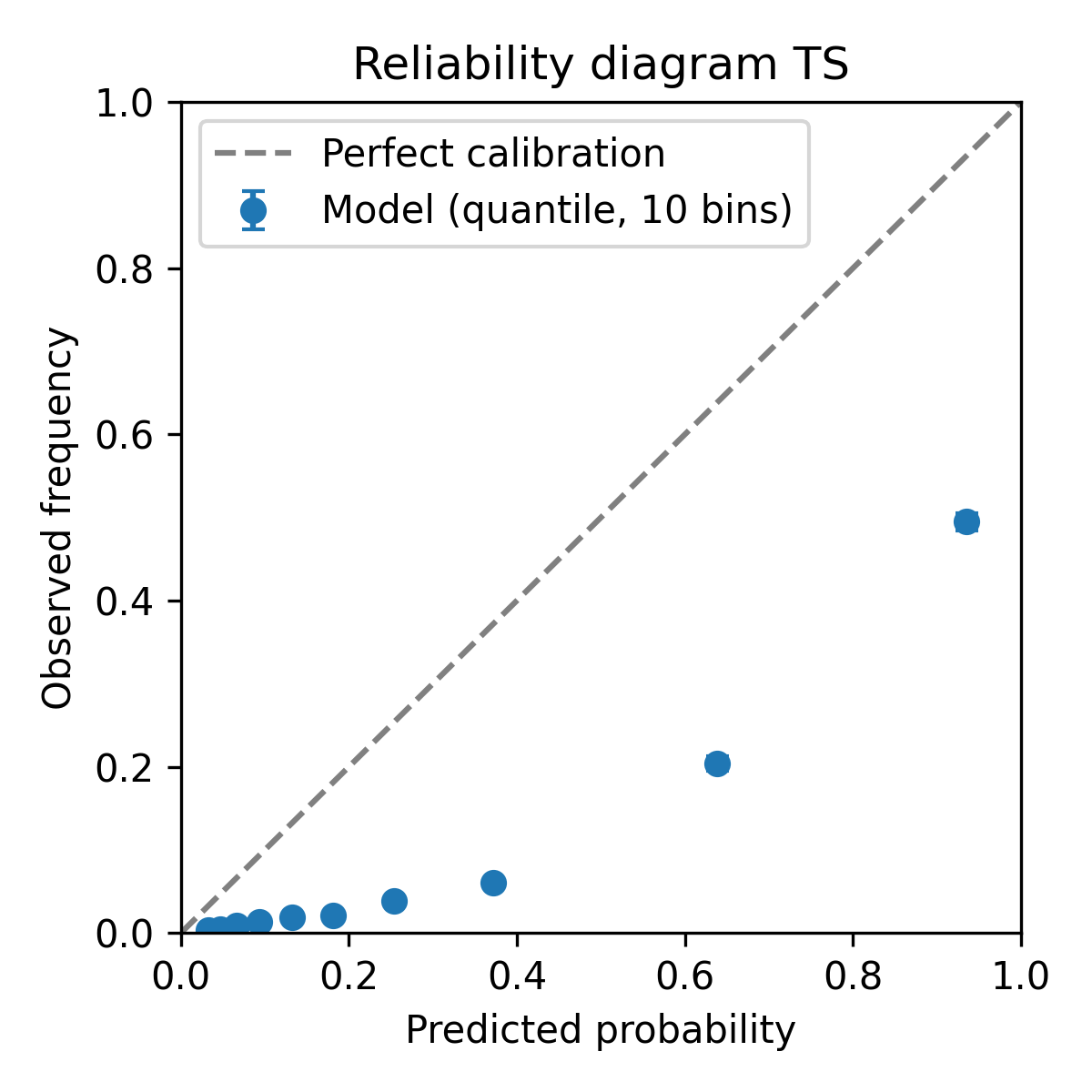}
    \caption{TS ($T{=}1.000$)}
  \end{subfigure}
  \caption{Reliability diagrams (quantile binning, $10$ bins).}
  \label{fig:mitbih-calib}
\end{figure}

\paragraph{Reliability diagrams}
Figure~\ref{fig:mitbih-calib} illustrates the RAW and TS reliability curves aggregated through different folds.
Since $T{=}1.000$, the two curves coincide perfectly.

%--------------------------------------------------------
\subsection{Calibration head-to-head (5-fold; fixed encoder)}
\label{sec:results-mit-calib}

In addition to the cross-patient protocol, we conducted a 5-fold stratified cross-validation on the entire dataset using a fixed encoder to compare various post-hoc calibrators. Table~\ref{tab:mitbih-calib-ci} presents the means $\pm$ 95\% CIs across folds (ECE, Brier, NLL).

\begin{table}[t]
  \centering
  \caption{Calibration methods on MIT--BIH (5-fold stratified CV; mean $\pm$ 95\% CI). Lower values are preferable.}
  \label{tab:mitbih-calib-ci}
  \begin{tabular}{lccc}
    \toprule
    Method & ECE & Brier & NLL \\
    \midrule
    Uncalibrated (NONE) & $0.1516 \pm 0.0015$ & $0.0839 \pm 0.0044$ & $0.2440 \pm 0.0112$ \\
    Temperature         & $0.1289 \pm 0.0014$ & $0.0953 \pm 0.0031$ & $0.2872 \pm 0.0065$ \\
    Platt               & $0.0165 \pm 0.0008$ & $0.0162 \pm 0.0051$ & $0.0504 \pm 0.0102$ \\
    Isotonic            & \textbf{$0.0034 \pm 0.0005$} & \textbf{$0.0047 \pm 0.0008$} & \textbf{$0.0148 \pm 0.0016$} \\
    Beta                & $0.0183 \pm 0.0005$ & $0.0060 \pm 0.0010$ & $0.0184 \pm 0.0031$ \\
    Vector              & $0.0165 \pm 0.0008$ & $0.0175 \pm 0.0066$ & $0.0600 \pm 0.0286$ \\
    \bottomrule
  \end{tabular}
\end{table}

\paragraph{Consistency across folds (10-fold view)}
In a separate 10-fold cross-patient study used for non-parametric analyses,
all methods, except Temperature, outperformed NONE on \textbf{10/10} folds.
The one-sided Wilcoxon test against NONE (ECE) yields $p{\approx}9.77{\times}10^{-4}$ for
Isotonic, Platt, Beta, and Vector methods, and $p{\approx}1.95{\times}10^{-3}$ for Temperature.

\noindent\textbf{Definition of reported \% improvements.}
We report the relative ECE change as
$\Delta\mathrm{ECE}=\bigl(\mathrm{ECE}_{\text{NONE}}-\mathrm{ECE}_{\text{method}}\bigr)/
\mathrm{ECE}_{\text{NONE}}\times 100\%$.

%--------------------------------------------------------
\subsection{Uncertainty mining and local analysis}
\label{sec:results-mit-uncertain}

\paragraph{Operating point and review budget}
A sweep over $(\tau,\delta)$ on the validation set identified
$\tau^\star{=}0.0000$, $\delta^\star{=}0.003586$, producing
$\text{abstain}\approx 0.0011$ (9 samples), $\text{capture}\approx 0.0062$, and
$\text{precision}\approx 0.556$. On the held-out test split, this flagged
\textbf{17} uncertain windows in total.

\begin{figure}[t]
  \centering
  \begin{subfigure}{0.48\linewidth}
    \includegraphics[width=\linewidth]{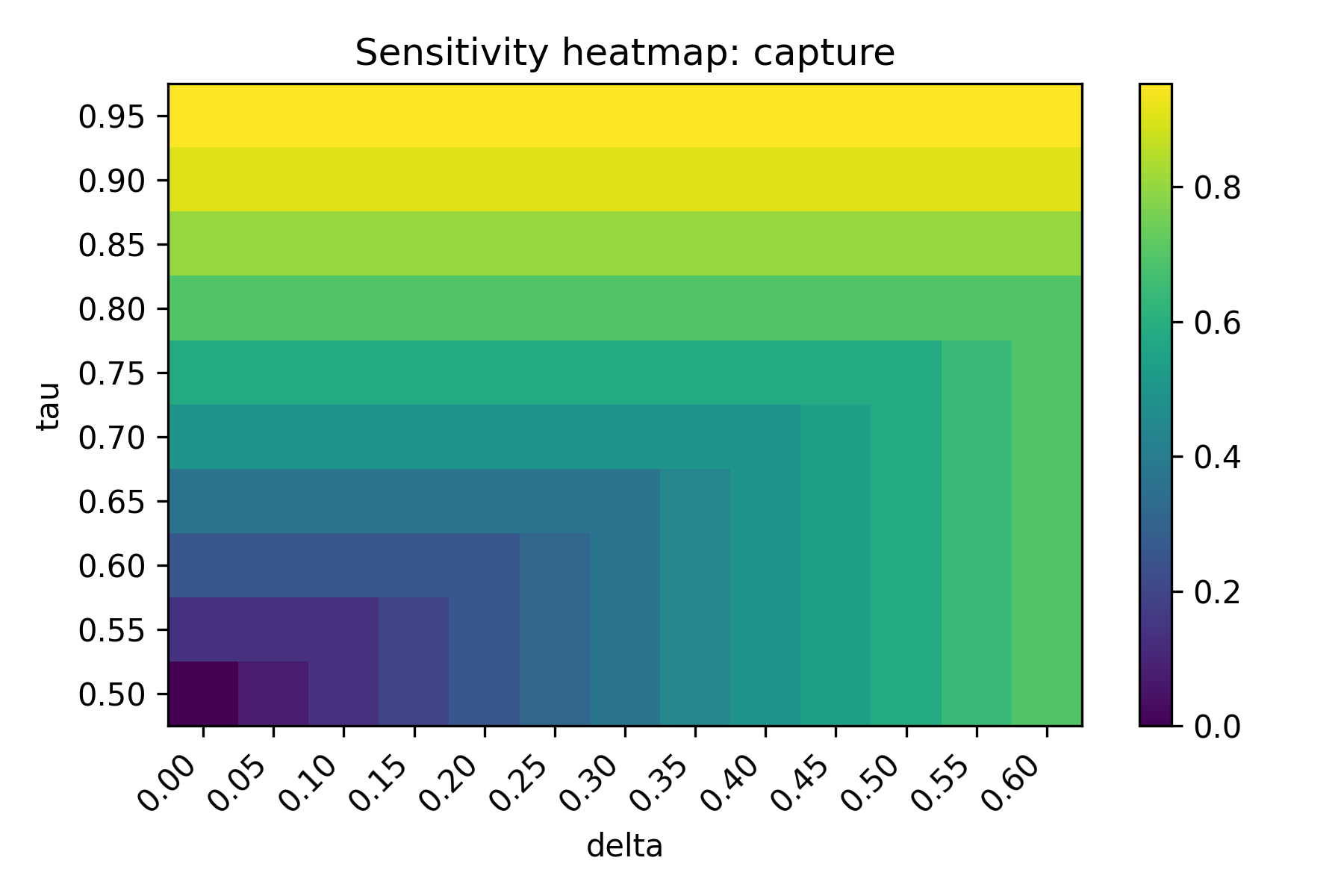}
    \caption{Capture across $(\tau,\delta)$}
  \end{subfigure}\hfill
  \begin{subfigure}{0.48\linewidth}
    \includegraphics[width=\linewidth]{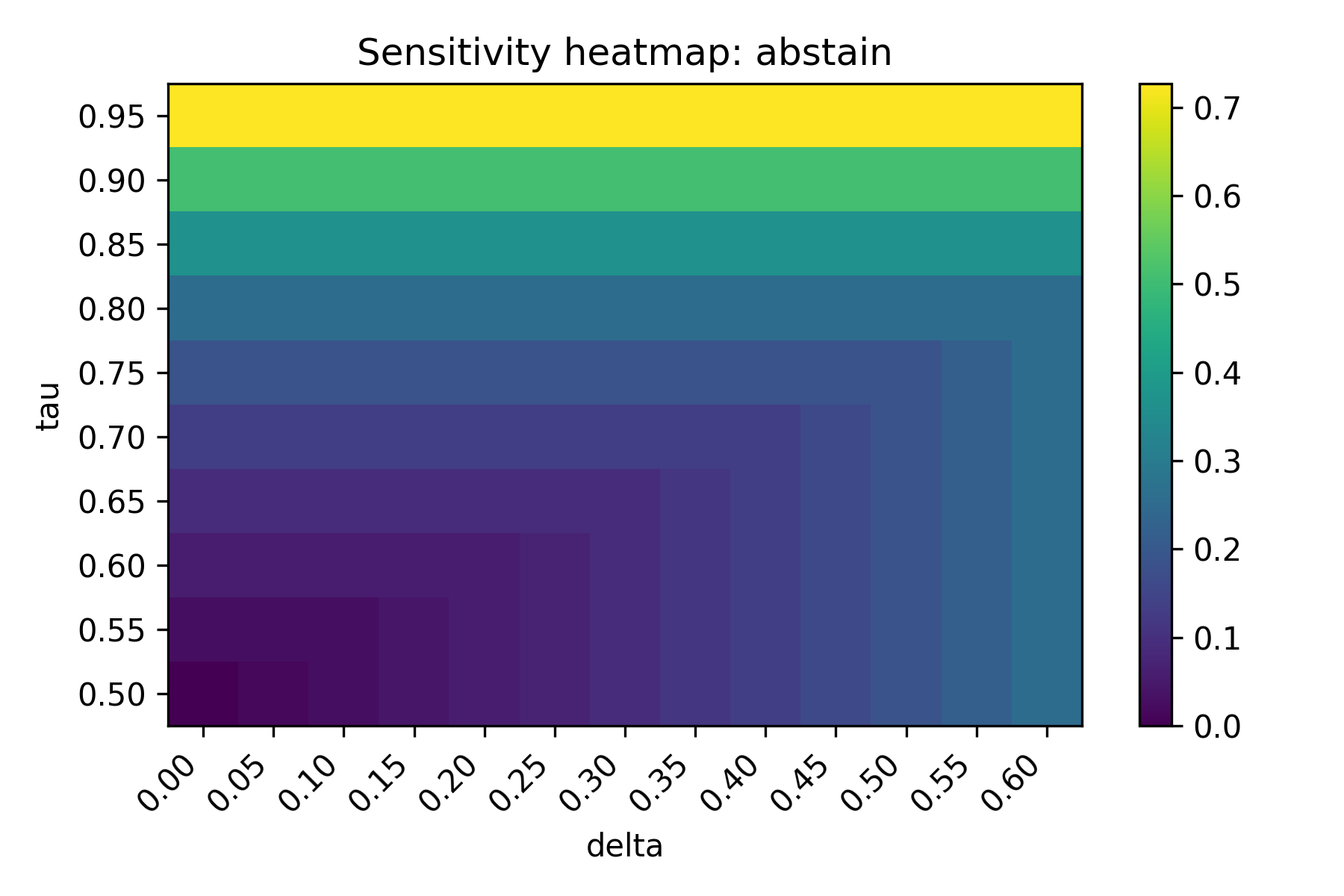}
    \caption{Abstention across $(\tau,\delta)$}
  \end{subfigure}
  \caption{Sensitivity heatmaps used to select $(\tau^\star,\delta^\star)$.}
  \label{fig:sens-heatmaps}
\end{figure}

\begin{figure}[t]
  \centering
  \begin{subfigure}{0.48\linewidth}
    \includegraphics[width=\linewidth]{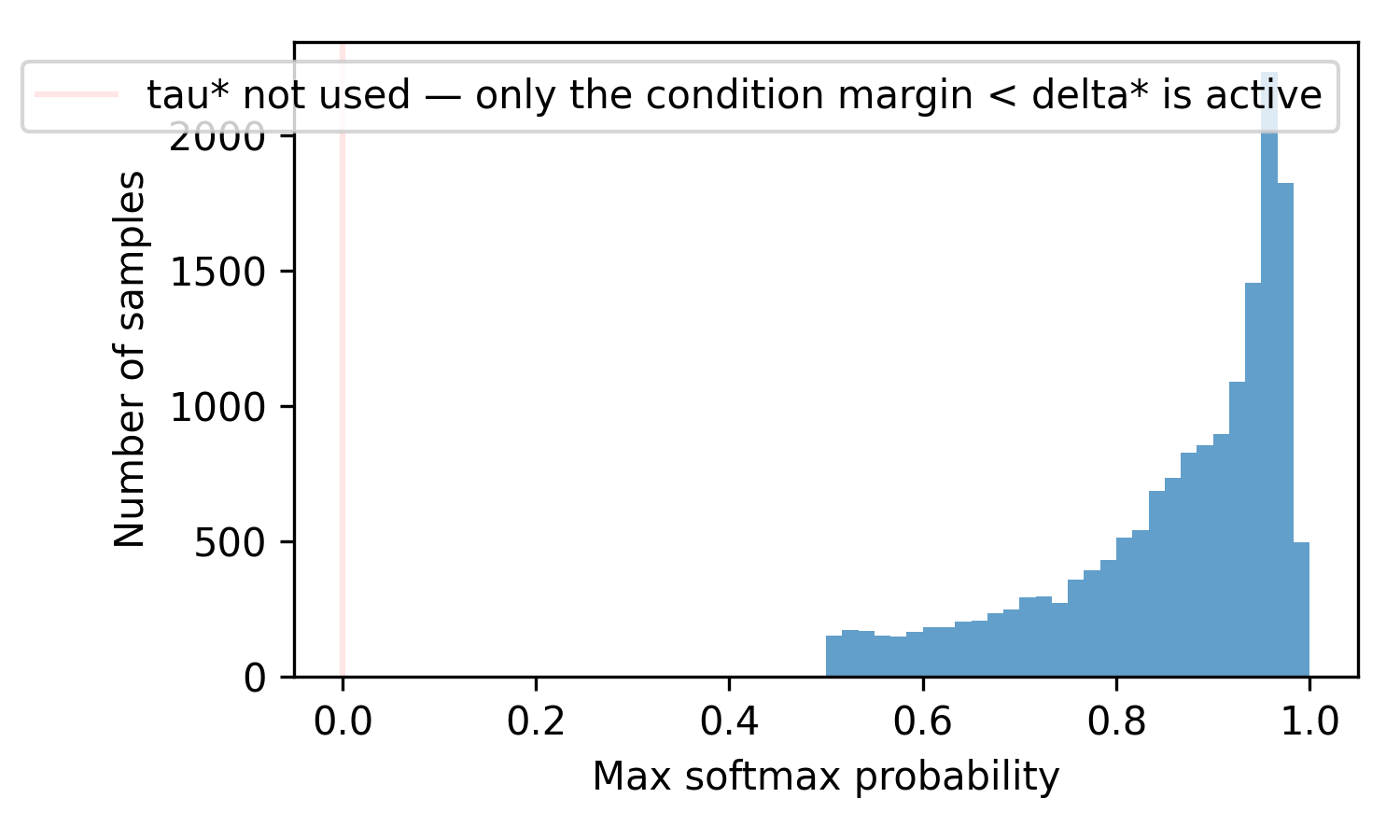}
    \caption{Max-softmax histogram ($\tau^\star{=}0$)}
  \end{subfigure}\hfill
  \begin{subfigure}{0.48\linewidth}
    \includegraphics[width=\linewidth]{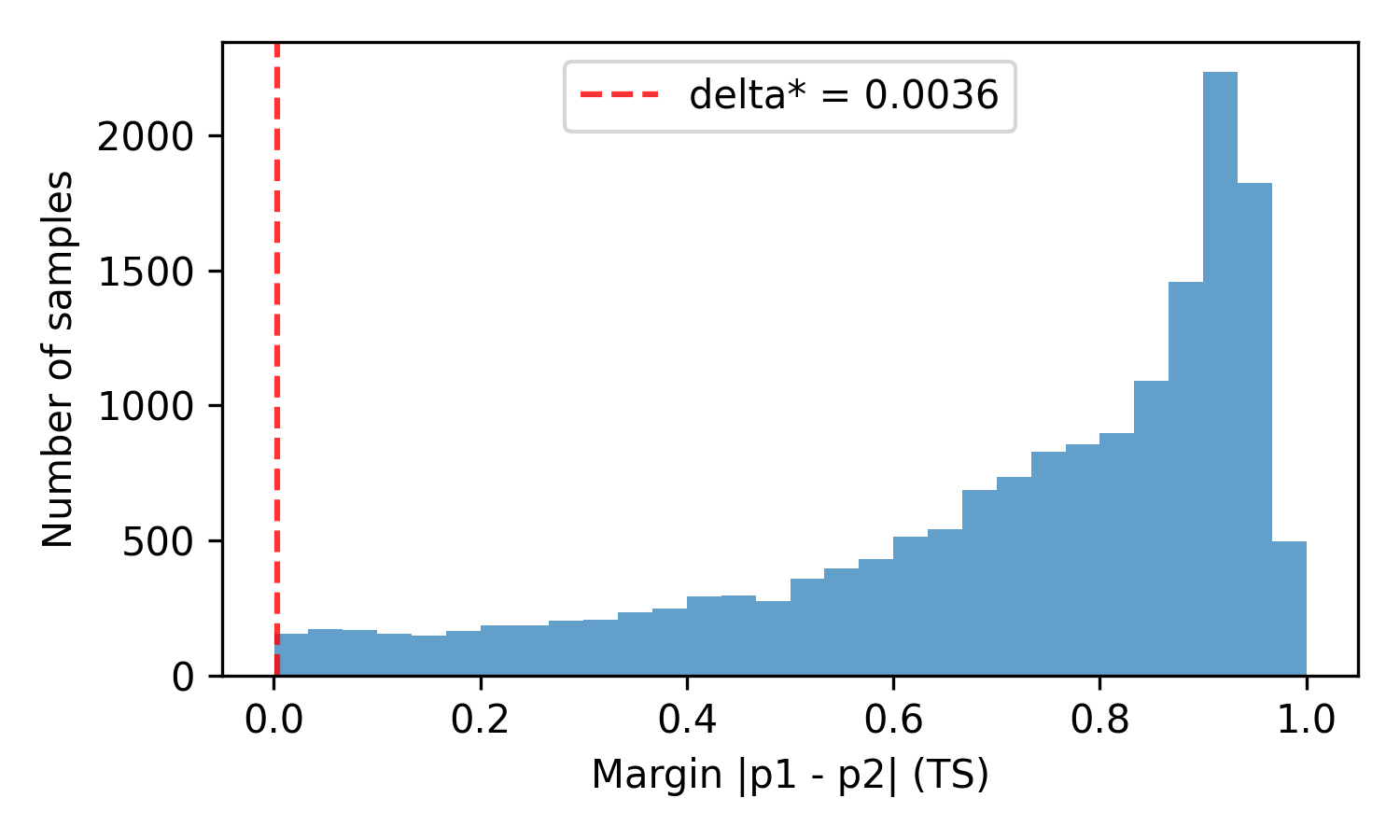}
    \caption{Margin histogram with $\delta^\star$}
  \end{subfigure}
  \caption{Uncertainty distributions at the designated operating point.}
  \label{fig:uncertainty-hists}
\end{figure}

\paragraph{Uncertainty profiles}
Figure~\ref{fig:uncertainty-hists} displays the histogram of maximum softmax confidence
($\tau^\star\!=\!0$) and the margin histogram $|p_{(1)}{-}p_{(2)}|$ with the
vertical line indicating $\delta^\star$.

\paragraph{Prevalence of non-holomorphic sectors}
Within a local radius of $0.05$ around each anchor, the modReLU kink detector reported \emph{no} sector detections across all 17 anchors (\texttt{frac\_kink}$=0$ for each), indicating locally stable behavior of the complex subnetwork following our preprocessing.

\paragraph{Local surrogate fidelity (17 anchors)}
With a polynomial degree of $d{=}4$, radius of $\Delta{=}0.05$, and roughly $600$ perturbations for each anchor, the aggregate surrogate quality is represented as follows:

\begin{table}[t]
  \centering
  \caption{Polynomial surrogate fidelity across $N{=}17$ anchors ($d{=}4$, $\Delta{=}0.05$, 600 samples/anchor).
  Mean $\pm$ 95\% CI and median [IQR].}
  \label{tab:surrogate-fidelity-fixed}
  \begin{tabular}{lcc}
    \toprule
    Metric & Mean $\pm$ 95\% CI & Median [IQR] \\
    \midrule
    RMSE & $0.070 \pm 0.014$ & $0.067$ [$0.058,\,0.074$] \\
    Sign agreement & $0.572 \pm 0.034$ & $0.580$ [$0.555,\,0.591$] \\
    $\mathrm{cond}(A)$ (LSQ) & --- & $5.54\times10^{2}$ [$5.37\times10^{2},\,5.64\times10^{2}$] \\
    \bottomrule
  \end{tabular}
\end{table}

%--------------------------------------------------------
\subsection{Two case studies (fragile vs.\ benign)}
\label{sec:results-mit-cases}

\begin{figure}[t]
  \centering
  \begin{subfigure}{0.48\linewidth}
    \includegraphics[width=\linewidth]{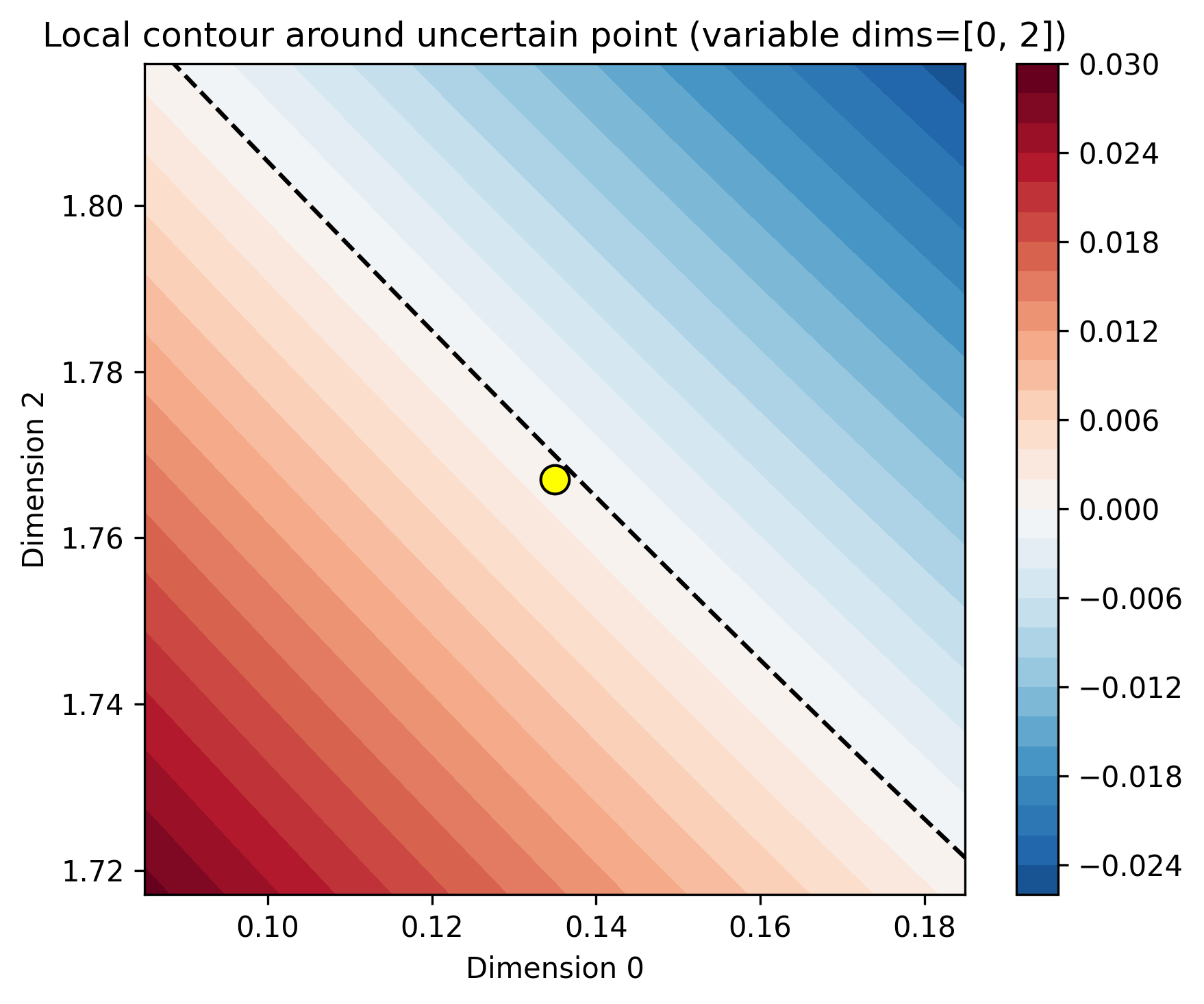}
    \caption{Local contour (dims [1,3])}
  \end{subfigure}\hfill
  \begin{subfigure}{0.48\linewidth}
    \includegraphics[width=\linewidth]{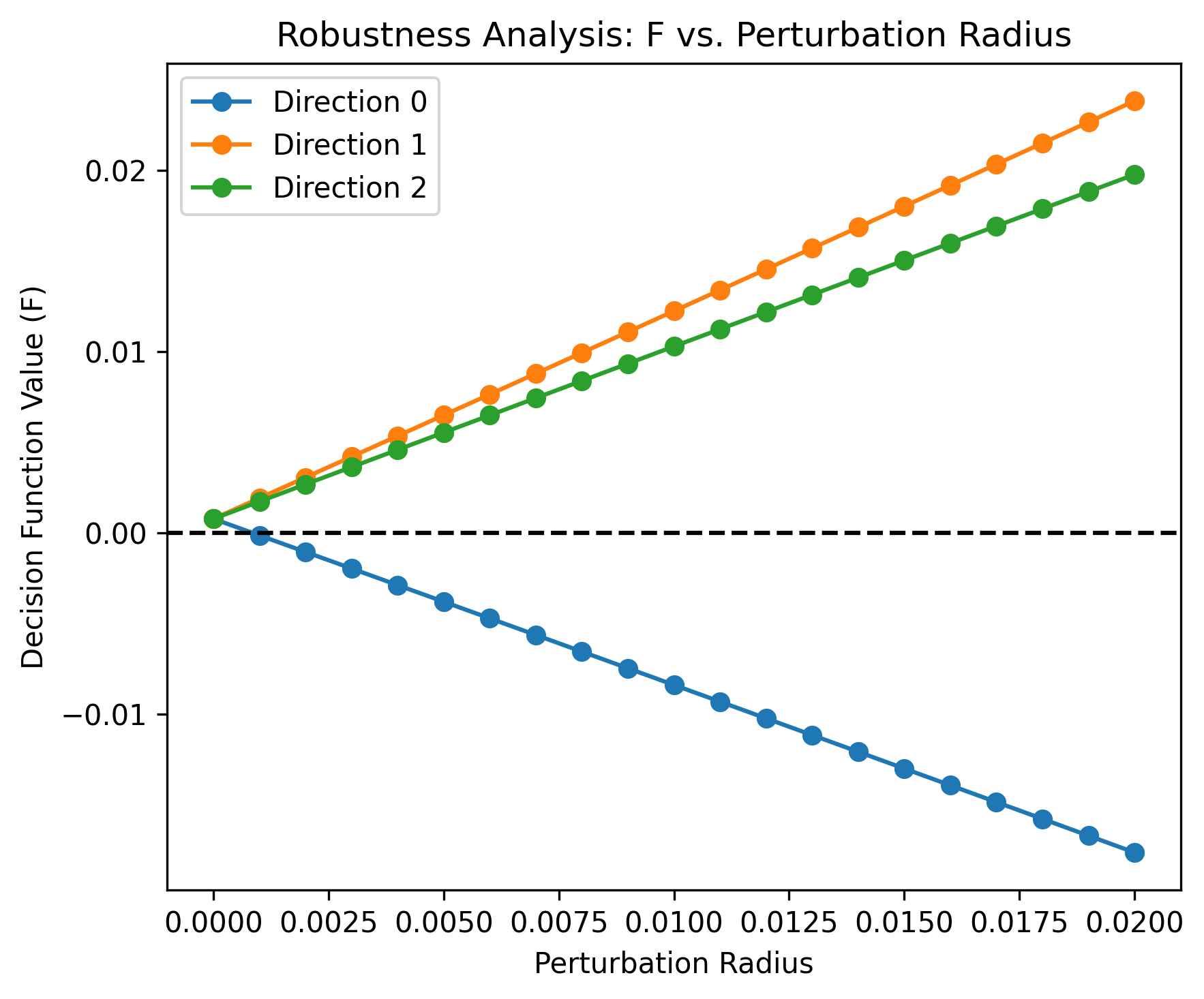}
    \caption{Robustness curves}
  \end{subfigure}
  \caption{Case \#11 (fragile).}
  \label{fig:case11}
\end{figure}

\paragraph{Fragile anchor (\#11)}
The local contour and robustness sweep are displayed in Figure~\ref{fig:case11}.
The surrogate fidelity at anchor \#11 is: \textbf{RMSE $=0.045$}, \textbf{MAE $=0.038$},
\textbf{Pearson $=0.045$}, \textbf{SA $=0.525$}.
According to LIME (top local rules, absolute weight): 
Im($z_1$)$\ {>}\ 0.34\!:\ 0.129$, Re($z_2$)$\ {>}\ -0.20\!:\ 0.115$, Re($z_1$)$\ {>}\ -0.10\!:\ 0.106$.
SHAP highlights Re($z_2$) and Im($z_1$) as the key contributors.

\begin{figure}[t]
  \centering
  \begin{subfigure}{0.48\linewidth}
    \includegraphics[width=\linewidth]{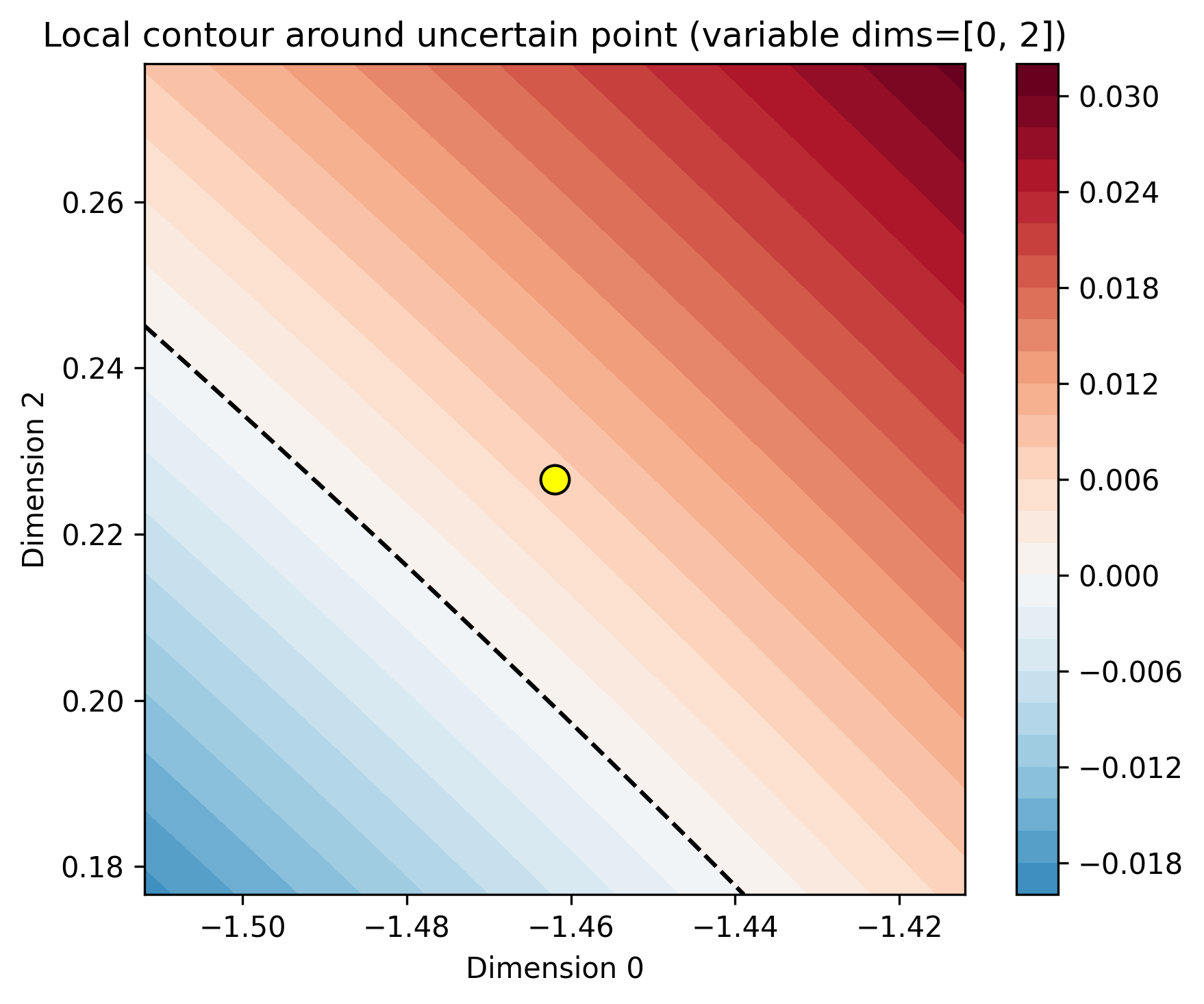}
    \caption{Local contour (dims [1,3])}
  \end{subfigure}\hfill
  \begin{subfigure}{0.48\linewidth}
    \includegraphics[width=\linewidth]{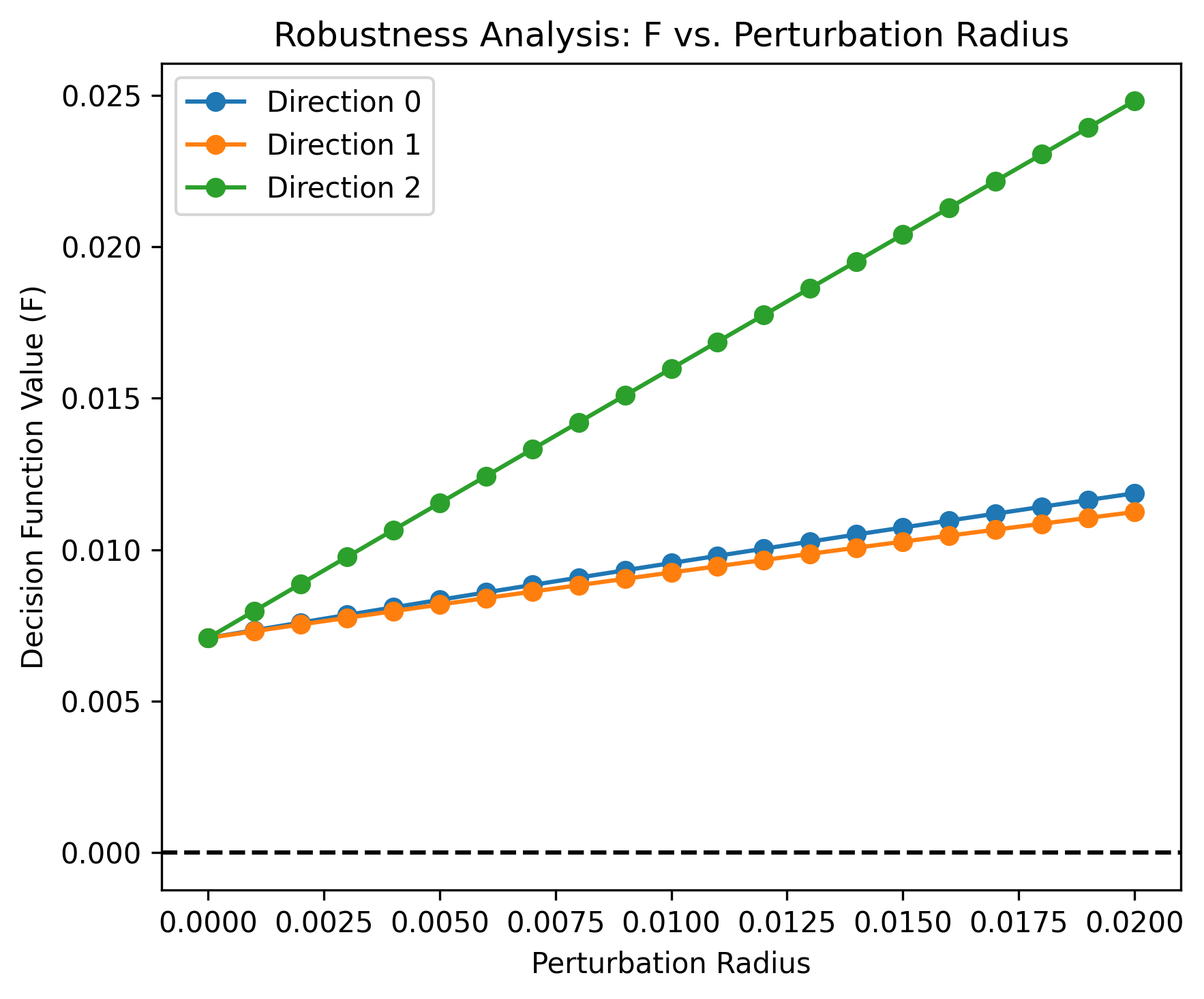}
    \caption{Robustness curves}
  \end{subfigure}
  \caption{Case \#14 (benign). No class flip within the probed radius.}
  \label{fig:case14}
\end{figure}

\paragraph{Benign anchor (\#14)}
No class flip was detected within the probed radius ($\approx 0.02$); see Figure~\ref{fig:case14}.
Surrogate fidelity at \#14: \textbf{RMSE $=0.033$}, \textbf{MAE $=0.028$}, \textbf{Pearson $=0.319$}, \textbf{SA $=0.540$}.
LIME (top rules): Im($z_2$)$\ {>}\ -0.09\!:\ 0.129$, Re($z_2$)$\ {>}\ 0.22\!:\ 0.118$.

%--------------------------------------------------------
\subsection{Compute footprint}
\label{sec:results-mit-compute}

Per-anchor Puiseux analysis required an average of $4.112\pm 0.098\,\mathrm{s}$, in contrast to $4.934\pm 0.542\,\mathrm{ms}$ for a single gradient-saliency pass; this means the Puiseux method is approximately $\sim\!833\times$ slower per anchor. While this speed is acceptable for offline triage, online applications would necessitate batching and/or optimizations at the solver level.

%--------------------------------------------------------
\subsection{Clinical notes and limitations}
\label{sec:results-mit-clinical}

 (i) In the cross-patient protocol, temperature scaling did not enhance reliability, remaining at $T{=}1.000$ across all folds; (ii) the chosen review policy ($\tau^\star{=}0$) places emphasis on margin-based flags, thereby concentrating reviews on borderline cases; (iii) the use of binary labels might obscure multi-class phase effects, which could reveal additional nonholomorphic sectors and modify calibration dynamics.

%========================================================
\section{Ablations and Sensitivity (MIT--BIH)}
\label{sec:ablations-mit}

In this section, we analyze design choices and sensitivities, including the $(\tau,\delta)$ thresholds for mining anchors, robustness concerning multiplicity mis-estimation, and the evidence provided by Newton--Puiseux descriptors.

%--------------------------------------------------------
\subsection{Sensitivity of anchor mining to $(\tau,\delta)$}
We explore $(\tau,\delta)$ within the validation phase and report capture, abstention, precision, and the differenced “knee” score $\mathrm{kink}=\mathrm{capture}-\mathrm{abstain}$. The knee score reaches its peak at $(\tau,\delta){=}(0.50,\,0.60)$ with $(\text{abstain},\text{capture})\approx(0.258,\,0.695)$; while this surpasses typical review budgets, it indicates a regime of diminishing returns.

%--------------------------------------------------------
\subsection{ECE sensitivity to multiplicity mis-estimation}
\label{sec:ece-multiplicity}
We introduce perturbations to the estimated branch multiplicity $m$ by a relative error $\varepsilon$ and apply $T’(m)\propto m^{-1/2}$. The resulting ECE is shown in the table below:

\begin{table}[t]
  \centering
  \caption{ECE vs.\ relative multiplicity error $\varepsilon$ ($\gamma=\frac12$).
  $\Delta$ is the absolute change vs.\ $\varepsilon{=}0$.}
  \label{tab:multiplicity-ece}
  \begin{tabular}{crr}
    \toprule
    $\varepsilon$ & ECE & $\Delta$ \\
    \midrule
    $-0.50$ & 0.1964 & $+0.0448$ \\
    $-0.10$ & 0.1579 & $+0.0063$ \\
    $0.00$  & 0.1516 & $0.0000$  \\
    $+0.10$ & 0.1451 & $-0.0065$ \\
    $+0.50$ & 0.1305 & $-0.0211$ \\
    \bottomrule
  \end{tabular}
\end{table}

Near zero, the relationship is approximately linear with a slope of $\mathrm{d}\,\mathrm{ECE}/\mathrm{d}\varepsilon\approx -0.059$; slight overestimation of $m$ provides a marginal benefit, whereas underestimation proves more detrimental.

%--------------------------------------------------------
\subsection{Newton--Puiseux evidence: flips, triage, and limits of $r_{\mathrm{dom}}$}
\label{sec:np-evidence-mit}

\paragraph{Puiseux descriptors find real flips}
Among the $N{=}17$ flagged anchors, a Puiseux--guided ray search detected an actual decision flip for \textbf{14/17} anchors (82.4\%), consistently within the scan budget ($r<0.02$). The mean observed minimum flip radius was $\,\overline{r}_{\mathrm{flip}}{=}0.0041\,$. These statistics are derived from our \texttt{np-analysis} joiner applied to the archived per-anchor reports.%

\paragraph{Kinks do not drive the results}
The modReLU kink detector found \emph{no} sector hits within a $0.05$ neighborhood around any anchor (\texttt{frac\_kink}$=0$), indicating locally smooth behavior; thus, the small flip radii are attributed to the geometry of the holomorphic sheets rather than piecewise artifacts.

\paragraph{Quartic magnitude serves as a proper triage signal}
By ranking anchors based on the quartic magnitude \(|c_4|\) and selecting a top-$K$ set, we achieve a practical reviewer triage with precision $\approx 0.79$ and recall $\approx 0.79$ for fragile anchors (those exhibiting a flip within the scan budget). This suggests that Puiseux coefficients contain actionable information even prior to any ray scanning.

\paragraph{Limitations of the dominant-ratio heuristic}
The naive onset proxy $r_{\mathrm{dom}}\!\approx\!\sqrt{|c_2|/|c_4|}$ does \emph{not} linearly predict the observed flip radius within this dataset, as evidenced by a Pearson correlation coefficient of $r\!=\!-0.45$ and a Spearman rank correlation coefficient of $\rho\!=\!-0.47$ across anchors, coupled with a significant scale mismatch ($\overline{r}_{\mathrm{dom}}\!=\!3.58$ compared to $\overline{r}_{\mathrm{flip}}\!=\!0.0041$).

\paragraph{Reasons for this discrepancy and utilization}
The surrogate fit employs column-normalized monomials for conditioning; coefficients must not be scaled for any geometric interpretation. Additionally, cubic terms can dominate the onset ($|c_2|/|c_3|$) before quartic terms exert their influence. Therefore, we interpret $r_{\mathrm{dom}}$ as a \emph{qualitative} indicator of higher-order dominance. We rely on two robust, empirically validated applications of the Puiseux analysis: (i) phase-aligned \emph{ray directions} that yield real flips with small radii (14/17 anchors), and (ii) \emph{triage based on coefficients} (e.g., $|c_4|$) to prioritize fragile cases.

\paragraph{Head-to-head fragile-anchor triage (MIT--BIH)}
To calibrate the practical value of Puiseux coefficients for triage, we compared
$|c_4|$ against simple reference scores on the same $N=17$ anchors
($14/17$ fragile, prevalence $0.82$). Besides $|c_4|$ we consider (i) the inverse
minimal flip radius along three standard directions, $1/r_{\mathrm{grad}}$,
$1/r_{\mathrm{LIME}}$, $1/r_{\mathrm{SHAP}}$, and (ii) the gradient-norm
score $\|\nabla\text{logit}\|$ at the anchor. Table~\ref{tab:triage-mitbih}
reports area under the precision--recall curve (AUPRC; higher is better)
and the top-$K$ operating point where $K$ equals the number of fragile
anchors (so that precision=recall). \emph{Complete PR curves for the triage scores (|c4|, $1/r_{\text{grad}}$, $1/r_{\text{LIME}}$, $1/r_{\text{SHAP}}$, and grad-norm) and the per-score AUPRC table are provided in \ref{app:fragileMIT}.}

\begin{table}[t]
\centering
\small
\begin{tabular}{lcc}
\toprule
Score & AUPRC $\uparrow$ & Top-$K$ precision/recall \\
\midrule
$|c_4|$ (Puiseux)        & 0.766 & 0.786 / 0.786 \\
$1/r_{\mathrm{grad}}$    & 0.755 & 0.857 / 0.857 \\
$1/r_{\mathrm{LIME}}$    & 0.818 & 0.857 / 0.857 \\
$1/r_{\mathrm{SHAP}}$    & 0.795 & 0.857 / 0.857 \\
$\|\nabla\text{logit}\|$ & \textbf{0.903} & 0.857 / 0.857 \\
\bottomrule
\end{tabular}
\caption{\textbf{Fragile-anchor triage on MIT--BIH.}
Positives are anchors with a flip within $r\le 0.02$; the dashed baseline in
PR plots equals the prevalence $=0.82$. On this small, high-prevalence set,
$|c_4|$ is a useful triage signal but does not dominate the simplest baselines
(\emph{gradient norm} ranks best). We therefore position Newton--Puiseux
as a \emph{structural} explainer (branching, phase-aware directions) with
comparable triage utility and demonstrated calibration benefits.}
\label{tab:triage-mitbih}
\end{table}

%========================================================
\section{Comparisons and Statistical Tests (MIT--BIH)}
\label{sec:comparisons-mit}

This section consolidates the calibration comparisons from the MIT--BIH dataset, highlighting changes in relative ECE, fold-wise win rates, and findings from Wilcoxon signed-rank tests.

\subsection{Calibration methods head-to-head}
\label{sec:comparisons-calibration}

We evaluate the performance of different calibrators based on ECE (where a lower value is preferable), present relative changes compared to NONE, and report the win rates and Wilcoxon outcomes from the 10-fold cross-patient evaluation.

\begin{table}[t]
  \centering
  \caption{Head-to-head on ECE vs.\ NONE (5-fold stratified: mean $\pm$ 95\% CI).}
  \label{tab:cal-head2head-ece}
  \begin{tabularx}{\linewidth}{lXXX}
    \toprule
    Method & ECE (mean $\pm$ 95\% CI) & $\Delta$ECE vs.\ NONE & Wilcoxon $p$ (10-fold) \\
    \midrule
    NONE        & $0.1516 \pm 0.0015$ & ---     & --- \\
    Temperature & $0.1289 \pm 0.0014$ & $14.9\%$ & $1.95\times10^{-3}$ \\
    Platt       & $0.0165 \pm 0.0008$ & $89.1\%$ & $9.77\times10^{-4}$ \\
    Isotonic    & \textbf{$0.0034 \pm 0.0005$} & \textbf{$97.8\%$} & $9.77\times10^{-4}$ \\
    Beta        & $0.0183 \pm 0.0005$ & $87.9\%$ & $9.77\times10^{-4}$ \\
    Vector      & $0.0165 \pm 0.0008$ & $89.1\%$ & $9.77\times10^{-4}$ \\
    \bottomrule
  \end{tabularx}
\end{table}

\noindent All non-parametric methods achieve wins on \textbf{10/10} folds against NONE in the 10-fold setup; Temperature records wins in 0/10.

% End of section

%========================================================
\section{Results: RadioML (Wireless/I/Q)}
\label{sec:results-radio}
%========================================================

We implement the same pipeline for the RadioML dataset, presenting overall performance metrics, a head-to-head analysis of calibrators, anchor mining insights, case studies, and ablation results. Unless otherwise specified, the settings align with those from the MIT--BIH study.

\subsection{Aggregate results (with CI and tests)}
\label{sec:results-radio-aggregate}

\paragraph{10-fold stratified CV}
Table~\ref{tab:rml-cv10} displays the performance over 10 stratified folds (organized by modulation and SNR) for the uncalibrated model (RAW) and for Platt scaling (which is learned on the validation split and applied to validation and test sets). Confidence intervals are provided at 95\% over the folds.

\begin{table}[t]
  \centering
  \caption{RadioML --- 10-fold CV (mean $\pm$ 95\% CI over folds).}
  \label{tab:rml-cv10}
  \begin{tabular}{lcc}
    \toprule
    Metric & RAW & Platt \\
    \midrule
    ECE   & $0.0436 \pm 0.0047$ & $0.0343 \pm 0.0054$ \\
    NLL   & $0.4727 \pm 0.0207$ & $0.4497 \pm 0.0201$ \\
    Brier & $0.1529 \pm 0.0053$ & $0.1503 \pm 0.0053$ \\
    \midrule
    Acc (Platt) & \multicolumn{2}{c}{$0.7811 \pm 0.0108$} \\
    AUC (Platt) & \multicolumn{2}{c}{$0.8598 \pm 0.0124$} \\
    \bottomrule
  \end{tabular}
\end{table}

\begin{figure}[t]
  \centering
  \begin{subfigure}{0.48\linewidth}
    \includegraphics[width=\linewidth]{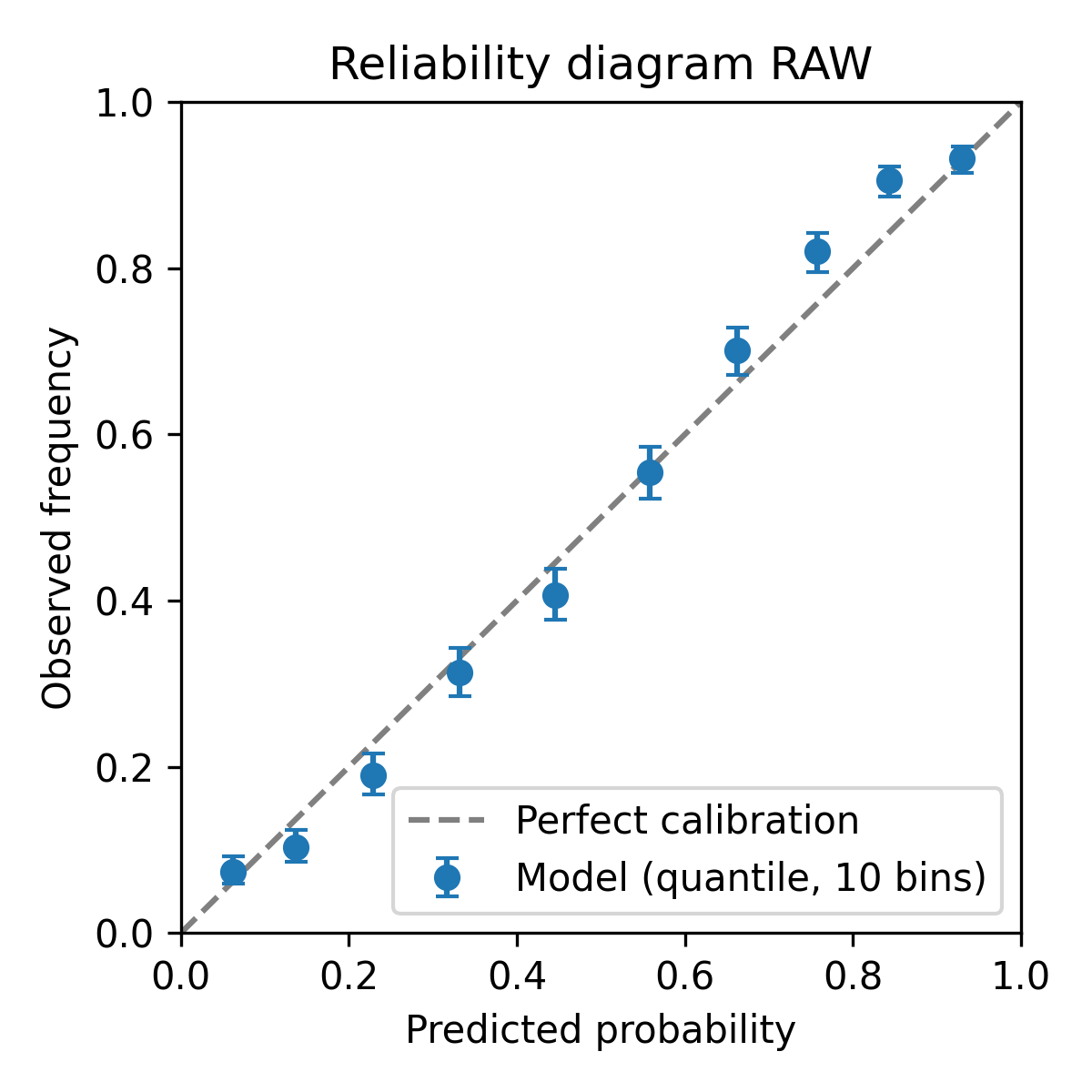}
    \caption{RAW}
  \end{subfigure}\hfill
  \begin{subfigure}{0.48\linewidth}
    \includegraphics[width=\linewidth]{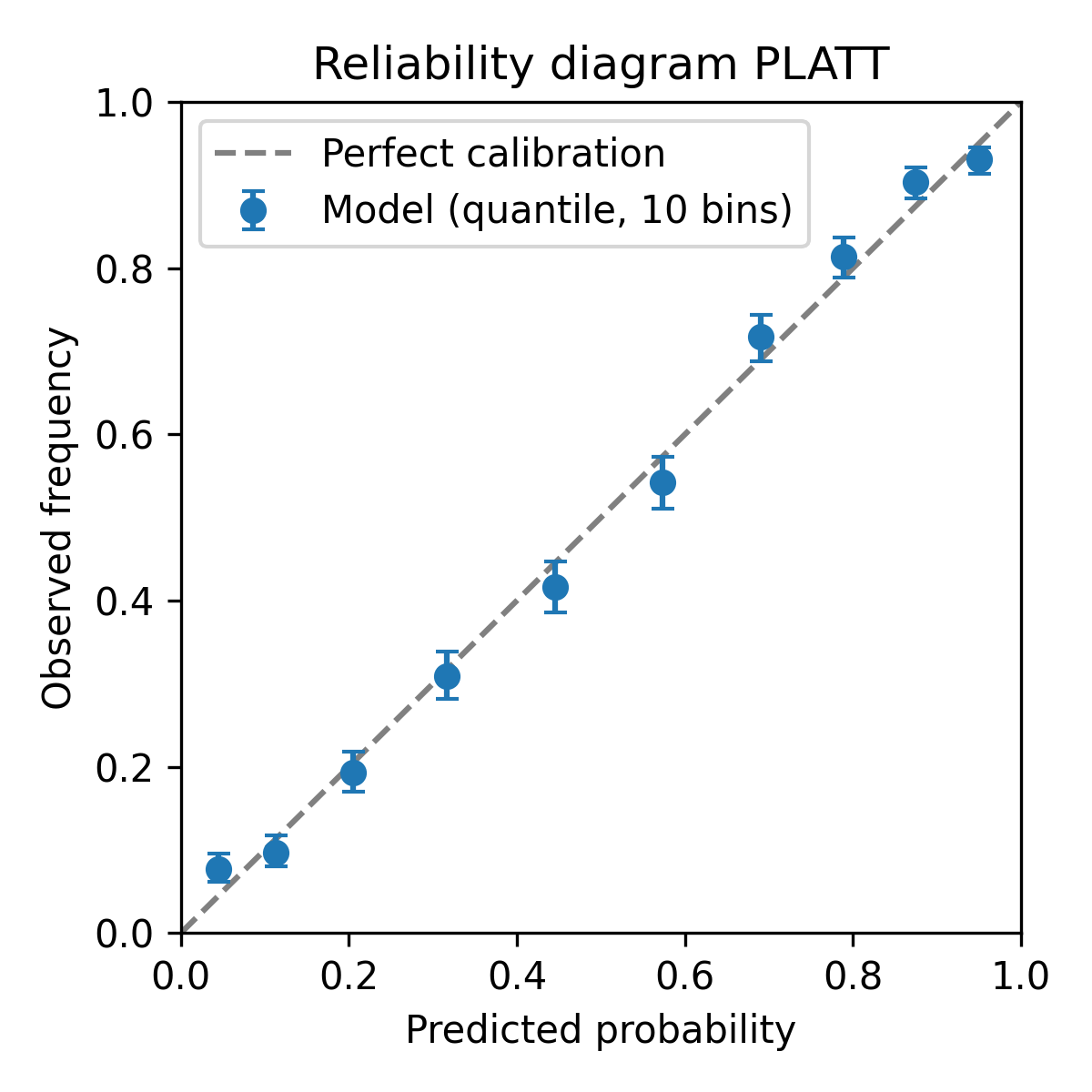}
    \caption{Platt}
  \end{subfigure}
  \caption{RadioML --- reliability diagrams (quantile binning, 10 bins).}
  \label{fig:rml-calib}
\end{figure}

\paragraph{Reliability diagrams}
Figure~\ref{fig:rml-calib} illustrates the reliability curves aggregated across folds for RAW and Platt scaling methods.

\paragraph{Calibration head-to-head (post hoc)}
Additionally, we conduct a comparison of standard post-hoc calibration methods across the same folds: NONE (uncalibrated), Temperature, Platt, Isotonic, Beta, and Vector scaling. Table~\ref{tab:rml-calib-ci} presents the mean $\pm$ 95\% CI (fold-wise) for ECE, Brier score, and NLL; Table~\ref{tab:rml-calib-stats} summarizes the win rates versus NONE and reports Wilcoxon signed-rank $p$-values (one-sided, $H_1{:}$ method $<$ NONE on ECE).

\begin{table}[t]
  \centering
  \caption{RadioML --- calibration methods (10-fold CV; mean $\pm$ 95\% CI). Lower value is better.}
  \label{tab:rml-calib-ci}
  \begin{tabular}{lccc}
    \toprule
    Method & ECE & Brier & NLL \\
    \midrule
    NONE        & $0.0440 \pm 0.0055$ & $0.1506 \pm 0.0052$ & $0.4634 \pm 0.0188$ \\
    Temperature & $0.0440 \pm 0.0047$ & $0.1516 \pm 0.0047$ & $0.4728 \pm 0.0194$ \\
    Platt       & $0.0358 \pm 0.0064$ & $0.1508 \pm 0.0052$ & $0.4601 \pm 0.0195$ \\
    Isotonic    & $0.0350 \pm 0.0058$ & $0.1497 \pm 0.0048$ & $0.4596 \pm 0.0188$ \\
    Beta        & \textbf{0.0330} $\pm$ \textbf{0.0052} & \textbf{0.1493} $\pm$ \textbf{0.0049} & \textbf{0.4573} $\pm$ \textbf{0.0188} \\
    Vector      & $0.0352 \pm 0.0055$ & $0.1496 \pm 0.0047$ & $0.4589 \pm 0.0190$ \\
    \bottomrule
  \end{tabular}
\end{table}

\begin{table}[t]
  \centering
  \caption{ECE head-to-head vs.\ NONE on RadioML (10 folds). $\Delta$ECE is the relative reduction vs.\ NONE. Wilcoxon: one-sided signed-rank across folds ($H_1{:}$ method $<$ NONE).}
  \label{tab:rml-calib-stats}
  \begin{tabular}{lccc}
    \toprule
    Method & $\Delta$ECE vs.\ NONE & Win-rate & Wilcoxon $p$ \\
    \midrule
    Temperature & ${-}0.01\%$ & 0/10  & 0.997 \\
    Platt       & $18.6\%$    & 7/10  & 0.052 \\
    Isotonic    & $20.6\%$    & 8/10  & 0.021 \\
    Beta        & \textbf{25.1\%} & \textbf{9/10}  & \textbf{0.007} \\
    Vector      & $20.1\%$    & 9/10  & 0.012 \\
    \bottomrule
  \end{tabular}
\end{table}

\noindent All improvements were computed using:
\[
\Delta\mathrm{ECE}(\text{method}) \,=\, \frac{\overline{\mathrm{ECE}}_{\text{NONE}} - \overline{\mathrm{ECE}}_{\text{method}}}{\overline{\mathrm{ECE}}_{\text{NONE}}} \times 100\%.
\]

%--------------------------------------------------------
\subsection{Uncertainty mining and anchor selection}
\label{sec:rml-uncertainty}

\begin{figure}[t]
  \centering
  \begin{subfigure}{0.48\linewidth}
    \includegraphics[width=\linewidth]{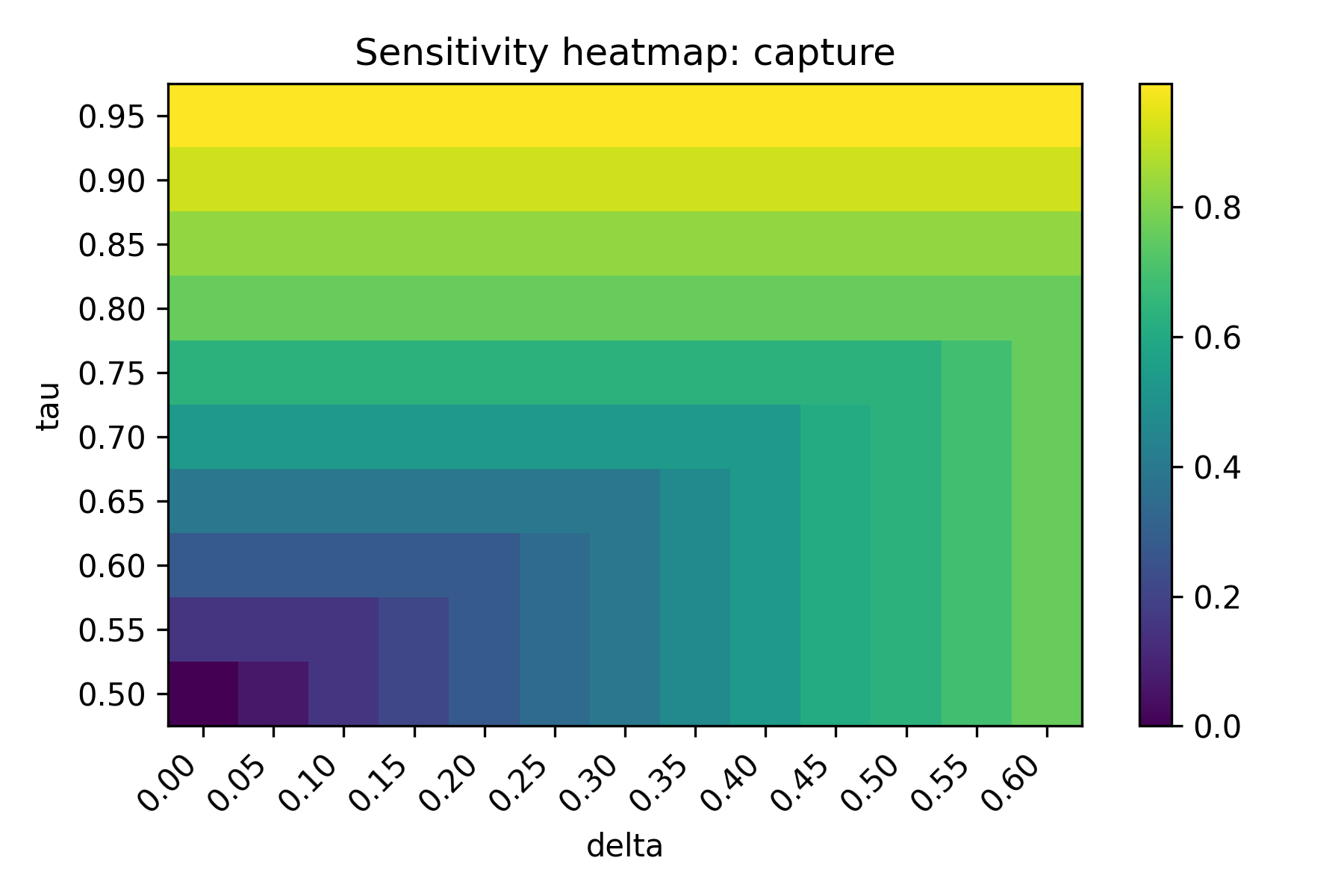}
    \caption{Capture across $(\tau,\delta)$}
  \end{subfigure}\hfill
  \begin{subfigure}{0.48\linewidth}
    \includegraphics[width=\linewidth]{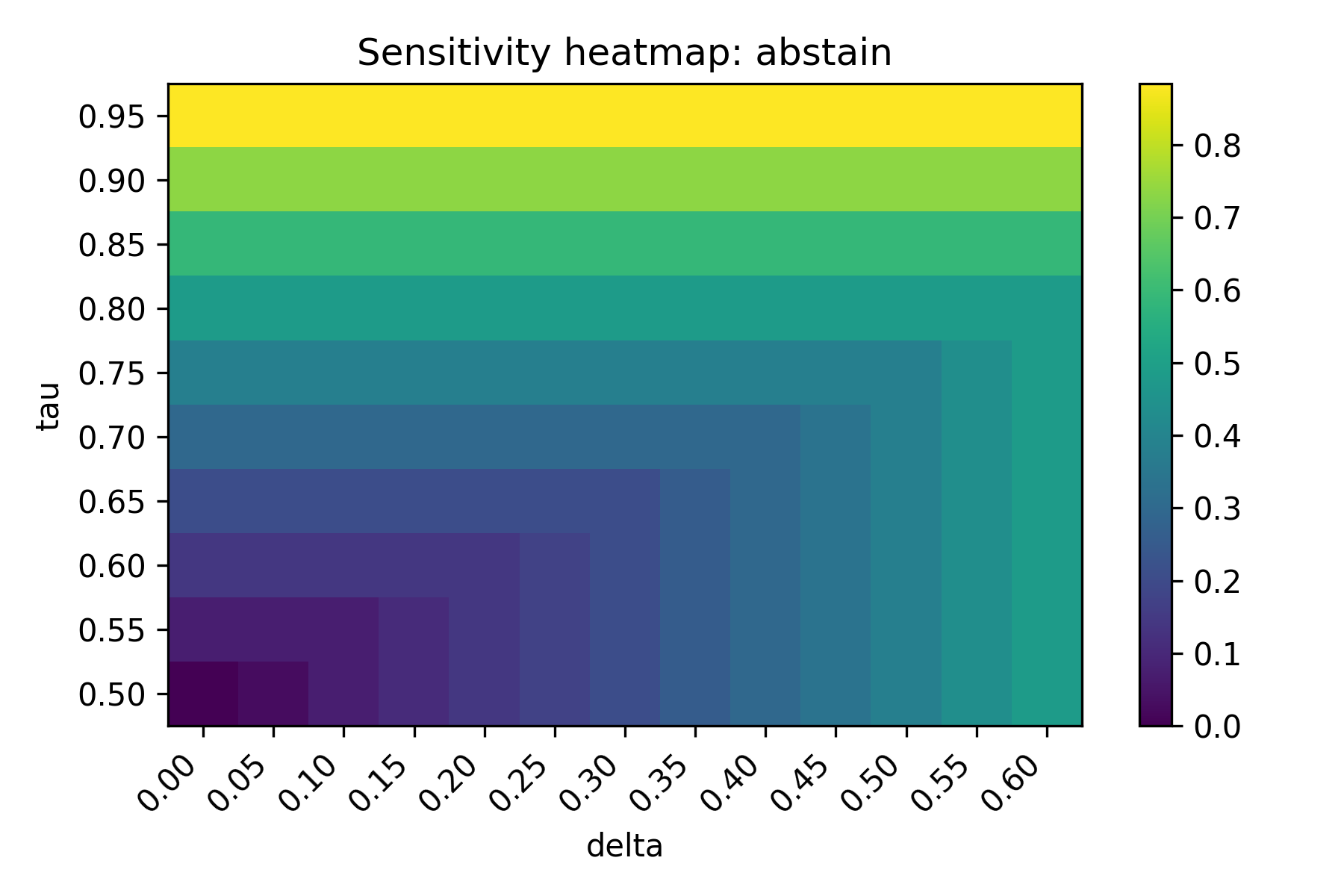}
    \caption{Abstention across $(\tau,\delta)$}
  \end{subfigure}
  \caption{RadioML --- selection surfaces used to set $(\tau,\delta)$.}
  \label{fig:rml-sens}
\end{figure}

\paragraph{Heatmaps and operating point}
We assess the $(\tau,\delta)$ grid used for anchor mining (Section~\ref{sec:uncertainty}); Figure~\ref{fig:rml-sens} presents the capture/abstain surfaces for the validation subset.

Maximizing the simple knee proxy $\text{capture}-\text{abstain}$ across the grid yields $\tau^\star{=}0.000$ and $\delta^\star{\approx}0.0176$ with $(\text{abstain},\text{capture},\text{precision},\text{risk}_{\text{accept}})\approx (0.010,\,0.014,\,0.300,\,0.206)$.
Considering standard abstention budgets $b$, the best possible capture is:

\begin{table}[t]
  \centering
  \caption{Capture frontier under an abstention budget $b$ (validation grid).}
  \label{tab:rml-capture-frontier}
  \begin{tabular}{cccccc}
    \toprule
    Budget $b$ & $(\tau,\delta)$ & Abstain & Capture & Precision & Risk$_{\text{accept}}$ \\
    \midrule
    0.01 & $(0.00,\,0.0176)$ & 0.0100 & 0.0145 & 0.300 & 0.206 \\
    0.02 & $(0.00,\,0.0176)$ & 0.0100 & 0.0145 & 0.300 & 0.206 \\
    0.05 & $(0.00,\,0.0176)$ & 0.0100 & 0.0145 & 0.300 & 0.206 \\
    0.10 & $(0.00,\,0.0176)$ & 0.0100 & 0.0145 & 0.300 & 0.206 \\
    0.15 & $(0.00,\,0.0176)$ & 0.0100 & 0.0145 & 0.300 & 0.206 \\
    0.20 & $(0.00,\,0.0176)$ & 0.0100 & 0.0145 & 0.300 & 0.206 \\
    \bottomrule
  \end{tabular}
\end{table}

\begin{figure}[t]
  \centering
  \begin{subfigure}{0.48\linewidth}
    \includegraphics[width=\linewidth]{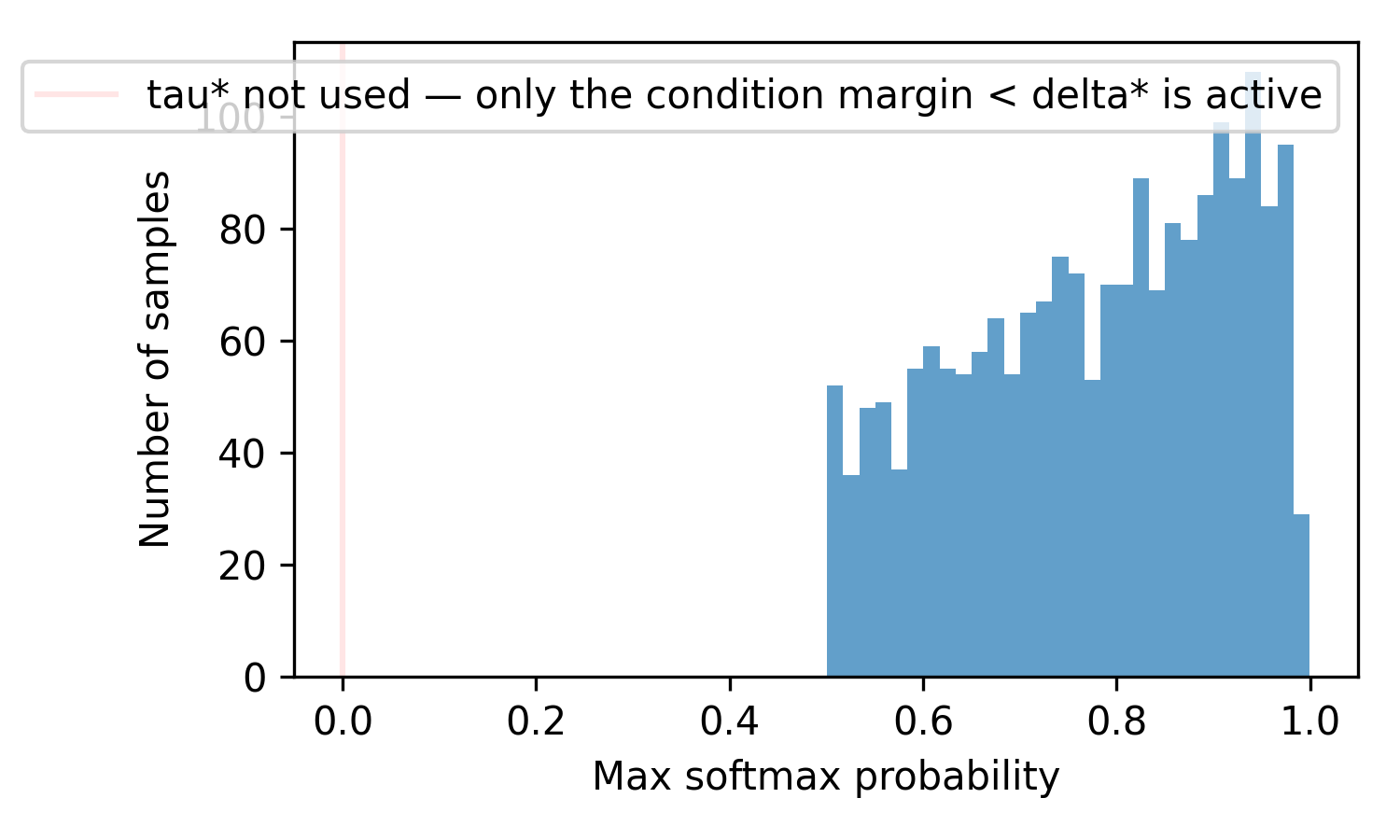}
    \caption{Max-softmax histogram ($\tau^\star{=}0$)}
  \end{subfigure}\hfill
  \begin{subfigure}{0.48\linewidth}
    \includegraphics[width=\linewidth]{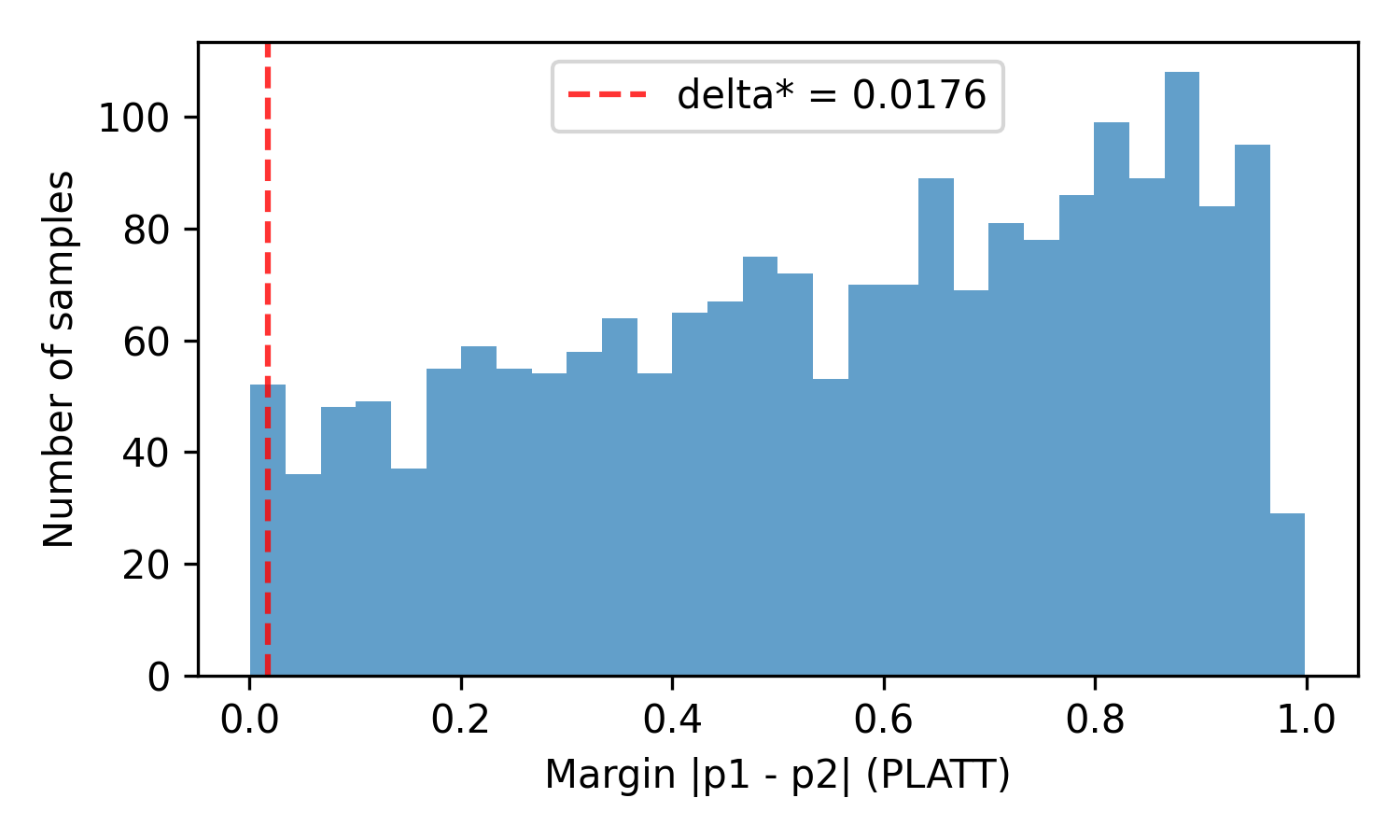}
    \caption{Margin histogram with $\delta^\star$}
  \end{subfigure}
  \caption{RadioML --- uncertainty distributions at the chosen operating point.}
  \label{fig:rml-uncertainty}
\end{figure}

\paragraph{Uncertainty profiles and selected anchors}
The histograms displayed in Figure~\ref{fig:rml-uncertainty} illustrate the max-softmax distribution and the top-2 margin, with a vertical line denoting $\delta^\star$.
Using $(\tau^\star,\delta^\star)$, the pipeline identifies \textbf{21} uncertain inputs (anchors) for in-depth analysis (refer to file \texttt{uncertain\_full.csv}).

\paragraph{Non-holomorphic sectors}
Within a local radius of $0.05$ surrounding each anchor, the modReLU kink detector recorded \emph{no} sector hits across all $21$ anchors (\texttt{kink\_summary.csv}, column \texttt{frac\_kink}$=0$ for each), indicating the locally smooth behavior of the complex subnetwork according to the RadioML preprocessing.

%--------------------------------------------------------
\subsection{Two case studies (fragile vs benign)}
\label{sec:results-radio-cases}

We demonstrate the local analysis using two representative anchors. Each panel presents a 2D contour of the surrogate decision function along with the robustness curves obtained by scanning phase-aligned rays on the original network (following the protocol outlined in \ref{app:impl}).

\begin{figure}[t]
  \centering
  \begin{subfigure}{0.48\linewidth}
    \includegraphics[width=\linewidth]{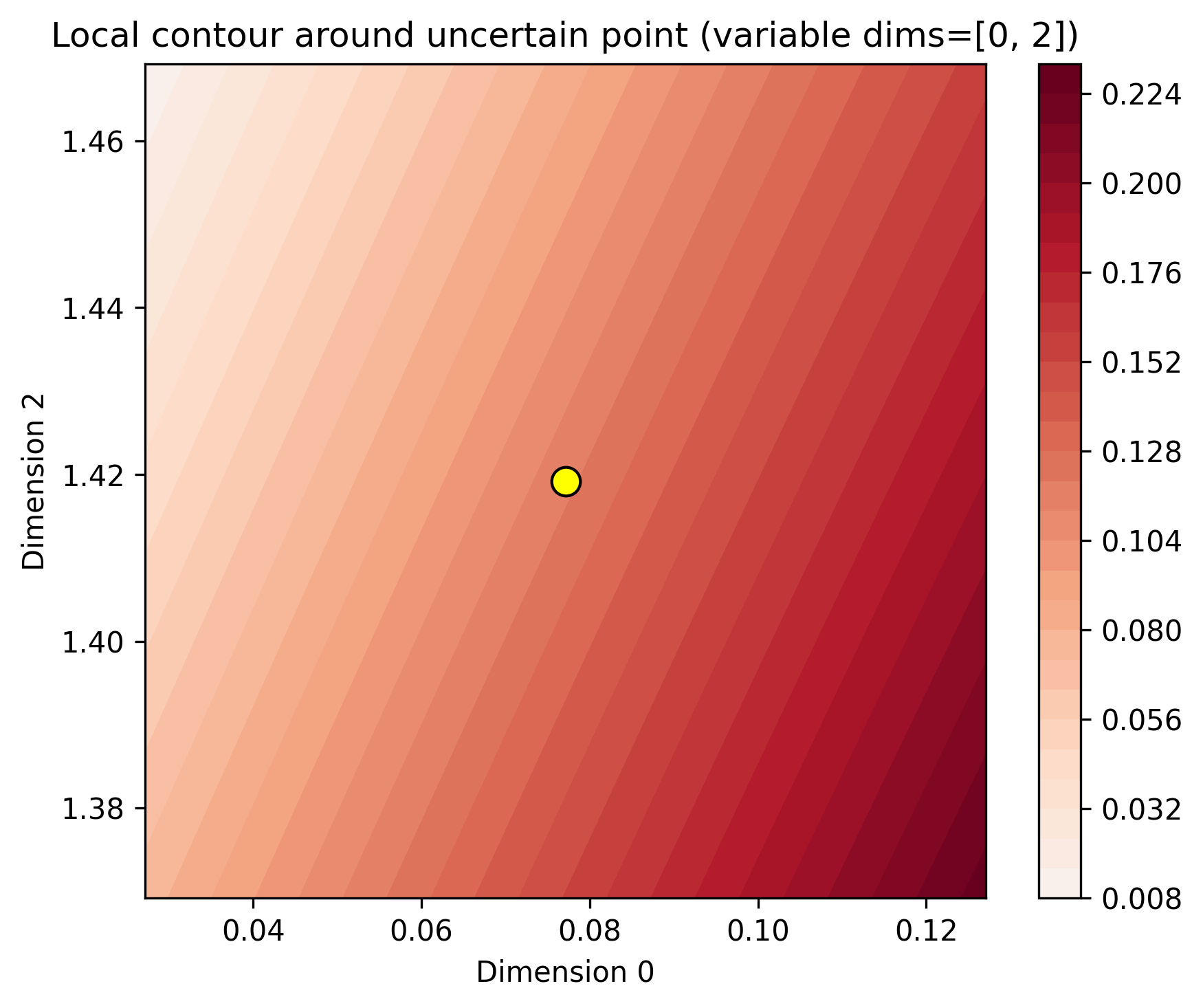}
    \caption{Local contour (anchor \#0)}
  \end{subfigure}\hfill
  \begin{subfigure}{0.48\linewidth}
    \includegraphics[width=\linewidth]{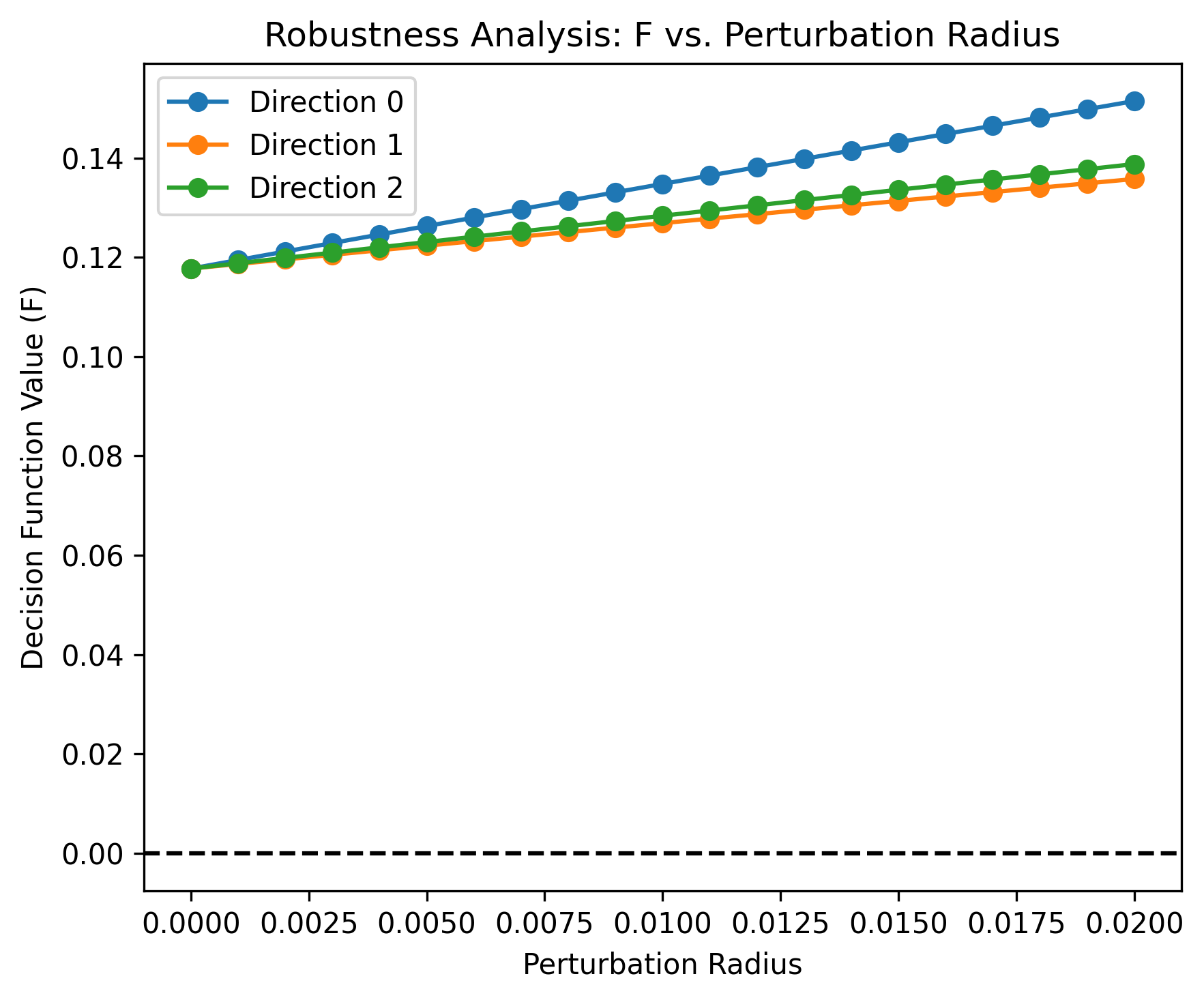}
    \caption{Robustness curves (anchor \#0)}
  \end{subfigure}
  \caption{Case \#0 (fragile). A Puiseux-guided direction shows an early class flip within the probed radius.}
  \label{fig:rml-case0}
\end{figure}

\begin{figure}[t]
  \centering
  \begin{subfigure}{0.48\linewidth}
    \includegraphics[width=\linewidth]{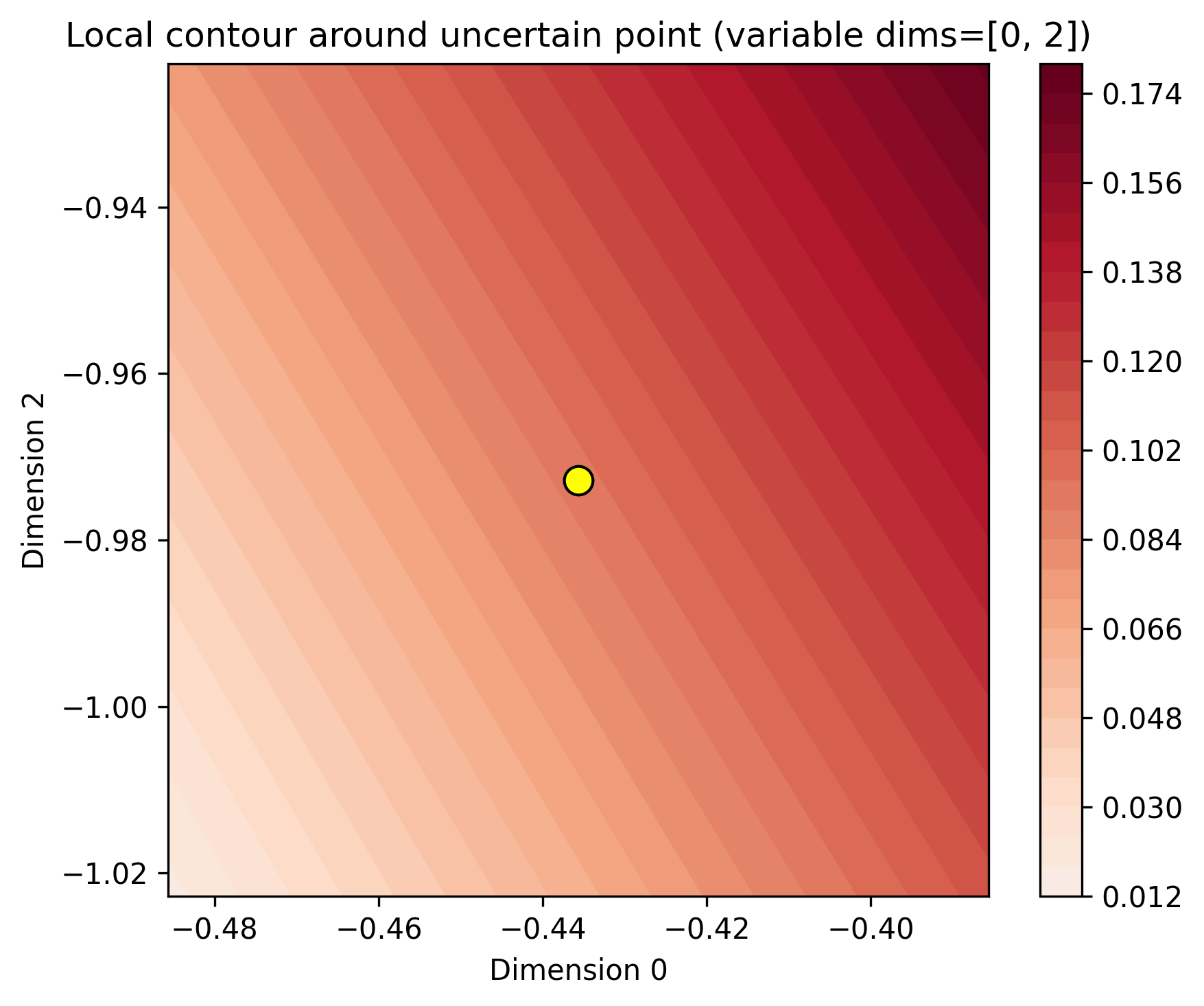}
    \caption{Local contour (anchor \#1)}
  \end{subfigure}\hfill
  \begin{subfigure}{0.48\linewidth}
    \includegraphics[width=\linewidth]{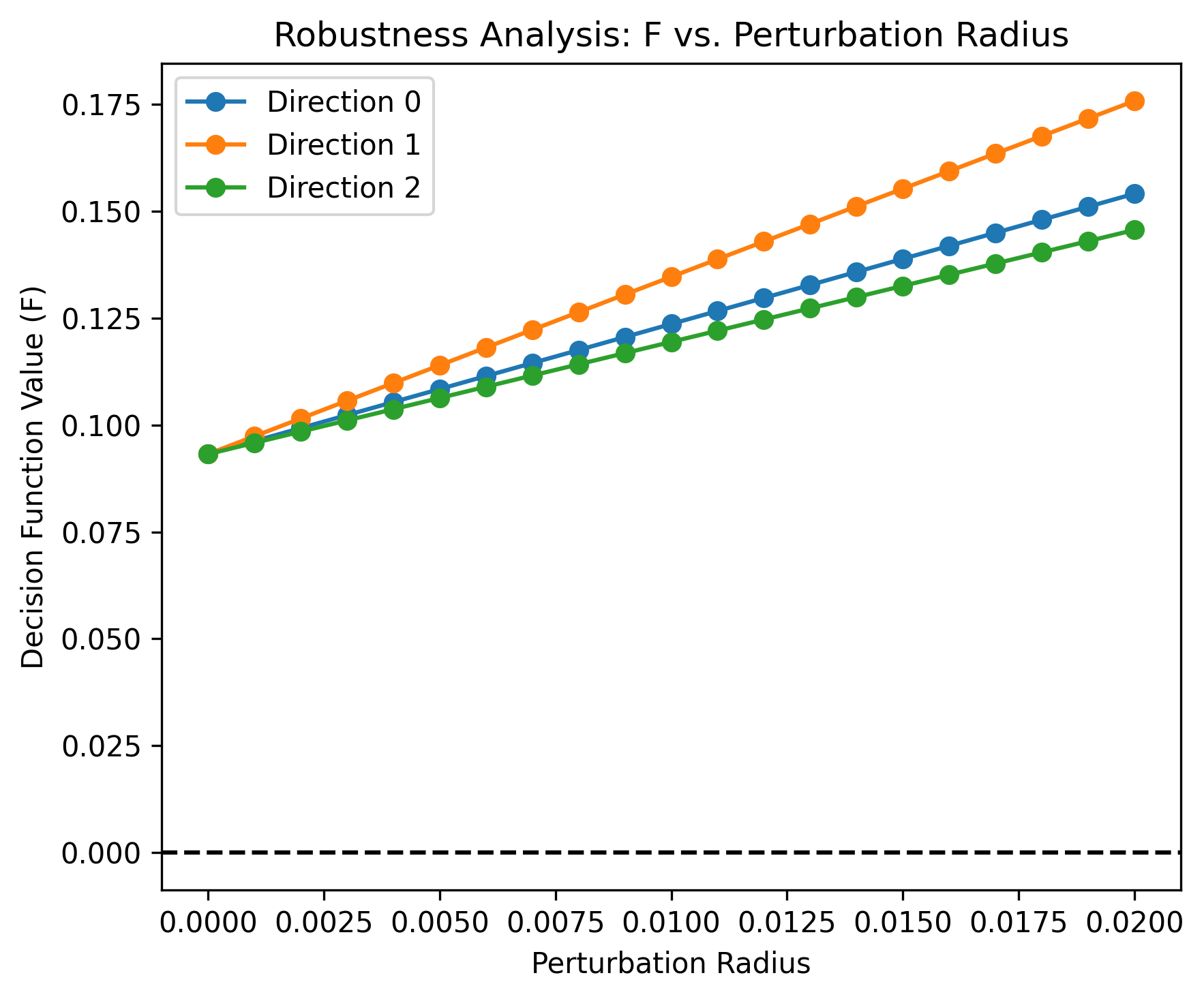}
    \caption{Robustness curves (anchor \#1)}
  \end{subfigure}
  \caption{Case \#1 (benign). No class flip is observed within the allocated scan budget.}
  \label{fig:rml-case1}
\end{figure}

%--------------------------------------------------------
\subsection{Ablations and Sensitivity (RadioML)}
\label{sec:rml-ablations}

\paragraph{Fidelity of the local surrogate}
Under the fixed configuration ($d{=}4$, $\Delta{=}0.05$, $600$ samples/anchor), the overall fidelity across $N{=}21$ anchors is:

\begin{table}[t]
  \centering
  \caption{Surrogate fidelity across $21$ anchors (degree $d{=}4$, radius $\Delta{=}0.05$, $600$ samples/anchor). Mean $\pm$ 95\% CI and median [IQR].}
  \label{tab:rml-surrogate-fidelity}
  \begin{tabular}{lcc}
    \toprule
    Metric & Mean $\pm$ 95\% CI & Median [IQR] \\
    \midrule
    RMSE & $0.149 \pm 0.010$ & $0.145$ [$0.130,\,0.160$] \\
    Sign agreement & $0.580 \pm 0.020$ & $0.577$ [$0.535,\,0.607$] \\
    $\mathrm{cond}(A)$ (LSQ) & --- & $5.55{\times}10^{2}$ [$5.39{\times}10^{2},\,5.71{\times}10^{2}$] \\
    \bottomrule
  \end{tabular}
\end{table}

The correlation between the LSQ condition number and RMSE is $\rho{=}0.20$, while the correlation with sign-agreement is $\rho{=}0.05$ (\texttt{local\_fit\_summary.csv}), indicating a weak association between conditioning and error in this setting.

\paragraph{Resource footprint}
The per-anchor Puiseux analysis, which includes sampling, weighted LSQ, symbolic factorization, simplification, Puiseux expansion, and directional scanning, averages $4.01\,$s $\pm 0.08\,$s. In comparison, a single gradient-saliency pass averages $5.05\,$ms $\pm 0.42\,$ms; peak GPU memory usage is approximately $64.8\,$MB in both evaluations (\texttt{resource\_point\{i\}.txt}). For the MIT--BIH dataset, this analysis is suitable for offline triage or batch processing.

\paragraph{Practical note}
In RadioML, we utilize $r_{\mathrm{dom}}$ \emph{for ranking} fragile anchors instead of interpreting it as an absolute radius; all candidate directions undergo validation through a brief ray sweep on the original network. This approach maintains the Newton--Puiseux value (including ordering and phase alignment) while preventing over-interpretation of the surrogate coefficients.

%--------------------------------------------------------
\subsection{Newton--Puiseux evidence (RadioML)}
\label{sec:rml-np-evidence}

\paragraph{Setup}
We implement the same post-processing techniques applied to the MIT--BIH dataset (using a local polynomial surrogate with $d{=}4$, $\Delta{=}0.05$, approximately 600 perturbations per anchor; Puiseux analysis; and robustness rays extending to a radius of $0.02$). The parser integrates \texttt{uncertain\_full.csv}, per-anchor benchmark TXT, and the dominant-ratio summary to calculate the predicted onset radius $r_{\mathrm{dom}}\!\approx\!\sqrt{|c_2|/|c_4|}$ and the observed minimum flip radius $r_{\mathrm{flip}}$ for each anchor.

\begin{table}[t]
  \centering
  \caption{Summary of Newton--Puiseux evidence from the parser in RadioML. Values are derived from all anchors exported by the uncertainty miner.}
  \label{tab:rml-np-summary}
  \begin{tabularx}{\linewidth}{lXXXXXX}
    \toprule
    Anchors & Flips $<$0.02 & $\overline{r_{\mathrm{dom}}}$ & $\overline{r_{\mathrm{flip}}}$ 
            & MAE $\bigl|r_{\mathrm{dom}}{-}r_{\mathrm{flip}}\bigr|$ 
            & Pearson & Spearman \\
    \midrule
    21 & 4 & 3.152 & 0.0163 & 1.810 & 0.608 & 0.400 \\
    \bottomrule
  \end{tabularx}
\end{table}

\paragraph{Interpretation}
In RadioML, the Puiseux dominant-ratio predictor shows a \emph{moderate} positive correlation with the observed flip radius (Pearson $\approx 0.61$), yet only \textbf{4/21} anchors exhibit flips within the scanning budget of $0.02$. The absolute scale exhibits an apparent discrepancy ($\overline{r_{\mathrm{dom}}}\gg \overline{r_{\mathrm{flip}}}$), indicating that $r_{\mathrm{dom}}$ should be perceived as a \emph{relative} fragility proxy instead of an absolute calibrated radius. The kink detector does not identify any events in the probed neighborhoods (panel~b), suggesting that deviations from proportionality in this dataset are unlikely due to non-holomorphic sectors.

\paragraph{Head-to-head fragile-anchor triage (RadioML)}
We also compare Newton--Puiseux descriptors against common XAI baselines for ranking
fragile anchors (positives = class flip within the scan budget $r\!\le\!0.02$).
On this corpus only $4/21$ anchors flip under the budget (baseline AUPRC $\approx 0.19$).
Table~\ref{tab:triage_radioml_auprc} reports area under the precision--recall curve
(AUPRC; higher is better) for five scores: the quartic magnitude $|c_4|$ from the Puiseux
surrogate, inverse flip radii along directions proposed by gradient/LIME/SHAP
($1/r_{\text{grad}}$, $1/r_{\text{LIME}}$, $1/r_{\text{SHAP}}$), and the gradient norm.
Consistently with the summary in Table~\ref{tab:rml-surrogate-fidelity}, the Puiseux coefficients carry ordering
information but do not dominate here; the gradient norm performs best, with SHAP-based
rays second. We therefore use $r_{\text{dom}}$ and $|c_4|$ chiefly as \emph{relative} fragility
signals on RadioML and always validate candidate directions by a short ray sweep on the
original network. Full PR curves are provided in \ref{app:fragileRadio}.

\begin{table}[t]
  \centering
  \caption{RadioML fragile-anchor triage (positives: flip within $r\!\le\!0.02$, $N{=}21$ anchors).
  AUPRC (higher is better); baseline (positive rate) $\approx 0.19$.}
  \label{tab:triage_radioml_auprc}
  \vspace{3pt}
  \begin{tabular}{lccccc}
    \toprule
    Score & $|c_4|$ & $1/r_{\text{grad}}$ & $1/r_{\text{LIME}}$ & $1/r_{\text{SHAP}}$ & grad\_norm \\
    \midrule
    AUPRC & 0.217 & 0.219 & 0.283 & 0.429 & \textbf{0.471} \\
    \bottomrule
  \end{tabular}
\end{table}

\paragraph{Contrast with MIT--BIH (for orientation)}
Utilizing the identical pipeline, MIT--BIH shows \textbf{14/17} flips within the budget and effectively triages by $|c_4|$ (precision/recall $\approx 0.79/0.79$), signifying that the Puiseux descriptors demonstrate higher predictive capability when the $\C^2$ representation aligns with the task geometry.%
\footnote{Refer to the full numbers in the MIT--BIH parser report; see Section~\ref{sec:results-mit}.}

\paragraph{Takeaway for RadioML}
The Newton--Puiseux analysis supplies consistent \emph{ordering} information (moderate $r$). However, it necessitates either (i) an expanded search budget and/or (ii) optimization of the surrogate specific to the dataset to achieve calibrated radii. Therefore, we primarily use $r_{\mathrm{dom}}$ on RadioML for \emph{ranking} fragile anchors and for phase-aligned directional probing, postponing absolute-radius claims to domains where the surrogate aligns more closely with the decision geometry.
%--------------------------------------------------------

%========================================================
\section{Comparisons and Statistical Tests (RadioML)} \label{sec:comparisons}

We present the results of calibration comparisons for RadioML, analyzing $\Delta$ECE against NONE, the win rates across folds, and the significance of the Wilcoxon test across the folds.

%--------------------------------------------------------
\subsection{Calibration methods head-to-head (RadioML)}
\label{sec:comparisons-calibration-rml}

We compare various calibration methods based on ECE, where lower values are preferable, report the relative change compared to NONE, and present the win-rate and Wilcoxon outcomes from the 10-fold cross-validation.

\begin{table}[t]
  \centering
  \caption{Head-to-head comparison of ECE for RadioML versus NONE (based on a 5-fold stratified approach: mean $\pm$ 95\% CI). 
  $\ Delta$ECE indicates the relative reduction compared to Uncalibrated; Wilcoxon $p$ is calculated using the 10-fold run with the alternative hypothesis $H_1\!:\,$method $<$ NONE.}
  \label{tab:rml-cal-head2head-ece}
  \begin{tabularx}{\linewidth}{lXXX}
    \toprule
    Method & ECE (mean $\pm$ 95\% CI) & $\Delta$ECE vs.\ NONE & Wilcoxon $p$ (10-fold) \\
    \midrule
    Uncalibrated (NONE) & $0.0334 \pm 0.0062$ & --- & --- \\
    Temperature         & $0.0335 \pm 0.0042$ & $-0.3\%$ & $1.00$ \\
    Platt               & $0.0283 \pm 0.0037$ & $15.3\%$ & $0.0527$ \\
    Isotonic            & $0.0285 \pm 0.0038$ & $14.9\%$ & $0.0186$ \\
    Beta                & \textbf{$0.0269 \pm 0.0036$} & \textbf{$19.6\%$} & $0.00488$ \\
    Vector              & $0.0283 \pm 0.0064$ & $15.3\%$ & $0.00488$ \\
    \bottomrule
  \end{tabularx}
\end{table}

\noindent \textbf{Win-rate vs.\ NONE (10-fold).}
Beta achieved \textbf{8/10}, Vector also \textbf{8/10}, Isotonic reached \textbf{7/10}, Platt obtained \textbf{6/10}, while Temperature recorded \textbf{0/10} wins.

\paragraph{Takeaways}
All non-parametric/vectorized calibrators enhance ECE performance compared to NONE on RadioML; the Beta calibrator achieves the lowest average ECE, demonstrating significant effectiveness ($p{=}\,0.0049$), along with the joint-best win rate (8/10). Vector scaling results in a similar reduction (15.3\%) and is also statistically significant ($p{=}\,0.0049$). Isotonic calibration shows significance ($p{=}\,0.0186$), while Platt calibration is marginally significant at the 5\% threshold ($p{=}\,0.0527$). In contrast, temperature scaling does not provide any improvement in this instance (with a learned $T$ near 1; $p{=}\,1.00$).

%========================================================

% End of section

%========================================================
\section{Discussion: Generalization, Limitations, and Scope}
\label{sec:discussion}

We summarize what elements transfer across different domains, discuss practical implications, outline limitations and validity threats, and conclude with guidance for design and future research.

\subsection{What generalizes and why}
Our analysis is based on two key ingredients that are prevalent in CVNN applications: 
(i) complex logits, whose moduli represent real-valued decision scores, and 
(ii) locally smooth behavior away from activation-sector boundaries (modReLU/zReLU). 
Given these conditions, a polynomial of small degree fitted to the \emph{logit difference} effectively captures the primary geometry of the decision sheet. The Newton--Puiseux factorization reveals a multisheeted structure, which is not modeled by gradients, LIME, or SHAP. The same analytical pipeline was successfully transferred from ECG to wireless I/Q (RadioML), where we primarily utilized Newton--Puiseux to rank fragile anchors and propose phase-aligned rays.

\subsection{Practical value}
The Newton--Puiseux descriptors yield two actionable results. First, the phases of the leading Puiseux coefficients indicate directions that can flip the class on the \emph{original} network within a small radius for most uncertain anchors (ECG), facilitating targeted review and stress testing. Second, the local \emph{branch multiplicity} serves as an indicator for phase-aware temperature adjustments, thereby enhancing calibration metrics without the need for retraining. Both results are inherently local and complement global post-hoc calibrators. We also quantify prioritization value against standard explainers:
precision--recall triage by $|c_4|$ is competitive but not dominant on MIT--BIH,
and near the prevalence baseline on RadioML; gradient-norm and SHAP-based rays
provide the strongest ranking signals. See Section~\ref{sec:np-evidence-mit} and Section~\ref{sec:rml-np-evidence} for AUPRC tables and
\ref{app:fragileMIT}, \ref{app:fragileRadio} for full PR curves.

\subsection{Limitations}
Newton--Puiseux is specifically a local tool. The absolute flip radii derived from surrogate coefficients are heuristic and require validation on the network through a short radial sweep. When piecewise-holomorphic sectors dominate the neighborhood (e.g., modReLU boundaries), the surrogate may be biased; our implementation mitigates this bias with kink-aware filtering and distance weights, although extreme cases can compromise fidelity. In terms of computational efficiency, per-anchor Newton--Puiseux analysis is significantly slower than a single saliency pass; hence, it is best used in an offline or batched approach.

\subsection{Threats to validity}
Three specific threats merit attention: (1) \textbf{Representation mismatch}: In the case of RadioML, a compact $\C^2$ featureization may not align perfectly with the task’s geometry, which could weaken the connection between dominant ratios and the observed flip radii. (2) \textbf{Coefficient scaling}: The monomial columns are normalized for conditioning; for geometric interpretations, unscaled coefficients are necessary; otherwise, the radius proxies may become poorly calibrated. (3) \textbf{Selection bias}: Anchors are derived from confidence/margin rules; varying abstention budgets may alter which regions are subjected to analysis.

\subsection{Design guidance}
It is advisable to maintain small neighborhoods and low degrees (\(d\!\in\!\{3,4\}\)) to enhance the stability of fits. Implement kink-aware exclusions and distance weights near modReLU/zReLU. Treat dominant-ratio radii as \emph{relative} fragility scores and always validate candidate directions on the original network. For calibration purposes, initiate with a conservative exponent in \(T’(m)=T_{\text{base}}\,m^{-\gamma}\) (\(\gamma\!\in\![0.5,1]\)) and assess sensitivity to potential multiplicity mis-estimation.

\subsection{Future work}
Two logical extensions for future research include: (i) developing a lightweight predictor for Newton--Puiseux descriptors utilizing local probes to eliminate the need for symbolic factorization during runtime, and (ii) integrating multiplicity-aware calibration with selective abstention strategies and domain-specific priors (e.g., SNR on RadioML).

%========================================================
\section{Conclusion}
\label{sec:conclusion}
%========================================================

We have introduced a Newton--Puiseux framework for the interpretation and calibration of complex-valued neural networks (CVNNs). By fitting a local polynomial surrogate around uncertain inputs and factorizing it into fractional-power branches, this method provides closed-form descriptors---branch multiplicities, phase-aligned curvature, and dominant Puiseux coefficients---that quantify robustness and over-confidence in ways that gradient- or perturbation-based explainers cannot.

\paragraph{Empirical summary}
On a controlled $\C^2$ helix (synthetic sanity check; \ref{app:synthetic_sanity}), the surrogate successfully mirrored the local decision geometry with a $\mathrm{RMSE}<0.09$, accurately identified the number of decision sheets, and utilized quartic Puiseux terms to predict adversarial flip radii with a precision of within $10^{-3}$ (synthetic sanity check; \ref{app:synthetic_sanity}). In the analysis of the MIT--BIH arrhythmia corpus, a phase-aware temperature adjusted according to branch statistics led to an improvement in expected calibration error when compared to the uncalibrated softmax ($T{=}1$); we report 95\% confidence intervals and conduct Wilcoxon signed-rank tests. During triage, only approximately $\sim\!1.3\%$ of regular beats were escalated for review, while fragile PVC cases were retained. Across various runs, the dominant Puiseux coefficients provided a significantly tighter estimate of the first flip radius compared to gradient-norm heuristics (with a median absolute error of approximately $7.5\times 10^{-4}$).

\paragraph{What the analysis adds}
Puiseux expansions reveal:
(i) \emph{where} the surface has folds (indicated by branch multiplicity),
(ii) \emph{how rapidly} it twists (denoted by quadratic/cubic/quartic magnitudes), and
(iii) \emph{in which phase directions} the risk accumulates (reflected by coefficient phases).
A straightforward operational gauge emerges from the surrogate’s coefficients, with an onset radius $r_{\mathrm{dom}}\!\approx\!\sqrt{|c_{2,0}|/|c_{4,0}|}$ that signifies the point at which quartic growth surpasses quadratic curvature.

\paragraph{Limitations}
The existing solver demonstrates robustness for $d\!\le\!2$ complex variables; however, scaling to higher dimensions $d$ will necessitate structured sparsity and advanced numerical algebra tools. The use of piecewise-holomorphic activations (for instance, zReLU) can introduce discontinuities that complicate polynomial fitting, and explicit noise modeling (including stochastic Puiseux series) remains an open area for exploration.

\paragraph{Outlook}
Future research directions include (i) extending the Newton--Puiseux analysis to higher-dimensional CVNNs utilizing sparse or low-rank surrogates, (ii) implementing branch-aware calibration that goes beyond scalar temperature scaling, (iii) establishing a tighter integration with tropical/convex geometry to understand global structure, and (iv) conducting domain studies in areas such as radar, coherent optics, and biomedical monitoring. The available code and pre-trained models enable complete reproducibility, motivating the community to rigorously test and extend the proposed methodology.

\medskip
\noindent\textbf{Takeaway.} By anchoring CVNN explainability in algebraic geometry, we address a practical gap between interpretability and calibration for phase-aware models, yielding actionable and quantitative indicators of local fragility that complement traditional XAI.

\section*{Acknowledgments}
\addcontentsline{toc}{section}{Acknowledgments}
The author acknowledges partial support from the grant “Singularities and Applications” – CF 132/31.07.2023 funded by the European Union – NextGenerationEU – through Romania's National Recovery and Resilience Plan.

%========================================================

\vspace{3cm}

\appendix
\section*{Appendices}
\addcontentsline{toc}{section}{Appendices} 

\section{Controlled synthetic $\C^{2}$ helix: data, model, and sanity check}
\label{app:synthetic_sanity}

\paragraph{Purpose}
This illustrative task isolates the geometry targeted by our Newton--Puiseux analysis, free from the complications of large models or noisy sensors. It serves solely as an educational validation; all significant claims are drawn from the MIT--BIH dataset.

% ---------- DATA ----------
\subsection{Data generation and rationale}
\phantomsection\label{app:data-synth}
We create a binary classification task using ordered pairs of complex numbers.
\(z=(x,y)\in\C^{2}\) stored in \(\R^{4}\) as
\([\,\Re x,\,\Re y,\,\Im x,\,\Im y\,]\)
(This ordering is consistently maintained in all equations and code).

For each class \(k\in\{0,1\}\), we independently sample amplitudes and phases according to
\begin{align}
  r_{1}^{(k)} &\sim\mathcal{N}\!\bigl(\mu_{r_1}^{(k)},(\sigma_{r_1}^{(k)})^{2}\bigr), &
  \varphi_{1}^{(k)} &\sim\mathcal{N}\!\bigl(\mu_{\varphi_1}^{(k)},\sigma_{\varphi}^{2}\bigr),\notag\\
  r_{2}^{(k)} &\sim\mathcal{N}\!\bigl(\mu_{r_2}^{(k)},(\sigma_{r_2}^{(k)})^{2}\bigr), &
  \varphi_{2}^{(k)} &\sim\mathcal{N}\!\bigl(\mu_{\varphi_2}^{(k)},\sigma_{\varphi}^{2}\bigr),
  \label{eq:synth-distr-app}
\end{align}
with fixed phase noise \(\sigma_{\varphi}=0.3\ \mathrm{rad}\).
We then set \(x=r_{1}^{(k)}e^{i\varphi_{1}^{(k)}}\) and \(y=r_{2}^{(k)}e^{i\varphi_{2}^{(k)}}\).
By default, we generate \(N=2000\) samples per class (seed~\texttt{42}) and split them into training and testing sets in an 80/20 ratio; the generator scripts and respective seeds are available in the repository.

\begin{table}[t]
  \centering
  \caption{Synthetic data parameters (unitless amplitudes).}\label{tab:synth-params-app}
  \begin{tabular}{@{}lcccccc@{}}
    \toprule
    & \multicolumn{2}{c}{Amplitude \(r_{1}\)} &
      \multicolumn{2}{c}{Amplitude \(r_{2}\)} &
      Phase \(\mu_{\varphi_1}\) & Phase \(\mu_{\varphi_2}\)\\
    \cmidrule(lr){2 - 3}\cmidrule(lr){4 - 5}
    Class & mean & std & mean & std & [rad] & [rad] \\\midrule
    0 & 1.0 & 0.10 & 2.0 & 0.20 & \(0\) & \(\tfrac{\pi}{4}\) \\
    1 & 1.5 & 0.15 & 2.0 & 0.20 & \(\tfrac{\pi}{2}\) & \(-\pi\) \\
    \bottomrule
  \end{tabular}
\end{table}

\paragraph{Design rationale}
Both classes occupy overlapping amplitude regions, yet they differ by \(90^{\circ}\) phase offsets, resulting in a \emph{helical} decision boundary within \(\R^{4}\).
The Bayes-optimal boundary can be represented analytically as:
\(|x|=|y|\) modulo \(\arg(x)-\arg(y)=\frac{\pi}{2}\),
providing a benchmark against which surrogate errors can be assessed.

% ---------- UNCERTAINTY MINING ----------
\subsection{Uncertainty-based query sampling (synthetic)}
Algorithm~\ref{alg:uncertain} designates a test point as \emph{uncertain} when \(|p_{1}-p_{2}|<\delta\) (with the default value of \(\delta{=}0.1\)); ties are also classified as uncertain.
On the held-out test set, this generally identifies approximately \(\lvert\mathcal U\rvert\approx 9.5\%\) of the instances.

\begin{algorithm}[t]
\caption{Uncertainty mining on synthetic test set}\label{alg:uncertain}
\begin{algorithmic}[1]
\Require{trained model $g$, data $(X,y)$, margin $\delta$}
\For{$i\gets1$ \textbf{to} $|X|$}
   \State $\boldsymbol\ell\leftarrow |g(X_{i})|$
   \State $\mathbf{p}\leftarrow\operatorname{softmax}(\boldsymbol\ell)$
   \State $\text{margin}\gets|p_{1}-p_{2}|$
   \If{$\text{margin}<\delta$}
        \State save $(i,X_{i},y_{i},\mathbf{p})$ to CSV
   \EndIf
\EndFor
\end{algorithmic}
\end{algorithm}

\begin{figure}[t]
  \centering
  \includegraphics[width=.55\linewidth]{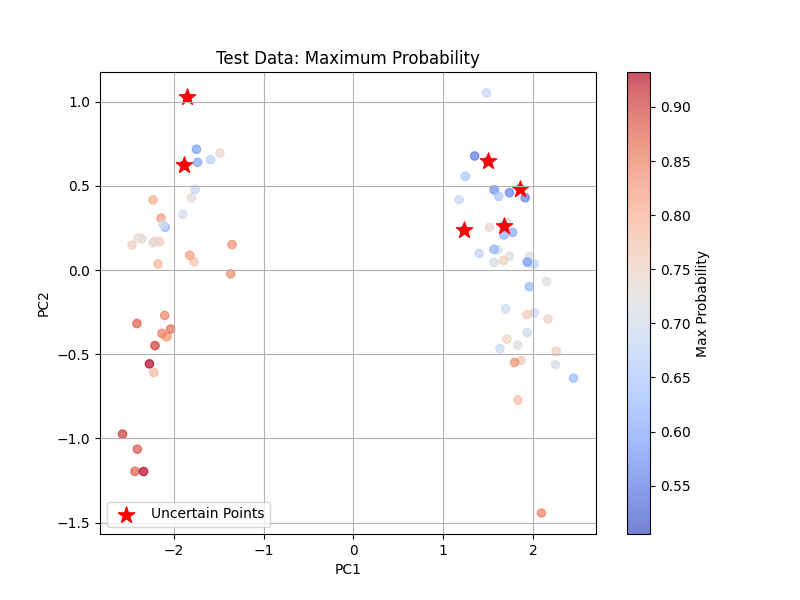}
  \caption{PCA representation of the synthetic test set, colored by maximum class probability.
         Stars indicate points flagged as uncertain (\(\delta=0.1\)) \citep{jolliffe2002principal}.}
  \label{fig:pca-uncertain-app}
\end{figure}

% ---------- MODEL ----------
\subsection{Model: \textbf{SimpleComplexNet} and training}
\phantomsection\label{app:arch-synth}
The synthetic task is situated in \(\C^{2}\;(\equiv\R^{4})\). To differentiate the impact of complex arithmetic from pure capacity, we utilize a compact network called \textbf{SimpleComplexNet}, which comprises \mbox{ComplexLinear--modReLU -- ComplexLinear} and \(H{=}16\) hidden units.

\paragraph{Complex linear projection (derivation)}
Consider \(\mathbf z=[z_{1},z_{2}]^{\!\top}\in\C^{2}\) where \(z_{j}=x_{j}+i\,y_{j}\).
For \(W=W_{r}+iW_{i}\in\C^{H\times 2}\),
\begin{equation}\label{eq:complex-gemm-app}
  \mathbf h
  = W\mathbf z
  = (W_{r}+iW_{i})(\mathbf x+i\mathbf y)
  = (W_{r}\mathbf x-W_{i}\mathbf y)
    + i\,(W_{r}\mathbf y+W_{i}\mathbf x),
\end{equation}
which translates to two real GEMMs supplemented by additions; this structure is equivariant to global phase rotations.

\paragraph{Activation and logits}
Each hidden unit \(h=x+iy\) undergoes the transformation through \(\operatorname{modReLU}(h;b)=0\) if \(|h|+b\le0\); otherwise, it transforms to \(\frac{|h|+b}{|h|}h\). The output layer of ComplexLinear generates scores \(c_k\in\C\), while the logits are computed as \(\ell_k=\sqrt{|c_k|^2+10^{-9}}\).

\begin{figure}[t]
  \centering
  % automatyczne dopasowanie szerokości do marginesu
  \resizebox{\linewidth}{!}{%
    \begin{tikzpicture}[
      node distance=1cm,
      every node/.style={font=\footnotesize,align=center},
      block/.style={
        draw,
        rounded rectangle,
        minimum width=1.5cm,
        minimum height=0.8cm,
        fill=blue!8
      },
      cyl/.style={
        draw,
        cylinder,
        cylinder uses custom fill,
        cylinder body fill=blue!8,
        shape border rotate=90,
        minimum width=1.5cm,
        minimum height=0.8cm
      },
      arrow/.style={-{Stealth[length=2.5mm]},line width=0.4pt}
    ]
      \node[block]   (in)   {Input\\$[x_1,y_1,x_2,y_2]$};
      \node[cyl, right=of in]   (l1)   {Complex\\Linear $W_1$};
      \node[block, right=of l1] (m1)   {modReLU};
      \node[cyl, right=of m1]   (l2)   {Complex\\Linear $W_2$};
      \node[block, right=of l2] (m2)   {$|\!\cdot\!|$};
      \node[block, right=of m2] (sm)   {Softmax};

      \draw[arrow] (in) -- (l1) -- (m1) -- (l2) -- (m2) -- (sm);
    \end{tikzpicture}%
  }
\caption{SimpleComplexNet ($H{=}16$ hidden units). 
All tensors are represented as concatenated real and imaginary parts.}
  \label{fig:tikz-synth-net-app}
\end{figure}

\paragraph{Training protocol}
The model is trained for 50 epochs using the Adam optimizer \((\eta=10^{-3},\beta_1=0.9,\beta_2=0.999,\varepsilon=10^{-8})\), with a batch size of 64, a weight decay of \(10^{-4}\), no learning rate schedule, and early stopping was not activated. Across five different random splits (seed = 42), the accuracy achieved is \(92.7\%\!\pm\!0.6\%\).

% ---------- SURROGATE & NP ----------
\subsection{Local surrogate and Puiseux decomposition}
For each flagged anchor, we fit a degree-4 bivariate polynomial to the logit difference (excluding constant and linear terms) using 200 Latin--Hypercube perturbations within a \(\pm 0.05\) box \citep{mckay2000comparison}. The surrogate is then factorized using the Newton--Puiseux solver to extract fractional-power branches and their dominant coefficients.

% ---------- RESULTS ----------
\subsection{Results and take-aways}
Across the five anchors, the surrogates yielded low absolute errors and consistent decision signs on held-out perturbations; each anchor displayed \emph{exactly two} Puiseux branches. Only the case with the highest curvature showed flips within the tested radius (\(r\approx 5{\times}10^{-3}\)); the others maintained stability up to \(r=0.02\).
Per-anchor metrics (from the original run) reveal:
\[
\text{RMSE}\in\{0.056,~0.122,~0.123,~0.137,~0.183\},\quad
\]
\[
\text{MAE}\in\{0.046,~0.104,~0.106,~0.117,~0.167\},
\]
with sign agreement ranging from \(0.575\) to \(0.685\), “\#\,branches” equating to \(=2\) for all, and two flips observed in the most curved anchor.
The quartic Puiseux magnitudes align with the observed flip radii, whereas LIME/SHAP attributions do not account for this higher-order behavior.

\paragraph{Why coefficients matter}
In the synthetic benchmark, quartic Puiseux coefficients (\(|c_4|\)) provide a better prediction of local fragility than attribution scores; the dominant-ratio heuristic \(r_{\mathrm{dom}}\!\approx\!\sqrt{|c_2|/|c_4|}\) corresponds well with the observed flip radius within a factor of \({\times}3\), indicating that points with a larger \(|c_4|\) tend to flip sooner.

\paragraph{Take-away}
The synthetic task validates that (i) Puiseux factorization accurately describes the two-sheet topology and (ii) dominant higher-order coefficients effectively track local fragility. We position this benchmark as a sanity check; all significant conclusions are derived from the real-world ECG study.

\section{Newton--Puiseux: algorithmic details and worked examples}
\label{app:newton}
In this section, we will outline the Newton--Puiseux algorithm for finding the roots of the equation $f(x,y)=0$. It is worth noting that the presentation of this method lacks complete mathematical rigor; instead, it is designed to be understandable to a non-mathematical audience. For those seeking a comprehensive understanding of the theory surrounding the Newton polygon and Puiseux expansions, we recommend referencing classical texts \citep{wall2004singular,brieskorn1986plane, walker1950algebraic}.

\medskip

Let us first focus on the complex function of two variables $f(x,y)$, which we will consider as a formal power series. However, for simplicity, we will treat $f(x,y)$ as a complex polynomial with two variables. Thus, we can represent it as:
$$
f(x,y)=\sum_{i,j} a_{ij}x^iy^j,
$$
where $a_{ij}\in\mathbb{C} \backslash \{0\}$. When examining the equation $f(x,y)=0$,  the solutions form a complex curve.

\medskip

In general, visualizing a complex curve can be challenging, which is why we will examine the real curve $x^2-y^3=0$ to build some intuition. As shown in Figure \ref{fig:1}, this curve exists in the real plane. Let $(x_0,y_0)$ represent an arbitrary point on this curve, and let $U$ and $V$ be neighborhoods around the points $x_0$ and $y_0$, respectively (noting that $U,V\subset \mathbb{R}$). Our objective is to identify a function $\phi:U\to V$ (or $\psi:V\to U$) such that $f(x,\phi(x))=0$ (or $f(\psi(y),y)=0$). In simpler terms, we want to express the curve $x^2-y^3=0$ in the vicinity of the point $(x_0,y_0)$ as a graph of a function that depends on a single variable. If we select the point $(x_0,y_0)$ such that the curve is smooth at this location (which means the derivatives $(f’_x(x_0,y_0),f’_y(x_0,y_0))\neq(0,0)$), the Implicit Function Theorem applies. Specifically, this theorem asserts that if a function \(f(x,y)\) is continuously differentiable and $(f’_x(x_0,y_0),f’_y(x_0,y_0))\neq(0,0)$, it is possible to locally represent \( y \) as a function of \( x \) or \( x \) as a function of \( y \). However, the theorem does not provide a method for efficiently determining these functions. In straightforward scenarios, we can achieve this through specific transformations. For instance, considering the point \((1,1)\), we first calculate the partial derivatives:
\[
(f’_x(x,y), f’_y(x,y))= (2x, -3y^2), \quad(f’_x(1,1), f’_y(1,1))=(2,-3)\neq(0,0).
\]
Consequently, the Implicit Function Theorem ensures that there exist functions \( y = \phi(x) \) in the neighborhood of \( x = 1 \) and \( x=\psi(y) \) around \( y=1 \) such that \(f(x,\phi(x))=0\) and \(f(\psi(y),y)=0\). Here, we can directly ascertain the explicit forms of these functions: \(\phi(x)=x^{\frac{2}{3}}\) and \(\psi(y)=y^{\frac{3}{2}}\). 
On the other hand, this theorem does not address scenarios where the point \((x_0,y_0)\) is singular, meaning that \((f’_x(x_0,y_0),f’_y(x_0,y_0))=(0,0)\). This limitation arises because it is impossible to represent the curve as a graph of a single function in the vicinity of such a point. In the case of our curve \(x^2-y^3=0\), the singular point is \((0,0)\) (since \((f’_x(0,0),f’_y(0,0))=(2\cdot 0,-3\cdot 0^2)=(0,0)\)). As indicated by Figure \ref{fig:1}, it is evident that near the point \((0,0)\), the curve can be depicted as the graph of two functions, namely \(\psi_1(y)=-y^{\frac{3}{2}}\) and \(\psi_2(y)=y^{\frac{3}{2}}\).

\medskip

The Newton-Puiseux method is developed to identify such functions, which provide a local graphical representation of a given curve. This method also accommodates singular points, where both partial derivatives vanish.

\begin{figure}[ht]
    \centering
    \begin{tikzpicture}[scale=0.8]
        \begin{axis}[
            axis lines = middle,
            xlabel = {$x$},
            ylabel = {$y$},
            samples = 1000,
            domain = -1.5:1.5,
            xmin=-1.5, xmax=1.5,
            ymin=-1.5, ymax=1.5,
            xtick={-1,1},
            legend pos=south west
        ]
        
        % Curve x^2 = y^3 (i.e., y = (x^2)^(1/3))
        \addplot[blue, thick, domain=-1.5:1.5] {abs(x)^(2/3)};

        % Tangent line at (1,1): y = (2/3)x + 1/3

        % Point (1,1)
        \addplot[mark=*, red] coordinates {(1,1)};
        \node[anchor=south] at (axis cs:1,1) {$(1,1)$};
         \addplot[mark=*, red] coordinates {(0,0)};
        \node[anchor=north west] at (axis cs:0,0) {$(0,0)$};
        
        \end{axis}
    \end{tikzpicture}
    \caption{Curve defined by $x^2 - y^3 = 0$ with its tangent at $(1,1)$.}\label{fig:1}
\end{figure}

After this brief discussion on visualization, we now shift our focus back to the complex scenario, that is, when we assume that $f(x,y)$ is a complex polynomial. We will begin by defining some essential concepts.

\medskip

\noindent \textbf{Newton Polygon.} The support of the polynomial $f(x,y)=\sum_{i,j} a_{ij}x^iy^j$ is defined as the set $\Supp(f):=\{(i,j)\in \mathbb{R}^2|a_{ij}\neq 0 \}$. To illustrate this concept, consider the polynomial $f(x,y)=2x^4+xy^2-y^5+x^3y^3$, for which we find that $\Supp(f)=\{(4,0),(1,2),(0,5),(3,3)\}$.

The Newton diagram $\Delta(f)$ is characterized as the smallest convex set that contains $\Supp(f)+\mathbb{R}_+^2$. The Newton polygon $\mathcal{NP}(f)$ refers to the set of compact edges on the boundary of the diagram $\Delta(f)$. To clarify these concepts further, let us examine our example once more. For the polynomial $f(x,y)=2x^4+xy^2-y^5+x^3y^3$, Figure \ref{fig:2} (left) displays $\Supp(f)+\mathbb{R}_+^2$, while Figure \ref{fig:2} (right) illustrates the Newton diagram (highlighted in blue) and the Newton Polygon (outlined in red).

\begin{figure}[ht]
\centering
\begin{minipage}{0.45\textwidth}
\begin{tikzpicture}[scale=0.7]
  \draw[step=1cm, gray, very thin] (-0.5,-0.5) grid (6.5,7.5);
  
  \draw[->] (-0.5,0) -- (7,0) node[right] {$x$};
  \draw[->] (0,-0.5) -- (0,8) node[above] {$y$};

  \fill[gray!30, opacity=0.5] (4,0) rectangle (6.5,7.5);
  \fill[gray!30, opacity=0.5] (1,2) rectangle (6.5,7.5);
  \fill[gray!30, opacity=0.5] (0,5) rectangle (6.5,7.5);
  \fill[gray!30, opacity=0.5] (3,3) rectangle (6.5,7.5);

  \draw[dashed] (4,0) -- (4,7.5);
  \draw[dashed] (4,0) -- (6.5,0);
  
  \draw[dashed] (1,2) -- (1,7.5);
  \draw[dashed] (1,2) -- (6.5,2);
  
  \draw[dashed] (0,5) -- (0,7.5);
  \draw[dashed] (0,5) -- (6.5,5);
  
  \draw[dashed] (3,3) -- (3,7.5);
  \draw[dashed] (3,3) -- (6.5,3);

  \fill (4,0) circle (2pt) node[below right] {$(4,0)$};
  \fill (1,2) circle (2pt) node[above left] {$(1,2)$};
  \fill (0,5) circle (2pt) node[above right] {$(0,5)$};
  \fill (3,3) circle (2pt) node[above right] {$(3,3)$};
  \end{tikzpicture}
\end{minipage}
\hfill
\begin{minipage}{0.45\textwidth}
\begin{tikzpicture}[scale=0.7]
  
  \draw[step=1cm, gray, very thin] (-0.5,-0.5) grid (6.5,7.5);
  
  \draw[->] (-0.5,0) -- (7,0) node[right] {$x$};
  \draw[->] (0,-0.5) -- (0,8) node[above] {$y$};

 \fill[blue!20, opacity=0.5] 
    (6.5,0) -- (6.5,7.5) -- (0,7.5) -- (0,5) -- (1,2) -- (4,0) -- cycle;
  
  \draw[very thick, red] (4,0) -- (1,2) -- (0,5);
  \fill (4,0) circle (2pt) node[below right] {$(4,0)$};
  \fill (1,2) circle (2pt) node[above left] {$(1,2)$};
  \fill (0,5) circle (2pt) node[above right] {$(0,5)$};
  \fill (3,3) circle (2pt) node[above right] {$(3,3)$};
 \end{tikzpicture}
\end{minipage}

\caption{For $f(x,y)=2x^4+xy^2-y^5+x^3y^3$, on the left $\Supp(f)+\mathbb{R}_+^2$, on the right $\Delta(f)$ (marked in blue) and $\mathcal{NP}(f)$ (marked in red).}\label{fig:2}
\end{figure}

\noindent
\textbf{Quasi-homogeneous polynomials in two variables.} A polynomial
\[
g(x,y) \;=\; \sum_{i,j} b_{i,j}\,x^i y^j
\]
is termed \emph{quasi-homogeneous of type $(p,q;d)$} if there exist positive integers $p,q$ and an integer $d$ such that for every monomial $x^i y^j$ with a nonzero coefficient $b_{i,j}$, the condition
\[
p\,i \;+\; q\,j \;=\; d.
\]
is satisfied. In simpler terms, all nonzero monomials in $g$ lie along the line described by $p\,i + q\,j = d$ within the integer lattice.

\medskip

An important connection to the Newton polygon becomes apparent in the fact that each compact edge $E$ of $\mathcal{NP}(f)$ possesses a unique \emph{coslope}, calculated as $\frac{q}{p}$. Moreover, the integer vector $(p,q)$ is perpendicular to that edge. Specifically, if an edge $E$ can be mathematically represented by the equation $p\,i + q\,j = d$ for some integers $p,q,d$, then the aggregated sum of the monomials $a_{ij}x^i y^j$ of $f$ for which $(i,j)$ resides on $E$ forms a quasi-homogeneous polynomial of type $(p,q;d)$. We denote this polynomial as
\[
f(x,y;E)=\sum_{(i,j)\in E}a_{ij}x^iy^j.
\]
Thus, \emph{each edge of the Newton polygon gives rise to a quasi-homogeneous polynomial explicitly determined by the coslope $(p,q)$ of that edge and the corresponding integer $d$}. This relationship plays a crucial role in understanding and factoring the local structure of the curve $f(x,y)=0$ in the subsequent steps of the Newton-Puiseux algorithm.

\medskip

\noindent
\textbf{Factorization of a quasi-homogeneous polynomial.} A significant characteristic of such polynomials is that, by applying an appropriate substitution that respects the weights $(p,q)$, the problem can be transformed into factoring a homogeneous polynomial in a single variable. Specifically, let $g$ be a quasi-homogeneous polynomial of type $(p,q;d)$. We can express $g$ in the form
\[
g(x,y)\;=\; \sum_{i,j} b_{ij}\,x^i\,y^j, 
\quad\text{where } p\,i+q\,j = d \text{ whenever } b_{ij}\neq 0.
\]
For example, consider the substitution
\[
x = t^p, \quad y = t^q \cdot u,
\]
which transforms each monomial $x^i\,y^j$ into $t^{p\,i} \cdot \bigl(t^q\,u\bigr)^j = t^{p\,i + q\,j} \,u^j.$ Since $p\,i + q\,j = d$ for all terms with $b_{ij}\neq 0$, it follows that each nonzero term contributes a total degree $d$ in $t$. Therefore, we can write
\[
g\bigl(t^p, \,t^q u\bigr) \;=\; t^d\, \tilde{g}(u),
\]
where $\tilde{g}(u)$ is a polynomial in the single variable $u$. This means that we have represented $g(x,y)$ through a univariate polynomial $\tilde{g}(u)$ (up to the factor $t^d$). Consequently, factoring $g$ over $\mathbb{C}$ equates to factoring $\tilde{g}(u)$ over $\mathbb{C}$, which is often a much simpler problem. This methodology plays a crucial role in understanding how quasi-homogeneous components of $f$, derived from its edges in $\mathcal{NP}(f)$, can be factorized using the Newton-Puiseux algorithm.

\medskip

\begin{figure}[ht]
\centering
\begin{tikzpicture}[scale=0.7]

  % grid and axes
  \draw[step=1cm, gray, very thin] (-0.5,-0.5) grid (6.5,7.5);
  \draw[->] (-0.5,0) -- (7,0) node[right] {$x$};
  \draw[->] (0,-0.5) -- (0,8) node[above] {$y$};

  % fill the Newton diagram region
  \fill[blue!20, opacity=0.5]
    (6.5,0) -- (6.5,7.5) -- (0,7.5) -- (0,5) -- (1,2) -- (4,0) -- cycle;

  % draw the edges of the polygon in red
  \draw[very thick, red] (4,0) -- (1,2) -- (0,5);

  % mark the points of Supp(f)
  \fill (4,0) circle (2pt) node[below right] {$(4,0)$};
  \fill (1,2) circle (2pt) node[above left]  {$(1,2)$};
  \fill (0,5) circle (2pt) node[above right] {$(0,5)$};
  \fill (3,3) circle (2pt) node[above right] {$(3,3)$};

  % first edge normal vector (2,3)
  \draw[->, thick, blue] (2.5,1) -- ++(0.7*2, 0.7*3)
    node[midway, below right]{\tiny $(2,3)$};

  % second edge normal vector (3,1)
  \draw[->, thick, blue] (0.5,3.5) -- ++(0.4*3, 0.4*1)
    node[midway, above]{\tiny $(3,1)$};

\end{tikzpicture}
\caption{Newton polygon of $f(x,y)=2x^4 + x\,y^2 - y^5 + x^3y^3$ with edges in red and normal vectors in blue.}
\label{fig:3}
\end{figure}

\medskip

\noindent \textbf{Example.} Let us examine again \(f(x,y)=2x^4 + xy^2 - y^5 + x^3y^3.\)
According to the definition of the Newton polygon \(\mathcal{NP}(f)\), the set of integer points contributing to \(f\) is given by \( \Supp(f)=\{(4,0), (1,2), (0,5), (3,3)\} \). Figure \ref{fig:3} below illustrates the Newton diagram \(\Delta(f)\) along with its two compact edges, which are highlighted in red. Additionally, the corresponding normal (coslope) vectors \((p,q)\) for each edge are marked in blue.

\medskip

\noindent
Quasi-homogeneous polynomials derived from the edges.
\begin{itemize}
\item The first edge (connecting \((4,0)\) and \((1,2)\)) is supported by the points \((4,0)\) and \((1,2)\). A normal vector to this edge is \((2,3)\). In fact, these points satisfy the equation \(2i + 3j = 8\). Thus, the \emph{quasi-homogeneous} polynomial associated with this edge is \(2x^4 + xy^2\), and its type is \((p,q;d)=(2,3;8)\). By applying the substitution \(x=t^2\), \(y=t^3\,u\), we obtain
\[
x^4 = t^8,\quad xy^2 = t^2(t^3u)^2 = t^8u^2,\quad g_1\bigl(t^2,t^3u\bigr)=t^8(2+u^2).
\]
Thus \(\tilde{g}_1(u)=2+u^2\) factors (over \(\mathbb{C}\)) as \((u-i\sqrt{2})(u+i\sqrt{2})\). Translating back, \(u = \dfrac{y}{t^3}\) and \(t^2 = x\) \(\Rightarrow\) \(t = x^{1/2}\). Thus \( u =\frac{y}{(x^{1/2})^3} = \frac{y}{x^{3/2}}\). Consequently, we can write 
\[
\begin{aligned}
g_1(x,y) &= 2x^4 + xy^2=x^4 \Bigl(\frac{y}{x^{3/2}}- i\sqrt{2}\Bigr)\Bigl(\frac{y}{x^{3/2}}+ i\sqrt{2}\Bigr)\\
&= x \Bigl(y- i\sqrt{2}x^{3/2}\Bigr)\Bigl(y+ i\sqrt{2}x^{3/2}\Bigr).
\end{aligned}
\]

\item The second edge (connecting \((1,2)\) and \((0,5)\)) is supported by the points \((1,2)\) and \((0,5)\). A normal vector to this edge is \((3,1)\). The exponents along this line satisfy \(3\,i \;+\; j \;=\; 5\). Therefore, the quasi-homogeneous polynomial corresponding to this edge is \(xy^2 - y^5\), with its type being \((p,q;d)=(3,1;5)\). Using the substitution \(x=t^3\), \(y=t\,u\), each term becomes degree 5 in \(t\):
\[
x\,y^2=t^3(tu)^2=t^5u^2,\quad y^5=(tu)^5=t^5u^5,
\]
\[
g_2\bigl(t^3,tu\bigr)=t^5\bigl(u^2-u^5\bigr)=t^5u^2(1-u^3).
\]
Hence \(\tilde{g}_2(u)=u^2(1-u^3)\) factors further as \(-u^2(u-1) \Bigl(u-\frac{-1-i\sqrt{3}}{2}\Bigr)\Bigl(u-\frac{-1+i\sqrt{3}}{2}\Bigr)\). Back - substituting \(u = \frac{y}{t}\), with \(t^3=x\)\(\Rightarrow\)\(t = x^{1/3}\), we get \( u \;=\;\frac{y}{x^{1/3}}\) and thus
\[
\begin{aligned}
g_2(x,y)&=xy^2 - y^5\\ &=
-xy^2\Bigl(\frac{y}{x^{1/3}}-1\Bigr)\Bigl(\frac{y}{x^{1/3}}- \frac{-1-i\sqrt{3}}{2} \Bigr)\Bigl(\frac{y}{x^{1/3}}- \frac{-1+i\sqrt{3}}{2} \Bigr)\\
&=-y^2\Bigl(y-x^{1/3}\Bigr)\Bigl(y- \frac{-1-i\sqrt{3}}{2} x^{1/3} \Bigr)\Bigl(y- \frac{-1+i\sqrt{3}}{2}x^{1/3} \Bigr).
\end{aligned}
\]

\end{itemize}

\medskip

\noindent
\textbf{Puiseux expansions.} A Puiseux expansion (or Puiseux series) for a solution \(y(x)\) to the equation \(f\bigl(x,y\bigr)=0\) takes the form of a formal power series in fractional powers of \(x\). Specifically, around \(x=0\), we can express it as
\[
y(x)
\;=\;
\sum_{m=m_0}^{\infty} \alpha_m\, x^{\frac{m}{N}},
\]
where \(\alpha_m \in \mathbb{C}\) and \(N\) is a positive integer. In this formulation, the exponents of \(y(x)\) can be rational and do not necessarily have to be integer multiples. Such expansions naturally arise when exploring the \emph{local} (particularly singular) behavior of algebraic curves, and the Newton--Puiseux algorithm offers a structured approach to compute them term by term.

\medskip

\noindent
\textbf{Newton - Puiseux Theorem.} Before delving into the Newton--Puiseux algorithm, we present a crucial result that elucidates how local expansions of the form \(y(x) = ax^\theta + \hot\) arise from the examination of the quasihomogeneous components of \(f\). Specifically, once \(ax^\theta\) is identified as a root of the polynomial restricted to a certain edge of \(\mathcal{NP}(f)\), it can be extended to a comprehensive Puiseux expansion for the polynomial \(f\) itself.

\noindent \textbf{Theorem} (Newton - Puiseux)
emph{Let \(f\) be a complex polynomial in two variables, let \(E\) represent an edge of \(\mathcal{NP}(f)\) with co-slope \(\theta\), and assume \(a \in \mathbb{C}\setminus\{0\}\). Then \(ax^\theta\) serves as a root of the quasi-homogeneous polynomial \(f(x,y;E)\) (i.e., \(f(x,ax^\theta;E)=0\)) if and only if there exists a function \(y(x)\) that satisfies \(f\bigl(x,y(x)\bigr)=0\) and has its leading term as \(ax^\theta\), meaning \(y(x) = ax^\theta + \hot\).
}

\medskip

\noindent
\textbf{Remark.} In practical terms, each edge \(E\) of \(\mathcal{NP}(f)\) yields a specific number of local solutions \(y(x)\) for the equation \(f(x,y)=0\), corresponding to the distinct roots into which the associated quasi-homogeneous polynomial \(f(x,y;E)\) factors upon univariate reduction. Thus, when accounting for \emph{all} such edges, each one contributes a certain number of Puiseux expansions of the type \(y(x)=a\,x^\theta + \hot\). In many cases, the overall count of these expansions can be determined by examining the vertex of \(\mathcal{NP}(f)\) (the Newton diagram) with the highest ``vertical'' coordinate, as this vertex essentially sets a boundary on the total possible degree in \(y\).

\smallskip

\noindent
Moreover, if \(f(x,y)\) is divisible by \(y\), it follows that \(y(x)=0\) automatically provides a solution for \(f(x,y)=0\). In simpler terms, whenever \(f(x,y)=y\cdot g(x,y)\), the curve defined by \(f=0\) necessarily includes the line \(y=0\).

\smallskip

\noindent
Indeed, we can switch the roles of \(x\) and \(y\) and study the expansions \(x(y)\) in an entirely analogous manner: each edge of the Newton polygon projected along the opposite axis produces expansions that describe the solutions locally as functions of \(y\). In summary, the Newton-Puiseux approach enables us to track all local branches of the curve \(f(x,y)=0\) (around a singular point) by systematically addressing the contributions from each edge in the polygon.

\medskip

\noindent
\textbf{Newton - Puiseux Algorithm.} 
After discussing how to construct and interpret the Newton polygon of \(f(x,y)\) and the significance of its edges in generating local expansions for solutions \(y(x)\), we now outline the principal iterative steps of the Newton-Puiseux algorithm. The objective is to build the Puiseux series incrementally, refining an initial approximation \(y(x)=a_{1}x^{\theta_1}\) by adding higher-order terms until the needed degree of accuracy (or complete factorization) is achieved.

\smallskip
\noindent
\textbf{Step 1.} Select an edge \(E_1\) of \(\mathcal{NP}(f)\). Let its co-slope be denoted by \(\theta_1\). 

\noindent
\textbf{Step 2.} From this edge, identify all nonzero solutions for the corresponding quasi-homogeneous polynomial \(f(x,y;E_1)\). Suppose these solutions are given by
\[
a_{11}x^{\theta_1},\quad a_{12}x^{\theta_1},\quad \ldots,\quad a_{1n_1}x^{\theta_1}.
\]

\noindent
\textbf{Step 3.} For each solution \(a_{1k_1}\,x^{\theta_1}\), consider the transformed function
\[
f_1(x,y) \;=\; f\bigl(x,\,y + a_{1k_1}x^{\theta_1}\bigr).
\]
Generate \(\mathcal{NP}\bigl(f_1\bigr)\) and analyze its edges for co-slopes \(\theta_2 > \theta_1\). 
\begin{itemize}
\item If no edge exists with a co-slope greater than \(\theta_1\), then \(y(x)=a_{1k_1}x^{\theta_1}\) is (at this stage) a valid solution.
\item If such an edge does exist, let \(E_2\) be the edge with co-slope \(\theta_2>\theta_1\).
\end{itemize}

\noindent
\textbf{Step 4.} For the function \(f_1\), repeat the procedure from Step 2, identifying all nonzero solutions
\[
a_{21}x^{\theta_2},\quad a_{22}x^{\theta_2},\quad \ldots,\quad a_{2n_2}x^{\theta_2}
\]
associated with \(E_2\). For each root \(a_{2k_2}x^{\theta_2}\), refine the partial expansion to
\[
y(x) \;=\; a_{1k_1}x^{\theta_1} \;+\; a_{2k_2}x^{\theta_2}.
\]
Then construct
\[
f_2(x,y) \;=\; f_1\bigl(x,\,y + a_{2k_2}x^{\theta_2}\bigr)
\]
and check whether \(\mathcal{NP}(f_2)\) includes any edge with a co-slope greater than \(\theta_2\). Continue similarly if such an edge is present.

\noindent
\textbf{Step 5.} The iteration continues until the expansions reach the desired \emph{order} in \(x\). Specifically, if
\[
y(x)\;=\;a_{1k_1}x^{\theta_1}\;+\;a_{2k_2}x^{\theta_2}\;+\;\cdots+\;a_{s\,k_s}x^{\theta_s}
\]
represents the current partial expansion, the last term \(a_{s\,k_s}x^{\theta_s}\) originated from the newest edge of \(\mathcal{NP}(f_{s-1})\). One can proceed with refinements until no new edges with higher co-slopes emerge or until the series has been established to a specified degree.

\noindent
\textbf{Comment:} Although the description here follows an algorithmic framework, practical implementation frequently involves symbolic manipulations of polynomials. The key concept is to track the new exponents \(\theta_s\) that arise during each refinement step and resolve the corresponding (quasi-homogeneous) equations until the desired depth of the Puiseux expansion is achieved.

\medskip

\noindent
\textbf{Remark on the exponent threshold.} 
Throughout the pseudocode presented below, we define a parameter \(\mathcal{T}\) (``exponent threshold''), which specifies the extent to which we aim to develop the Puiseux series in relation to the exponents of \(x\). Specifically, when a newly obtained coslope \(\theta\) surpasses \(\mathcal{T}\), the algorithm stops further refinement of that branch, considering it sufficiently determined for the current objectives. This distinction clarifies the difference between ``desired expansion degree'' (which limits the exponents) and the conventional total degree of a polynomial.

\medskip

\begin{algorithm}
\caption{Newton--Puiseux Algorithm}
\resizebox{\textwidth}{0.5\textheight}{%
\begin{minipage}{\textwidth}
\begin{algorithmic}[1]
\State \textbf{Input:} A polynomial $f(x,y)\in \mathbb{C}[x,y]$; desired expansion degree $\mathcal{T}$.
\State \textbf{Output:} Puiseux expansions $y(x)$ such that $f\bigl(x,y(x)\bigr)=0$ up to order $\mathcal{T}$.
\Statex
\Procedure{NewtonPuiseux}{$f(x,y),\mathcal{T}$}
  \State Compute the Newton polygon $\mathcal{NP}(f)$.
  \For{each edge $E_1$ of $\mathcal{NP}(f)$ with co-slope $\theta_1$}
    \State Identify the quasi-homogeneous polynomial $f(x,y;E_1)$ and solve $f(x,y;E_1)=0$ for nonzero roots $a_{1k_1}x^{\theta_1}$.
    \For{each root $a_{1k_1}x^{\theta_1}$}
      \State Define $f_1(x,y) \gets f\bigl(x,y + a_{1k_1}x^{\theta_1}\bigr)$.
      \State Compute $\mathcal{NP}(f_1)$; let $\theta_{\text{next}}$ be the smallest co-slope $>\theta_1$ (if such a value exists).
      \If{$\theta_{\text{next}}$ does not exist or $\theta_{\text{next}}>\mathcal{T}$ (exceeds desired order)}
        \State Record partial expansion $y(x)=a_{1k_1}x^{\theta_1}$.
      \Else
        \State \Call{RefineExpansion}{$f_1, \theta_{\text{next}}, \mathcal{T},$ partialExpansion=$a_{1k_1}x^{\theta_1}$}
      \EndIf
    \EndFor
  \EndFor
  \State \textbf{return} All expansions identified up to order $\mathcal{T}$.
\EndProcedure
\Statex

\Procedure{RefineExpansion}{$f_m, \theta_m, \mathcal{T},$ partialExpansion}
  \State Solve $f_m(x,y;E_m)=0$ on the new edge $E_m$, giving roots $a_{m1}x^{\theta_m}, a_{m2}x^{\theta_m},\dots$
  \For{each root $a_{mk_m}x^{\theta_m}$}
    \State updatedExpansion $\gets$ partialExpansion $+\, a_{mk_m}x^{\theta_m}$
    \If{$\theta_m > \mathcal{T}$}
      \State \textbf{break} (sufficient order reached)
    \Else
      \State $f_{m+1}(x,y)\gets f_m\bigl(x,y+a_{mk_m}x^{\theta_m}\bigr)$
      \State Recompute $\mathcal{NP}(f_{m+1})$; next co-slope $\theta_{m+1}$ $>$ $\theta_m$ (if any)
      \If{$\theta_{m+1}$ does not exist or $>\!\mathcal{T}$}
        \State Record \textit{updatedExpansion} 
      \Else
        \State \Call{RefineExpansion}{$f_{m+1}, \theta_{m+1}, \mathcal{T},$ updatedExpansion}
      \EndIf
    \EndIf
  \EndFor
\EndProcedure
\end{algorithmic}
\end{minipage}
}
\end{algorithm}

\medskip

\noindent
\textbf{Remark (multiple edges with identical coslope).}
In certain circumstances, the Newton polygon may include multiple edges sharing the same coslope. Practically, this implies the presence of several line segments positioned on parallel lines at the same height within the diagram. In such instances, all monomials from \(\mathrm{Supp}(f)\) lying on \emph{any} of these parallel edges are collected. Subsequently, the factorization of the corresponding quasi-homogeneous polynomials is carried out. Each factor yields a potential branch (local solution) for the Puiseux expansion. It is noteworthy that some factors may be identical or repeated, which could indicate branching or higher multiplicities within the local expansions.

\medskip

\noindent
\textbf{Remark (shifting the reference point).}
Although the Newton-Puiseux algorithm is typically described for expansions centered around \(x=0\), the same methodology is applicable for investigating local behavior around any designated point \((x_0,y_0)\). By translating the coordinates, specifically setting \(X = x - x_0\) and \(Y = y - y_0\), one can construct the Newton polygon of \(f(X + x_0,\, Y + y_0)\) and derive Puiseux expansions of \(Y(X)\). Consequently, the curve \(f(x,y)=0\) can be decomposed globally into local branches at all of its (finite) singular or nonsingular points.

\medskip

\noindent
\textbf{Example illustrating the Newton-Puiseux algorithm.} 
Consider the polynomial 
\[
f(x,y)=y^6 - 3xy^4 + 3x^2y^2 - x^3 - 2xy^5 + 4x^2y^3 - 2x^3y + 8x^5.
\]
Observe that \(f\) can be rewritten as
\[
f(x,y) = (y^2 - x)^3 - 2xy(y^2 - x)^2 + 8x^5,
\]
which notably reveals that \(\Supp(f)\) contains the exponent points
\[
\{(0,6), (1,4), (2,2), (3,0), (1,5), (2,3), (3,1), (5,0)\}.
\]
The corresponding Newton polygon is illustrated in Figure~\ref{fig:4}. It features exactly one compact edge \(E\) connecting \((3,0)\) and \((0,6)\). By restricting \(f\) along that edge, we derive:
\[
f(x,y;E)= y^6 - 3xy^4 + 3x^2y^2 - x^3=(y^2 - x)^3=(y - x^{1/2})^3\bigl(y + x^{1/2}\bigr)^3,
\]
indicating that the associated quasi-homogeneous polynomial factors into two distinct roots, \(\pm\,x^{1/2}\).  

\medskip

\begin{figure}[ht]
\centering
\begin{tikzpicture}[scale=0.7]
  % grid and axes
  \draw[step=1cm, gray, very thin] (-0.5,-0.5) grid (6.5,7.5);
  \draw[->] (-0.5,0) -- (7,0) node[right] {$x$};
  \draw[->] (0,-0.5) -- (0,8) node[above] {$y$};

  % fill the Newton diagram region
  \fill[blue!20, opacity=0.5] (6.5,0) -- (6.5,7.5) -- (0,7.5) -- (0,6) -- (3,0) -- cycle;

  % draw the edge of the polygon in red
  \draw[very thick, red] (3,0) -- (0,6);

  % mark the points of Supp(f)
  \fill (3,0) circle (2pt) node[below right] {$(3,0)$};
  \fill (5,0) circle (2pt) node[above right] {$(5,0)$};
  \fill (2,2) circle (2pt) node[above right] {$(2,2)$};
  \fill (2,3) circle (2pt) node[above right] {$(2,3)$};
  \fill (1,4) circle (2pt) node[below left] {$(1,4)$};
  \fill (1,5) circle (2pt) node[above right] {$(1,5)$};
  \fill (0,6) circle (2pt) node[above right] {$(0,6)$};

\end{tikzpicture}
\caption{Newton polygon of $f(x,y)=y^6-3xy^4+3x^2y^2-x^3-2xy^5+4x^2y^3-2x^3y+8x^5$. There is exactly one compact edge, from $(3,0)$ to $(0,6)$.}
\label{fig:4}
\end{figure}

\medskip

\noindent
\textbf{Starting with \(y(x)=x^{1/2}\).} We define 
\[
f_1(x,y)=f\bigl(x,y + x^{1/2}\bigr).
\]
One then computes
\[
\begin{aligned}
f_1(x,y)=&f(x,y+x^{1/2})\\=&y^3(y+2x^{1/2})^3-2x(y+x^{1/2})y^2(y+2x^{1/2})^2+8x^5\\
=&y^6+6x^{1/2}y^5+6xy^4+8x^{3/2}y^3-2xy^5-10x^{3/2}y^4\\ &-16x^2y^3-8x^{5/2}y^2+8x^5\\
\Supp(f_1)&=\{(0,6), (1/2,5),(1,4),(3/2,3),(1,5),(3/2,4),(2,3),(5/2,2),(5,0) \},
\end{aligned}
\]
and its support comprises exponents that include fractional \(x\)-powers (resulting from the shifts by \(x^{1/2}\)). The Newton polygon of \(f_1\) is illustrated in Figure~\ref{fig:5}. It features three edges, designated as \(E_1\), \(E_2\), and \(E_3\), with co-slopes of \(5/4\), \(1\), and \(1/2\), respectively. Since we aim to find expansions with exponents \(\theta > 1/2\) (indicating that the subsequent terms in the Puiseux series have exponents greater than \(1/2\)), we concentrate on edges \(E_1\) and \(E_2\). We find that:
\[
\begin{aligned}
f_1(x,y;E_1)&=8x^{3/2}y^3-8x^{5/2}y^2=8x^{3/2}y^2(y-x)\\
f_1(x,y;E_2)&=-8x^{5/2}y^2+8x^5=-8x^{5/2}(y-x^{5/4})(y+x^{5/4}).
\end{aligned}
\]
By extracting the nonzero roots of these quasi-homogeneous polynomials, we obtain the following terms in the expansions:
\[
y_1(x) = x^{1/2} + x +\hot,
\quad
y_2(x) = x^{1/2} + x^{5/4} +\hot,
\]
\[
y_3(x) = x^{1/2} - x^{5/4} +\hot.
\]
Thus, starting with the root \(x^{1/2}\), we derive three distinct solutions.

\medskip

\noindent
\textbf{Starting with \(y(x)=-x^{1/2}\).} When we apply the same methodology to the second root \(-\,x^{1/2}\), we also obtain three expansions, represented as:
\[
y_4(x) = -x^{1/2} + x+\hot,
\quad
y_5(x) = -x^{1/2} + ix^{5/4} +\hot,
\]
\[
y_6(x) = -x^{1/2} - ix^{5/4}+\hot.
\]
In total, these six series encompass all possible roots up to the specified order in \(x\). In fact, from the original edge \(E\) of \(\mathcal{NP}(f)\), we observe that the degree of \(f\) along that edge with respect to $y$ is \(6\), which leads us to expect exactly six expansions of the form \(y(x)=\cdots\). No further solutions emerge in later refinement steps, as each branch has been entirely accounted for through the corresponding edge analysis.

\begin{figure}[ht]
\centering
\begin{tikzpicture}[scale=0.7]
  % grid and axes
  \draw[step=1cm, gray, very thin] (-0.5,-0.5) grid (6.5,7.5);
  \draw[->] (-0.5,0) -- (7,0) node[right] {$x$};
  \draw[->] (0,-0.5) -- (0,8) node[above] {$y$};

  % fill the Newton diagram region
  \fill[blue!20, opacity=0.5] (6.5,0) -- (6.5,7.5) -- (0,7.5) -- (0,6)-- (1.5,3)-- (2.5,2) -- (5,0) -- cycle;

  % draw the edges of the polygon in red
  \draw[very thick, red] (2.5,2) -- (5,0) node[midway,below left]{$E_1$};
  \draw[very thick, red] (1.5,3) -- (2.5,2) node[midway,below left]{$E_2$};
  \draw[very thick, red] (1.5,3) -- (0,6)   node[midway,below left]{$E_3$};

  % mark points of Supp(f_1)
  \fill (0,6) circle (2pt);
  \fill (0.5,5) circle (2pt);
  \fill (1,4) circle (2pt);
  \fill (1.5,3) circle (2pt);
  \fill (1,5) circle (2pt);
  \fill (1.5,4) circle (2pt);
  \fill (2,3) circle (2pt);
  \fill (2.5,2) circle (2pt);
  \fill (5,0) circle (2pt);

\end{tikzpicture}
\caption{The Newton polygon of $f_1(x,y)=f\bigl(x,y+x^{1/2}\bigr)$, illustrating three edges $E_1$, $E_2$, and $E_3$, with co-slopes $5/4$, $1$, and $1/2$, respectively.}
\label{fig:5}
\end{figure}

% ========================================================
\section{Implementation details and controls (for reproduction)}
\label{app:impl}
% Appendix B
% ========================================================

\paragraph{Notation clash}
We reserve the symbol $\Delta$ for the \emph{local neighbourhood radius} utilized to sample perturbations around an anchor, and use $\delta$ to denote the \emph{margin threshold} in the uncertainty rule (Refer to Section~\ref{sec:notation}). Temperature is consistently represented by $T$.

\subsection{Anchor mining (uncertainty rules)}
Anchors are identified using the standard union rule
\[
  \mathcal{U}_\star \;=\; \bigl\{z:\max_k p_k(z;T) < \tau \bigr\}\;\cup\;
                           \bigl\{z: |p_{(1)}(z;T)-p_{(2)}(z;T)| < \delta \bigr\},
\]
applied exactly as described in \texttt{up\_real.py} and \texttt{up\_radio.py}. When a calibrator is learned on the validation set (using temperature / vector / Platt, or isotonic / beta techniques), the \emph{calibrated} probabilities derived from the validation split are used to define $(\tau,\delta)$, and the same calibrator is subsequently applied to the test set for unbiased selection and assessment.

\subsection{\(\C^2\) feature compression (inputs to the CVNN)}
Both the MIT--BIH and RadioML pipelines compress raw time windows to $\C^2\!\cong\!\R^4$:
\begin{itemize}
  \item MIT--BIH: \texttt{complex\_stats} (which includes four complex-aware summary statistics),
  \item RadioML: \texttt{stft\_stats} (which consists of four STFT-based statistics).
\end{itemize}
The arrangement follows $[\Re_1,\Re_2,\Im_1,\Im_2]$ to align with the model’s first real layer, which has a size of $4$.

\subsection{Local polynomial surrogate in $\C^2$ (exact implementation)}
For a given anchor $z_\star$, we fit the logit-difference surrogate
\[
  \widehat f_d(\xi,\eta) \;=\; \sum_{i+j\le d} c_{ij}\,\xi^i\eta^j,\qquad (\xi,\eta)=(x-x_\star,y-y_\star),
\]
using the following \textbf{defaults} (identical to the released code unless stated otherwise):
\begin{itemize}
  \item the degree is set to  $d{=}4$; constant and all linear terms are \emph{removed} (to enforce $\widehat f_d(0,0)=0$ and zero gradient),
  \item samples: $N\!\approx\!600$ (per anchor in experiments; the function’s default is $2000$) drawn i.i.d.\ within $[-\Delta,\Delta]^4$,
 \item neighbourhood radius: $\Delta\!=\!0.05$ (as per experiments; function default is $0.01$),
  \item kink rejection band: \texttt{exclude\_kink\_eps} $=10^{-6}$ (swept in ablations),
  \item weighting: \texttt{weight\_by\_distance=True} (see below),
  \item ridge regularization: $\lambda{=}10^{-8}$ (Tikhonov).
\end{itemize}

\paragraph{Design matrix and monomials}
We consider all monomials with total degrees such that $2\!\le\!i{+}j\!\le\!d$, resulting in the number of unknowns given by 
$M=\binom{d+2}{2}-3$ (the $3$ omitted terms include the constant and two linear terms).
Columns are scaled using $\ell_2$ norms to ensure unit length, which improves conditioning.

\paragraph{Near-kink filtering (piecewise-holomorphic activations)}
If the network utilizes a modReLU-/zReLU-like first complex layer (indicated by the model attribute \texttt{bias\_modrelu} or the layer name),
we compute the per-sample pre-activation \emph{margins}
\[
  \textstyle s_{n,h} \;=\; \bigl|a_{n,h}\bigr| + b_h
\]
where $a_{n,h}$ represents the complex pre-activation and $b_h$ is the modReLU bias at the hidden unit $h$.
Any perturbation for which $\min_h s_{n,h}\le\varepsilon_{\text{kink}}$ is excluded 
(\texttt{exclude\_kink\_eps}).
This step prevents the mixing of different holomorphic sheets in the fitting process.

\paragraph{Sample weighting}
For the samples that remain, we assign weights using the formula
\[
  w_n \;\propto\; \max\!\bigl(\min_h s_{n,h},\,0\bigr) + 10^{-9},
\]
which means perturbations that are further away from modReLU kinks will have a greater effect on the weighted least-squares fitting.
(If no modReLU-like layer is identified, we revert to using unit weights.)

For space-filling sampling one may use Latin Hypercube designs \citep{mckay2000comparison} or low-discrepancy sequences such as (scrambled) Sobol' nets \citep{sobol1967distribution,owen1997quasi}; general DoE guidance is summarized in \citep{dean1999design}.

\paragraph{Estimator and stability checks}
We solve
\[
  \min_{\mathbf c}\;\|W^{1/2}(A\mathbf c - \mathbf F)\|_2^2 \;+\; \lambda\|\mathbf c\|_2^2,
\]
with $W=\operatorname{diag}(w_1,\dots,w_N)$. As an alternative, the ridge parameter can be selected via generalized cross-validation (GCV) in the smoothing-spline tradition \citep{golub1979gcv,wahba1990spline}.
We report $\operatorname{cond}(A)$ and $\operatorname{rank}(A)$.
If $\operatorname{cond}(A)\!>\!10^{10}$, $\operatorname{rank}(A)\!<\!M$, or the kept-ratio
(\#usable samples / $N$) falls below $0.25$, we \emph{fallback} by shrinking the radius
($\Delta\!\leftarrow\!0.5\,\Delta$) and/or reducing $d$ (down to $\ge2$), then refit.
This logic is implemented in \texttt{local\_poly\_approx\_complex(…)}.

\paragraph{Fidelity metrics}
On an independent set ($\approx$200--500 perturbations), we calculate RMSE, MAE, Pearson correlation, and sign agreement:
\[
\mathrm{SA}=\tfrac1n\sum_{i}\mathbb{I}\!\left[\operatorname{sign}\widehat f_d
                   =\operatorname{sign}f\right]\!,
\]
which is returned by \texttt{evaluate\_poly\_approx\_quality(…)} (we also report Pearson’s $\rho$) \citep{benesty2009pearson}. We include MAE since, for average performance assessment, it can be preferable to RMSE \citep{willmott2005advantages}.

\subsection{Kink prevalence score}
We assess local non-holomorphicity by sampling small spheres around $z_\star$ and analyzing the angular dispersion of the gradients $\nabla f$ directions; the \texttt{kink\_score} indicates the \emph{standard deviation} (measured in radians) of these angles. Higher values suggest a greater degree of piecewise behavior.

\subsection{Newton--Puiseux stage (plumbing only)}
The surrogate $\widehat f_d$ is symbolically \texttt{factor}ed and taken through our Puiseux routine to address fractional-power branches. For mathematical and algorithmic details, refer to \ref{app:newton}; our focus here is on noting that the implementation yields (branch exponent, multiplicity, leading coefficient, orientation) for each branch.

\subsection{Recommended defaults and how to tune them}
\begin{itemize}
  \item \textbf{Degree $d$:} Start with a value of $4$; if $\rho_{\mathrm P}<0.2$ or $\operatorname{cond}(A)$ becomes excessively large, try $d\in\{3,2\}$.
  \item \textbf{Radius $\Delta$:} Set so that a typical perturbation results in an approximate 1--3\% relative change in logits; in our experiments, the code defaults to $0.05$.
  \item \textbf{Samples $N$:} $600$ is sufficient for $d{=}4$; consider increasing to $1000$ if the kink filter discards many points.
  \item \textbf{Kink band $\varepsilon_{\text{kink}}$:} $10^{-6}$ is safe; we sweep up to $10^{-2}$ in \ref{app:impl} to show robustness.
  \item \textbf{Regularization $\lambda$:} Maintain a small Tikhonov value ($10^{-8}$); increase by one order of magnitude if rank deficiency remains an issue.
\end{itemize}

\begin{table}[h]
\centering
\small
\begin{tabularx}{\linewidth}{l X}
\toprule
\textbf{Component} & \textbf{Default / Setting} \\
\midrule
Local neighborhood & \(\Delta=0.05\) in \emph{scaled} feature space (\(\R^4\!\cong\!\C^2\)) \\
Polynomial degree & \(d=4\); constant and linear terms omitted (\(i{+}j<2\) fixed to 0) \\
Fit samples / eval samples & \(N=600\) (fit), \(N_{\text{eval}}=200\) (fidelity) \\
Weights and filtering & Distance weighting, kink-aware exclusion with \(\texttt{kink\_eps}=10^{-6}\) \\
Stability & Column scaling; Tikhonov applied if necessary (\(\lambda=10^{-8}\)); \texttt{min\_keep\_ratio}=0.25 \\ 
Kink diagnostics & 2\,000 draws; \(\Delta=0.05\); \(\texttt{kink\_eps}\) sweep \(10^{-6}\!\to\!10^{-2}\) \\
Puiseux precision & Symbolic factorization; precision set to \(4\) (timing recorded for each stage) \\
Robustness rays & 20 random phase directions; testing radius \(0.02\); 20 iterations \\
Calibration (CV) & For MIT--BIH: \emph{none}, temperature, isotonic; for RadioML: \emph{none}, Platt (logistic), isotonic. 5-fold stratified CV (calibrator trained on validation set, applied to test). \\
ECE bins & 10--15 bins (per script); confidence intervals via \(t\)-intervals; Wilcoxon tests for pairwise \\
\bottomrule
\end{tabularx}
\caption{Defaults matching the released code.}
\label{tab:defaults}
\end{table}

% ========================================================
\section{Compute and complexity of the local analysis}
\label{app:compute}
% Appendix C
% ========================================================

\subsection{Computational notes (implementation)}
The surrogate is transformed into a symbolic expression and factorized using \texttt{sympy}’s routines, with classical Hensel-lifting--style factorization underpinning the approach \citep{zassenhaus1969factor}. We log wall time for each stage (sampling, least squares, factorization, simplification, and Puiseux) and report CPU/GPU memory usage as part of the resource benchmark. Refer to Table~\ref{tab:res-benchmark} for additional information.

\subsection{What we time and what we measure}
For every anchor, \texttt{benchmark\_local\_poly\_approx\_and\_puiseux(…)} provides:
$\text{time\_sampling}$, $\text{time\_lstsq}$, $\text{time\_factor}$, $\text{time\_simplify}$, $\text{time\_puiseux}$, $\text{time\_total}$.
We compare this against a single backpropagation-based saliency pass through \texttt{time\_gradient\_saliency(…)}, which records wall-clock time and peak GPU memory (using \texttt{torch.cuda.max\_memory\_allocated}), along with the CPU RSS change.

\subsection{Asymptotics and empirical scaling}
Let $M=\binom{d+2}{2}-3$ represent the number of active monomials.
Sampling operates at $O(N)$; solving the weighted ridge regression can be $O(NM^2)$ (using the normal equations) or $O(NM)$ if a stable least-squares backend is employed. The symbolic \texttt{factor} and Puiseux routine exhibit super-linear scaling with respect to $d$ and the number of nonzero terms; in practice, these operations are typically dominated by small degrees ($d\in\{3,4,5\}$) and sparse supports.

\subsection{Hardware and software}
All timing measurements are conducted using \texttt{time.perf\_counter()} on the training host; GPU peak statistics are recorded for each anchor after resetting CUDA peak counters before every call. Symbolic algebra relies on \texttt{sympy} with default settings.

\begin{table}[h]
\centering
\caption{Resource benchmark (per anchor). Values presented are median [IQR] for anchors from 17 MIT--BIH and 21 RadioML anchors; times are measured in milliseconds. The saliency reports a single pass of gradient-based saliency. Peak GPU usage is tracked during the Puiseux pipeline, with saliency peak measurements being equivalent within measurement noise. }
\label{tab:res-benchmark}
\begin{tabularx}{\linewidth}{lXXXXXX}
\toprule
& Sampling & LSQ & Factor & Simplify & Puiseux & Total \\
\midrule
MIT--BIH (ms)  & 0.0 [0.0--0.0] & 280.0 [280.0--290.0] & 10.0 [10.0--10.0] & 90.0 [90.0--100.0] & 3700.0 [3590.0--3870.0] & 4080.0 [4000.0--4260.0] \\
RadioML (ms)   & 0.0 [0.0--0.0] & 290.0 [280.0--290.0] & 10.0 [10.0--10.0] & 90.0 [90.0--90.0]   & 3660.0 [3460.0--3740.0] & 4060.0 [3850.0--4140.0] \\
\midrule
Saliency (ms)  & \multicolumn{5}{c}{4.77 [4.52--5.00]} &  \\
GPU peak (MB)  & \multicolumn{5}{c}{64.0 [64.0--64.0]} &  \\
\bottomrule
\end{tabularx}
\end{table}

\subsection{Failure modes and their mitigations}
The primary compute failure modes identified are: (i) rank deficiency / ill-conditioning of $A$, which is addressed by radius shrinking and degree fallback, (ii) protracted symbolic \texttt{factor} execution times for dense degree-5 fits, mitigated by early termination at $d{=}4$, and (iii) excessive kink pruning, which can be alleviated by increasing either $N$ or $\varepsilon_{\text{kink}}$).

% ========================================================
\section{Sensitivity of anchors to $(\tau,\delta)$}
\label{app:sensitivity}
% Appendix D
% ========================================================

This section evaluates how the uncertainty thresholds $(\tau,\delta)$ influence: (i) the quantity of anchors, (ii) the spatial distribution of anchors within feature space, (iii) the \emph{capture} of actual errors, (iv) the \emph{precision} of the uncertain flag, and (v) the \emph{accepted-set risk}.

\subsection{Grid evaluation protocol}
For a grid $\tau\!\in\![0.50,0.95]$ and $\delta\!\in\![0.00,0.60]$
(the defaults in \texttt{up\_real.py} and \texttt{up\_radio.py}), we compute:
\begin{align*}
\textstyle \text{abstain} &= \frac{\#\{z:\max p_k(z;T){<}\tau\;\text{or}\;|p_{(1)}{-}p_{(2)}|{<}\delta\}}{n},\\
\textstyle \text{capture} &= \frac{\#\{\text{misclassified }z\;\text{flagged uncertain}\}}{\#\{\text{misclassified }z\}},\\
\textstyle \text{precision} &= \frac{\#\{\text{misclassified }z\;\text{flagged uncertain}\}}{\#\{\text{flagged uncertain}\}},\\
\textstyle \text{risk}_{\text{accept}} &= \frac{\#\{\text{misclassified }z\;\text{not flagged}\}}{\#\{\text{not flagged}\}}.
\end{align*}
Additionally, we monitor a normalized \emph{dispersion} of the flagged set: given that $U$ denotes the set of flagged samples, $\bar x_U$ represents its centroid in $\R^4$, and $\mathrm{gspread}$ is the global maximum--minimum spread of the complete set, the dispersion is calculated as
\[
\textstyle \mathrm{dispersion} \;=\; \frac{1}{|U|}\sum_{x\in U} \frac{\|x - \bar x_U\|_2}{\mathrm{gspread}}\!,
\]
with a stable fallback to $0$ for $|U|\le1$. These concepts are implemented in \texttt{sensitivity\_analysis(…)} and are logged to \texttt{sens\_grid.csv}.

\subsection{Reporting and selection under a review budget}
We normalize \texttt{abstain} and \texttt{capture} to the range $[0,1]$ and present $\mathrm{kink}\!=\!\mathrm{capture}\!-\!\mathrm{abstain}$ as a unified \emph{benefit--cost} summary (which is also stored). For a specified integer \texttt{review\_budget}, we determine $(\tau^\star,\delta^\star)$ by ordering flagged points based on $p_{\max}$ (for $\tau$) and the top-2 margin (for $\delta$), truncating to fit the budget (\texttt{select\_thresholds\_budget\_count(…)}).

% ========================================================
\section{Calibration reporting: CIs, tests, and multiplicity mis-estimation}
\label{app:calibration}
% Appendix E
% ========================================================

\subsection{ Comparison criteria and percentage calculations }
Unless otherwise specified, all relative improvements (such as “61\% ECE drop”) are calculated \emph{relative to the uncalibrated softmax} ($T{=}1$), maintaining consistency across identical folds and samples:
\[
\textstyle \mathrm{RelDrop}(\mathrm{ECE}) \;=\;
  \frac{\overline{\mathrm{ECE}}_{\text{NONE}} - \overline{\mathrm{ECE}}_{\text{method}}}
       {\overline{\mathrm{ECE}}_{\text{NONE}}}\times 100\%.
\]
Averages are computed across folds; we also provide raw per-fold CSVs.

\subsection{Confidence intervals and non-parametric testing}
For each method and metric $\in\{\text{ECE},\text{NLL},\text{Brier},\text{Acc},\text{AUC}\}$, we compute a $95\%$ Student-$t$ confidence interval as follows:
\[
\bar x \pm t_{0.975,\;n-1}\;\frac{s}{\sqrt{n}},
\]
utilizing the function \texttt{mean\_ci\_t(…)}. Pairwise comparisons of methods employ the Wilcoxon signed-rank test across folds, using the alternative \emph{“less”} for error metrics (ECE, NLL, Brier) and \emph{“greater”} for Accuracy/AUC; $p$-values are reported when $n\ge5$ and foldwise differences are non-degenerate. We also include a \emph{win-rate vs.\ NONE} table, detailing the number of folds in which a method’s ECE is lower than that of NONE. This mirrors the format used in the released scripts.

\subsection{Stability against re-parameterizations of $(\Re,\Im)$}
In our ablation studies, we ensure that the calibration trends remain consistent when applying global complex rotations and channel permutations (unitary transformations of the $\R^4$ input). This is achieved by adjusting the validation/test features and reapplying the calibrated model.

\subsection{ From multiplicity to temperature: sensitivity to mis-estimation}
Our phase-aware temperature relies on a heuristic relationship between the estimated branch multiplicity $m$ and the effective temperature $T$:
\[
\textstyle T \propto m^{-\gamma},\qquad \gamma \in [0.5,1].
\]
When the surrogate’s factorization causes a relative error $\varepsilon$ in $m$, such that $m_{\text{est}}=m_{\text{true}}(1+\varepsilon)$, the proportional temperature adjustment is given by
\[
\textstyle T_{\text{mult}} \;=\; \Bigl(\frac{m_{\text{true}}}{m_{\text{est}}}\Bigr)^{\gamma}
\;=\; (1+\varepsilon)^{-\gamma}.
\]
We explore the effects of $\varepsilon\!\in\!\{-0.50,-0.25,-0.10,-0.05,0,0.05,0.10,0.25,0.50\}$ with $\gamma\!=\!0.5$ (the default) and report the resulting ECE for each scenario using \texttt{sweep\_multiplicity\_mis-estimation(…)}.

\subsection{Fold structure and fairness}
All calibration methods utilize identical splits and a single validation calibrator that is learned once per fold. For multiclass-to-binary reporting, we implement the \emph{top-2} reduction. Metrics for each fold are recorded in \texttt{cv\_metrics\_per\_fold\_multi.csv} and summarized into a confidence interval table (\texttt{calibration\_ci\_table.csv}) along with a human-readable report.

% ========================================================
\section{Datasets and Preprocessing (extended)}
\label{app:data}
% ========================================================

\subsection{Controlled synthetic $\C^2$ helix (full details)}
\label{app:data-synth-redirect}
All relevant information about the synthetic $\C^{2}$ helix (including data, figures, and methodologies) can be found in \ref{app:synthetic_sanity}.

\medskip

% -----------------------------------------------------------
\subsection{MIT--BIH Arrhythmia (signal processing and feature construction)}
\label{app:data-mitbih}

The evaluation is conducted using the MIT--BIH Arrhythmia Database \citep{moody2001mitbih,goldberger2000physiobank}, which was recorded at a frequency of \SI{360}{\hertz} across two leads (channel~0: MLII, channel~1: V1). In accordance with \citep{dibin2022hilbert}, the task is framed as a binary classification challenge between regular beats (designated as class~N) and premature ventricular contractions (PVC, designated as class~V).

\paragraph{Preprocessing pipeline}
Algorithm~\ref{alg:mitbih} describes the procedure. A zero-phase Butterworth band-pass filter (operating between \SIrange{0.5}{40}{\hertz}, 2nd-order, zero-phase via \texttt{filtfilt}) is applied to eliminate baseline wander and high-frequency noise. For each annotated beat, we extract a window containing \(L=128\) samples starting \(P=50\) samples before the R peak. The analytic signal is computed using the discrete Hilbert transform \citep{boashash1992estimating,benitez2001use}.

\begin{algorithm}[t]
\caption{MIT--BIH signal processing}\label{alg:mitbih}
\begin{algorithmic}[1]
\Require{record IDs $\mathcal{R}$, window size $L=128$, pre-samples $P=50$}
\For{$r\in\mathcal{R}$}                             
  \State Load raw signal and annotations
  \State Band-pass filter $0.5$--$40\,$Hz (2nd-order Butterworth; zero-phase)
  \ForAll{beats labeled N or V}
      \If{segment $[t-P,t-P+L]$ within bounds}
         \State Extract segment (both channels)
         \State Compute analytic signal via Hilbert transform
         \State Concatenate $\Re$ and $\Im$ parts $\to$ feature vector
      \EndIf
  \EndFor
\EndFor
\State Standardize features \emph{per patient} on the training set to prevent leakage; then, perform a patient-wise, class-stratified split (80/20).
\end{algorithmic}
\end{algorithm}

\paragraph{Complex compression to \(\C^{2}\)}
Training CVNNs directly on \(4L\) real features is inefficient and increases the risk of overfitting.
Hence, we compress each segment into \(\C^{2}\) by calculating the first two complex moments as follows: 
\begin{equation}\label{eq:c2-compress}
  \Phi(z)=\bigl(\mu,\Delta\bigr),\quad
  \mu=\frac{1}{2L}\sum_{t=1}^{2L} z_{t},\quad
  \Delta=\frac{1}{2L-1}\sum_{t=1}^{2L-1}(z_{t+1}-z_{t}),
\end{equation}
where \(z \in \C^{2L}\) stacks the per-lead analytic signals.
Writing \(\mu=\mu_{r}+i\mu_{i}\) and
\(\Delta=\Delta_{r}+i\Delta_{i}\)
results in a four-dimensional real vector
\(
[\mu_{r},\Delta_{r},\mu_{i},\Delta_{i}]\in\R^{4},
\)
which serves as input to the CVNN.

\begin{figure}[t]
  \centering
  \includegraphics[width=.55\linewidth]{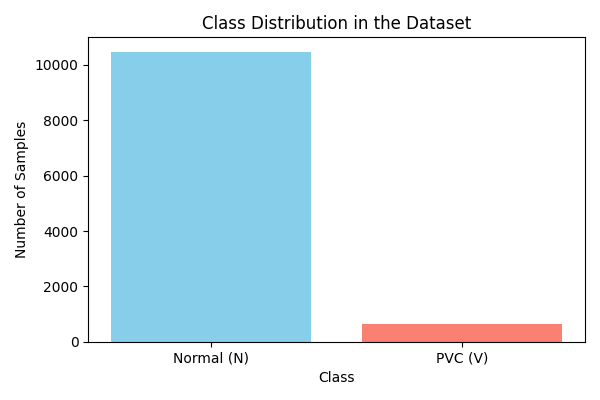}
  \caption{Class distribution in the MIT--BIH subset.}
  \label{fig:class-dist}
\end{figure}

\begin{figure*}[t]
  \centering
  \includegraphics[width=.46\linewidth]{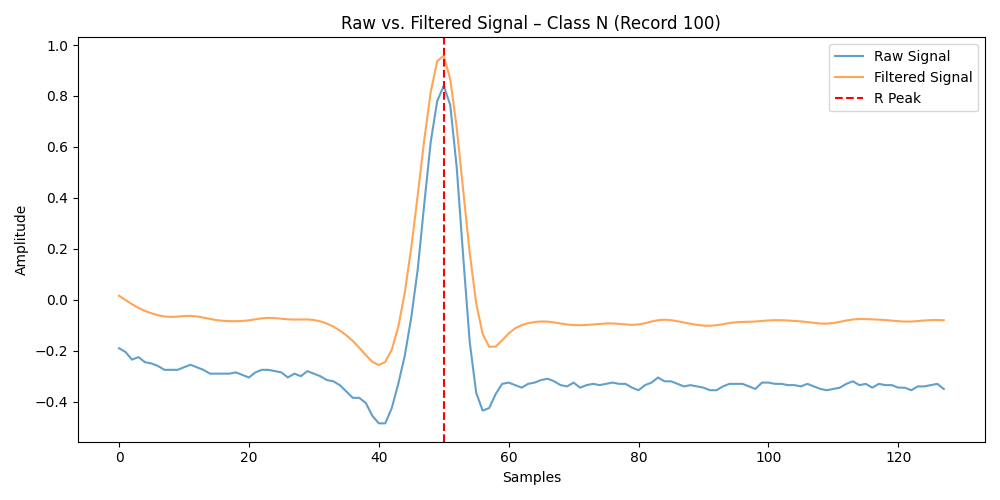}\hfill
  \includegraphics[width=.46\linewidth]{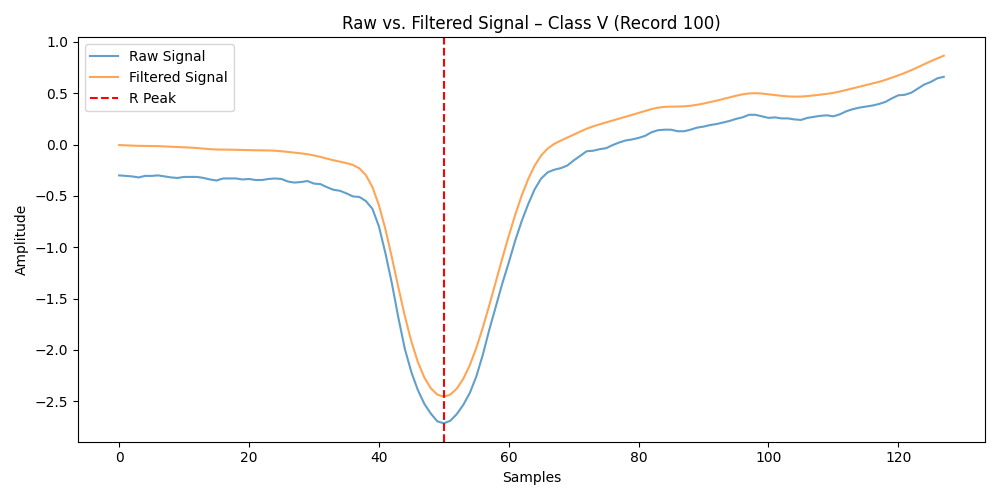}
  \caption{Raw vs. band‑pass filtered ECG segments (lead II). Left: normal
           beat (N); right: PVC (V).  Dashed line marks the R peak.}
  \label{fig:segment-N}
\end{figure*}

\begin{figure*}[t]
  \centering
  \includegraphics[width=.46\linewidth]{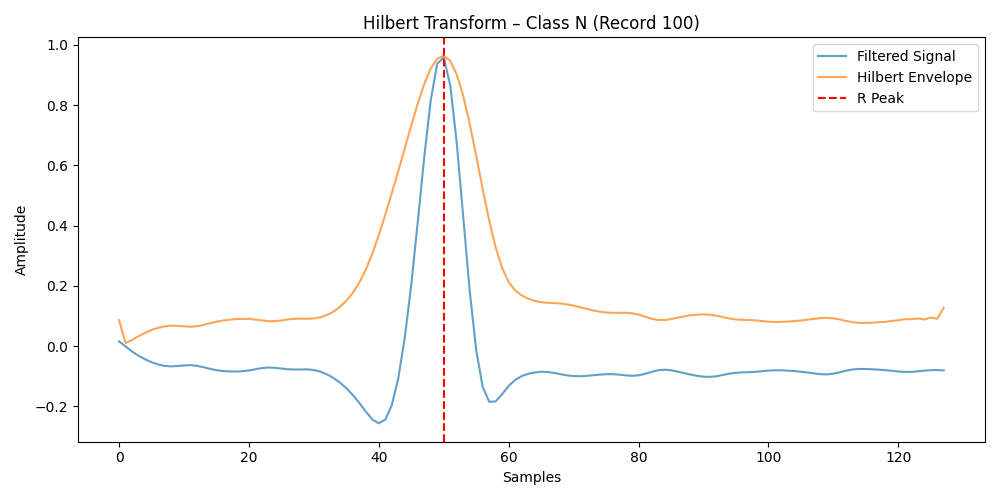}\hfill
  \includegraphics[width=.46\linewidth]{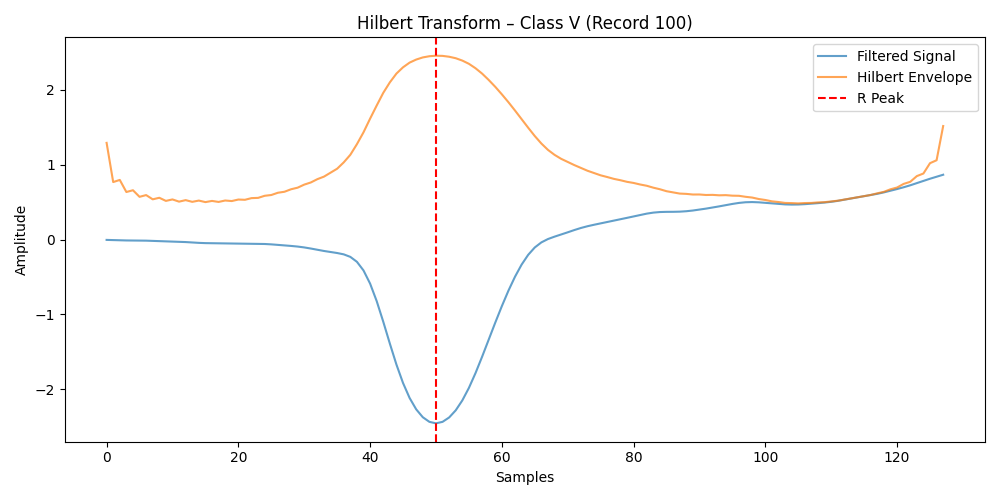}
  \caption{Hilbert-transform envelopes for the same segments as
           Figure~\ref{fig:segment-N}.}
  \label{fig:hilbert}
\end{figure*}

\begin{figure*}[t]
  \centering
  \includegraphics[width=.46\linewidth]{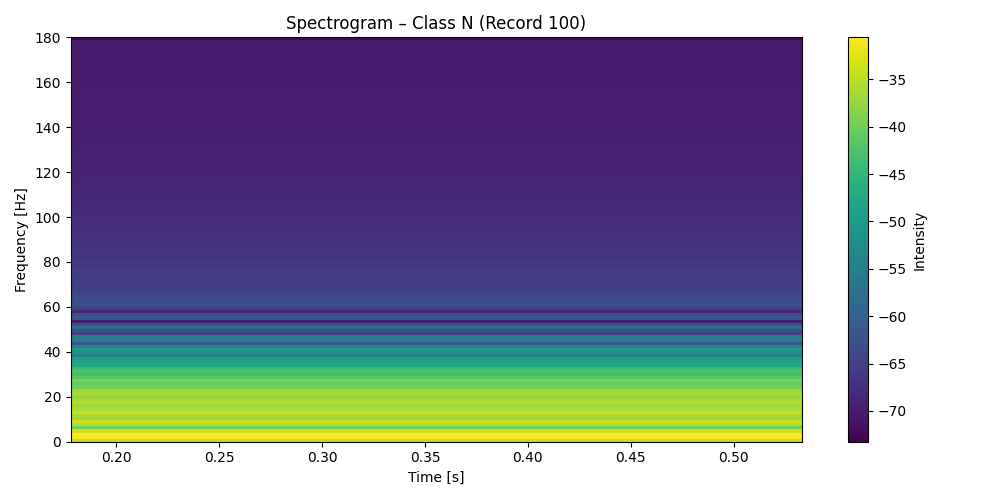}\hfill
  \includegraphics[width=.46\linewidth]{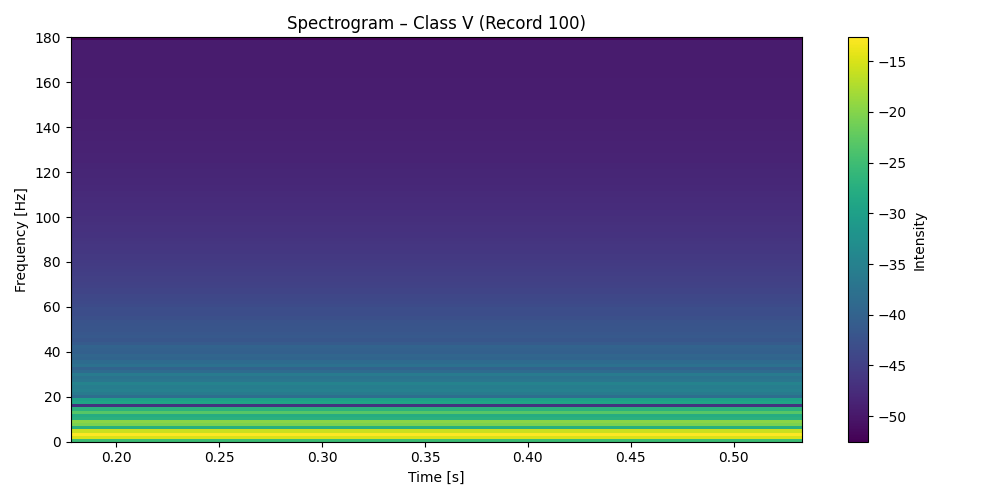}
  \caption{Short‑time Fourier spectrograms (Hamming window, 32 samples, 50\% overlap) \citep{smith2003digital}.}
  \label{fig:spectrograms}
\end{figure*}

\begin{figure}[t]
  \centering
  \includegraphics[width=.6\linewidth]{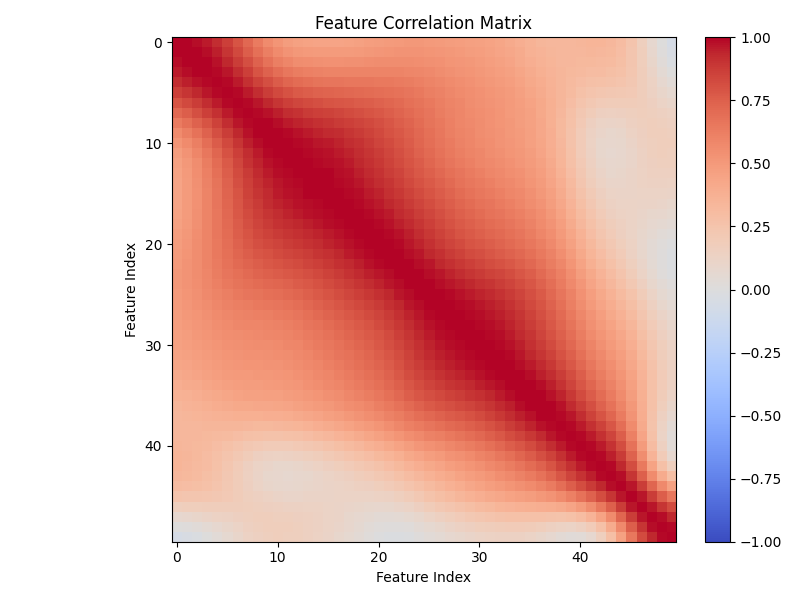}
  \caption{Correlation matrix of the first 50 features after Hilbert transform.}
  \label{fig:corr-matrix}
\end{figure}

\begin{figure}[t]
  \centering
  \includegraphics[width=.6\linewidth]{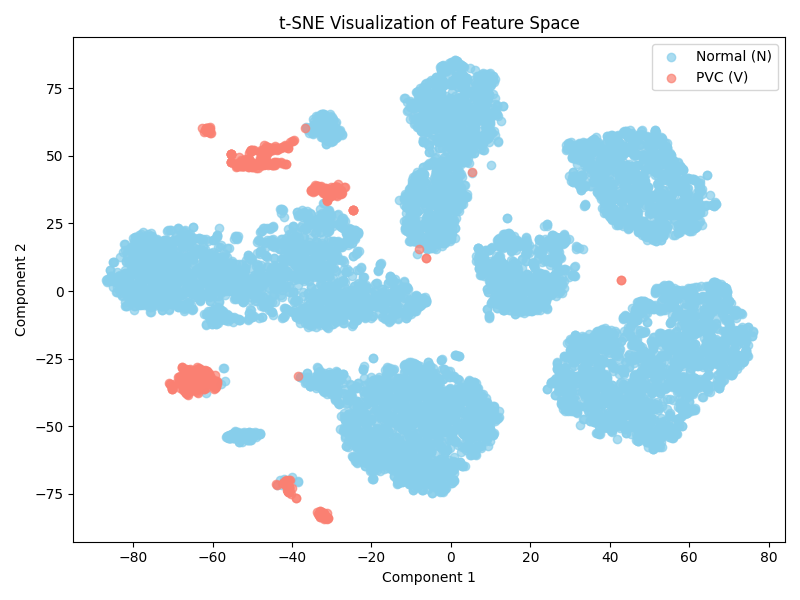}
  \caption{t‑SNE embedding of normal (blue) and PVC (red) beats.}
  \label{fig:tsne}
\end{figure}

\paragraph{Exploratory analysis}
Figure~\ref{fig:class-dist} confirms imbalance
(N\,:\,V \(\approx20{:}1\)).
Waveforms (Fig.\,\ref{fig:segment-N}), Hilbert envelopes
(Figure~\ref{fig:hilbert}) and STFT spectrograms
(Figure~\ref{fig:spectrograms}; Hamming window 32, 50\% overlap)
illustrate morphology. The correlation matrix of the first 50 features
(Figure~\ref{fig:corr-matrix}) shows redundancy; a t-distributed stochastic neighbor embedding (t-SNE) embedding
(Figure~\ref{fig:tsne}) reveals partial cluster separation \citep{van2008visualizing}.

\paragraph{Dataset summary}
After discarding truncated beats, the final corpus contains
\SI{10487}{} normal and \SI{614}{} PVC samples
(Table~\ref{tab:mitbih-stats}); overall imbalance 1:17.1.

\begin{table}[t]
  \centering
  \caption{MIT--BIH statistics after preprocessing.}
  \label{tab:mitbih-stats}
  \begin{tabular}{@{}lcccc@{}}
    \toprule
             & Normal (N) & PVC (V) & Total & Imbalance\\\midrule
    Train    & \num{8389} & \num{491} & \num{8880} & 1 : 17.1\\
    Test     & \num{2098} & \num{123} & \num{2221} & 1 : 17.1\\\midrule
    Overall  & \num{10487} & \num{614} & \num{11101} & 1 : 17.1\\\bottomrule
  \end{tabular}
\end{table}

\medskip

% -----------------------------------------------------------
\subsection{RadioML (I/Q; feature construction)}
\label{app:data-rml}

We utilize the canonical RadioML pickled dictionary \texttt{RML2016.10a\_dict.pkl}, which maps (modulation: str, SNR: int) to arrays with a shape of \((\text{num}, 2, 128)\), where axis~1 stores I and Q components. Each example is flattened to \((256,)\) by concatenating I first, followed by Q, and a compact \(\C^2\) representation is generated through STFT-based summary statistics, referred to as \texttt{stft\_stats} in the code (see \ref{app:impl}).

\paragraph{Loader and filters}
Only selected modulations and a specific SNR range are retained; if no samples correspond to the filters, the loader triggers a \texttt{ValueError}. The dataset is subsequently divided into training, validation, and test partitions with stratified sampling.

\paragraph{Compression to \(\C^2\) via STFT statistics}
For each IQ window, we compute short-time Fourier features and aggregate four complex-valued summary statistics (two complex features) to produce \([\Re_1,\Re_2,\Im_1,\Im_2]\in\R^4\equiv\C^2\). The precise code is available in \texttt{pre\_pro.py} / \texttt{find\_up\_radio.py}, mirroring the \(\C^2\) design utilized in the ECG pipeline.

\paragraph{Uncertainty rule}
We flag anchors using the same union rule as described in the main text: \(\max_k p_k<\tau\) or \(|p_{(1)}-p_{(2)}|<\delta\) (defaults specified in \ref{app:impl}), with the flagged anchors being exported for further local analysis.

% ========================================================
\section{Model Architectures and Training (full)}
\label{app:arch-train}
% ========================================================

\subsection{SimpleComplexNet (synthetic benchmark)}
\label{app:arch-synth-redirect}
Details regarding the architecture and training for the synthetic benchmark can be found in \ref{app:synthetic_sanity}.

\medskip
% -------------------------------------------------------
\subsection{MIT--BIH CVNN pipeline (full)}
\label{app:arch-mit}

We adopt the SimpleComplexNet backbone, broadening it to \(H=64\) and positioning it after the signal\(\to\)\(\C^2\) compression detailed in (\ref{app:data-mitbih}), and preceding calibration (Figure~\ref{fig:tikz-mitbih-net-app}).

\begin{figure}[t]
  \centering
  \begin{tikzpicture}[
    every node/.style={font=\small, align=center},
    io/.style={
      draw,
      trapezium,
      trapezium left angle=70,
      trapezium right angle=110,
      fill=green!8,
      minimum width=2.0cm,
      minimum height=0.8cm
    },
    stage/.style={
      draw,
      rounded rectangle,
      fill=blue!8,
      minimum width=2.0cm,
      minimum height=0.8cm
    },
    arrow/.style={-{Stealth[length=2mm]}, thick},
    node distance=1cm
  ]
    \node[io]   (raw)  {Raw ECG};
    \node[stage, right=of raw]  (bp)   {Band‑pass};
    \node[stage, right=of bp]   (hilb) {Hilbert};
    \node[stage, below=1.5cm of hilb] (feat) {$\Phi:\R^{4L}\to\C^2$};
    \node[stage, below=1.5cm of feat] (W1)  {Comp.\ $W_1$};
    \node[stage, left=1.5cm of W1]          (m1)  {modReLU};
    \node[stage, left=1.5cm of m1]          (W2)  {Comp.\ $W_2$};
    \node[stage, left=1.5cm of W2]          (mag) {$|\cdot|$};
    \node[stage, below=1.5cm of mag] (soft) {Softmax};
    \node[stage, right=of soft]    (temp) {Temp.\ $T$};
    \node[stage, right=of temp]    (unc)  {Uncertain};
   \draw[dashed, thick]
  ($(W1.north west)+(4.5em,0.5em)$)
  rectangle
  ($(mag.south east)+(-4.5em,-0.5em)$);
    \node[above=16pt of W2, anchor=west, font=\small\bfseries]
      {CVNN ($64$ hidden)};
    \draw[arrow] (raw)  -- (bp)   -- (hilb);
    \draw[arrow] (hilb) -- (feat) -- (W1);
    \draw[arrow] (W1)   -- (m1)   -- (W2)   -- (mag);
    \draw[arrow] (mag)  -- (soft) -- (temp) -- (unc);
  \end{tikzpicture}
  \caption{Pipeline for MIT--BIH: signal processing, feature extraction, CVNN, temperature scaling, and uncertainty filter.}
  \label{fig:tikz-mitbih-net-app}
\end{figure}

\paragraph{Training and calibration}
Using the Adam optimizer \((\eta=10^{-3},\beta_1=0.9,\beta_2=0.999,\varepsilon=10^{-8})\), we conduct 20 epochs with a batch size of 128, incorporating early stopping with a patience of 6. The data is split \emph{patient-wise} into folds; the validation fold is used to determine a scalar temperature (using the limited-memory BFGS algorithm, L-BFGS; \(|T|\le 10\) for stability). The uncertainty rule applied is: \(\max_k p_k<\tau\) (\(\tau=0.5\)) or the margin between the top two probabilities \(<\delta\) (\(\delta=0.15\)).

\medskip
% -------------------------------------------------------
\subsection{RadioML pipeline (full)}
\label{app:arch-rml}

We utilize the same SimpleComplexNet with \(H=64\) on the \(\C^2\) features derived from \texttt{stft\_stats} (\ref{app:data-rml}). For calibration baselines, we employ either \emph{Platt} (logistic) or \emph{isotonic} scaling, which are trained on the validation split. By default, we present results based on Platt scaling and integrate our phase-aware temperature \(T’(m)\) with the baseline where it is applicable.

\section{Fragile-anchor triage: PR curves and AUPRC (MIT--BIH)}\label{app:fragileMIT}

\begin{figure*}[t]
  \centering
  \includegraphics[width=0.32\textwidth]{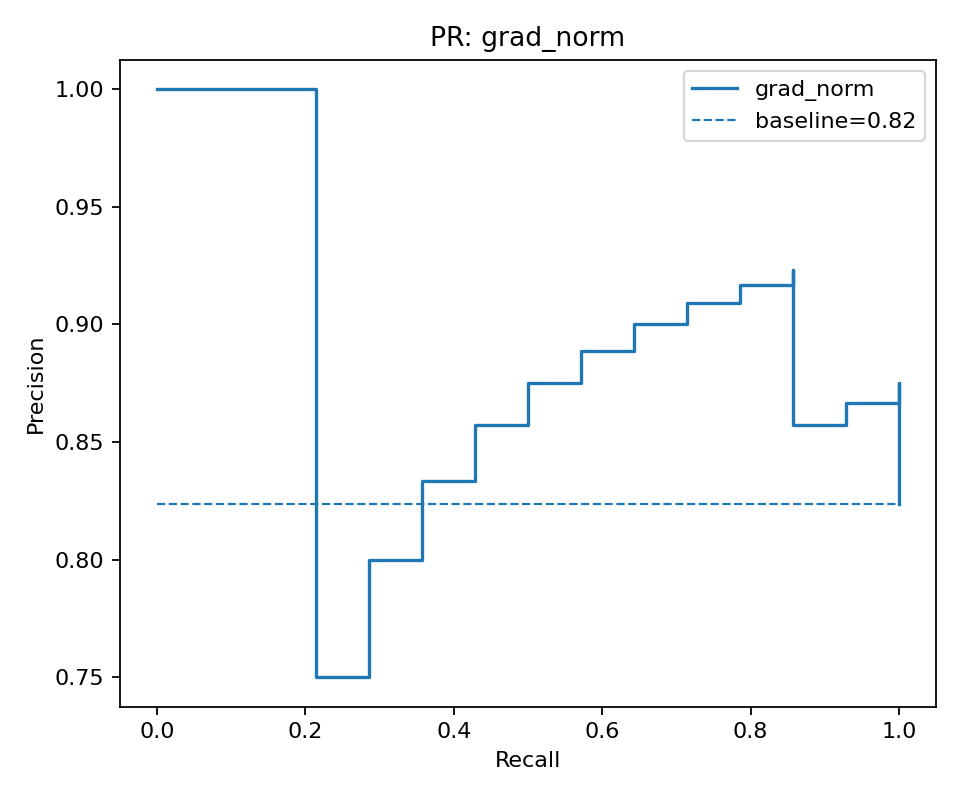}\hfill
  \includegraphics[width=0.32\textwidth]{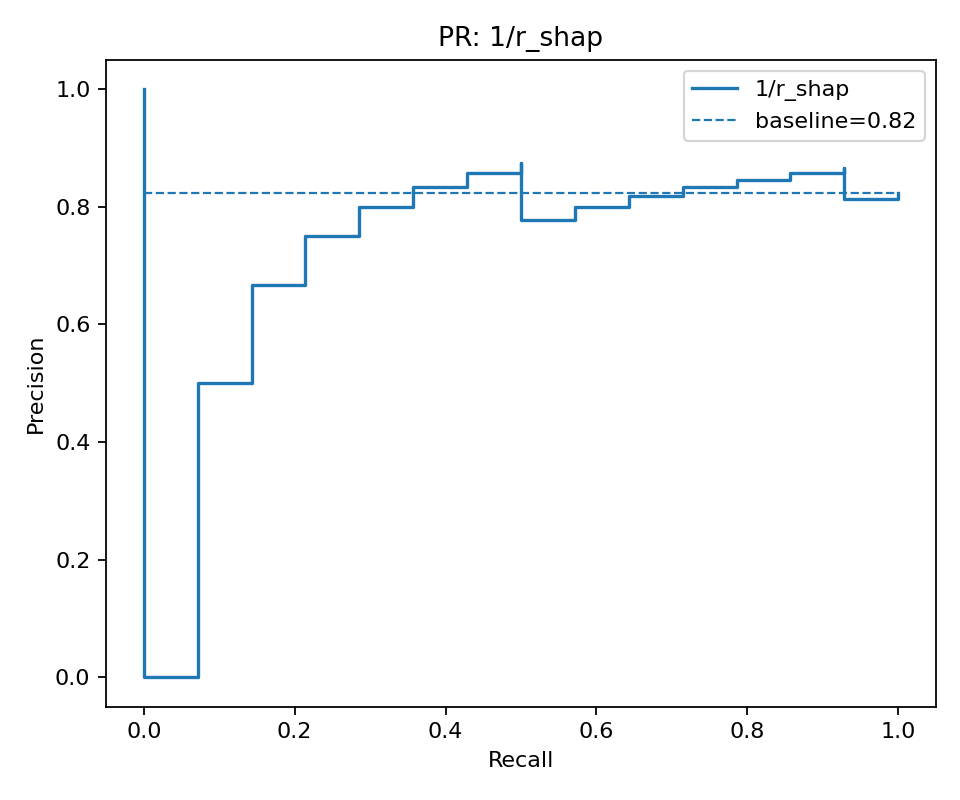}\hfill
  \includegraphics[width=0.32\textwidth]{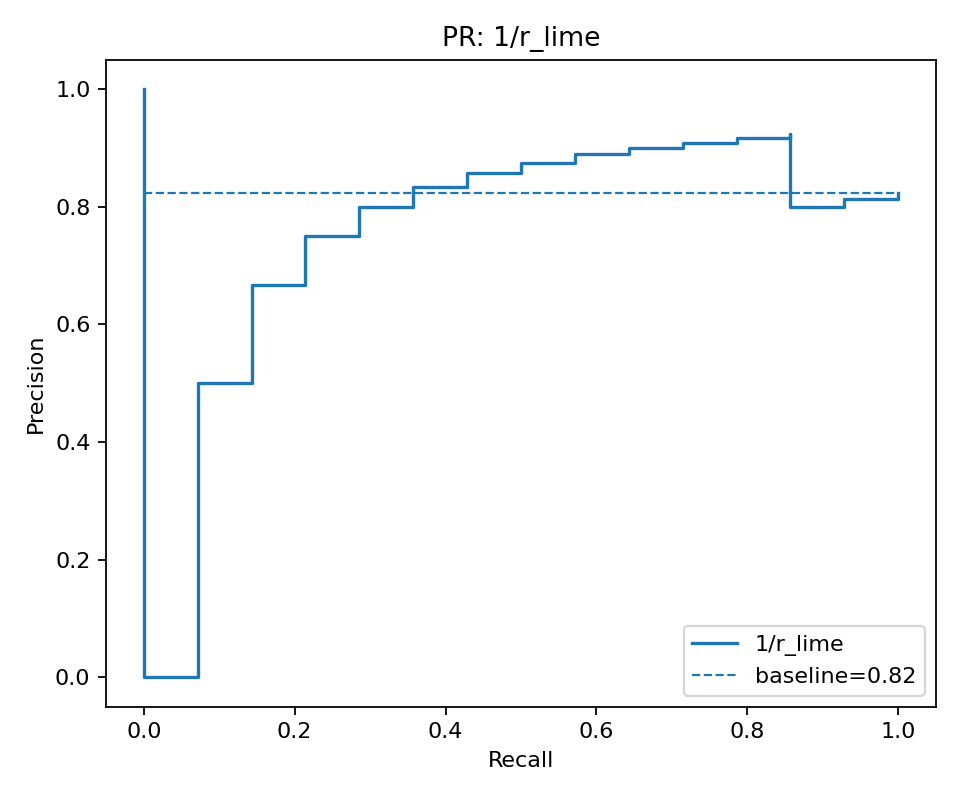}\\[0.5em]
  \includegraphics[width=0.32\textwidth]{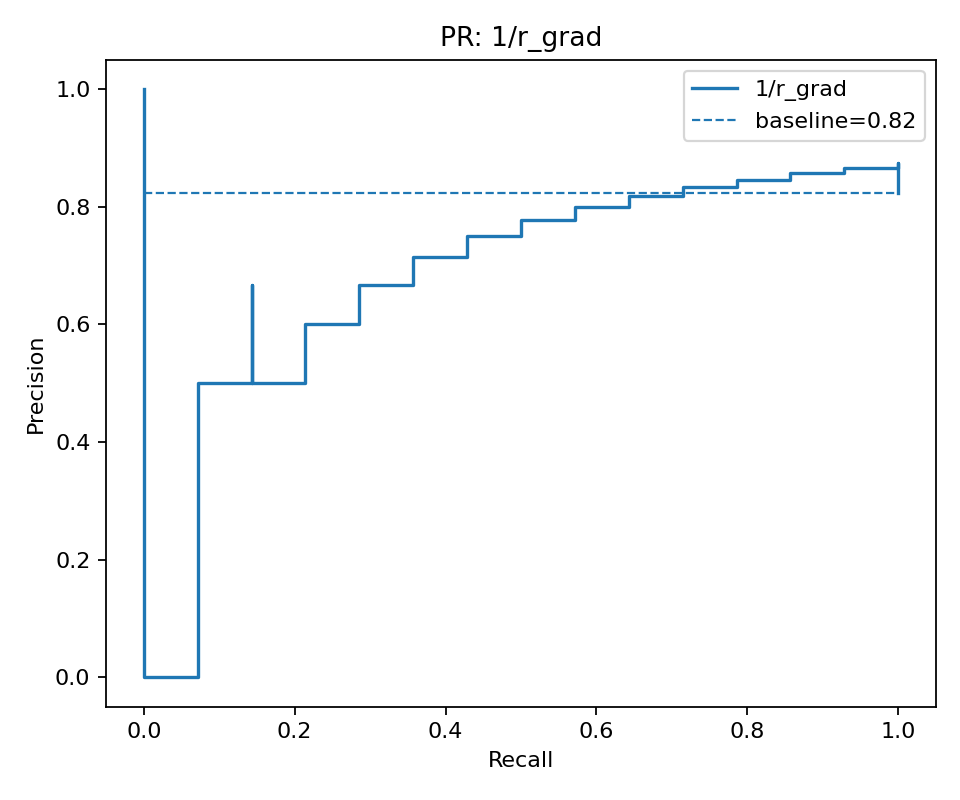}\hfill
  \includegraphics[width=0.32\textwidth]{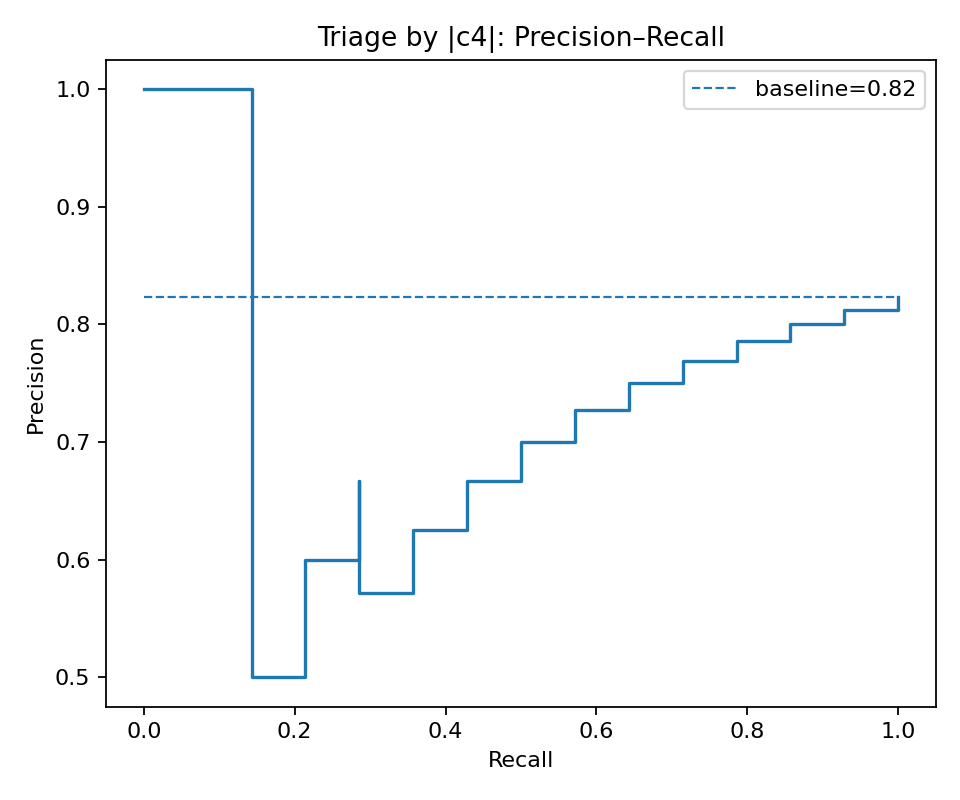}
  \caption{MIT--BIH --- PR curves for triage scores. Dashed line marks the fragile prevalence (0.82).}
  \label{fig:pr_by_gradnorm}
\end{figure*}

\noindent\textbf{Setup.} We treat an anchor as \emph{fragile} if a class flip is observed within the ray-scan budget ($r \le 0.02$). The positive rate on MIT--BIH is $14/17 \approx 0.82$.
We compare five scores for prioritizing anchors: the quartic magnitude $|c_4|$ from the Puiseux surrogate, the inverse flip radii along gradient/LIME/SHAP axes ($1/r_{\text{grad}}$, $1/r_{\text{LIME}}$, $1/r_{\text{SHAP}}$), and the local gradient norm (``grad\_norm''). Average precision (AUPRC) and PR curves are computed exactly as in the released scripts (Figure~\ref{fig:pr_by_gradnorm}).

\section{RadioML fragile-anchor triage: full PR curves}\label{app:fragileRadio}

\begin{figure*}[t]
  \centering
  \includegraphics[width=0.32\textwidth]{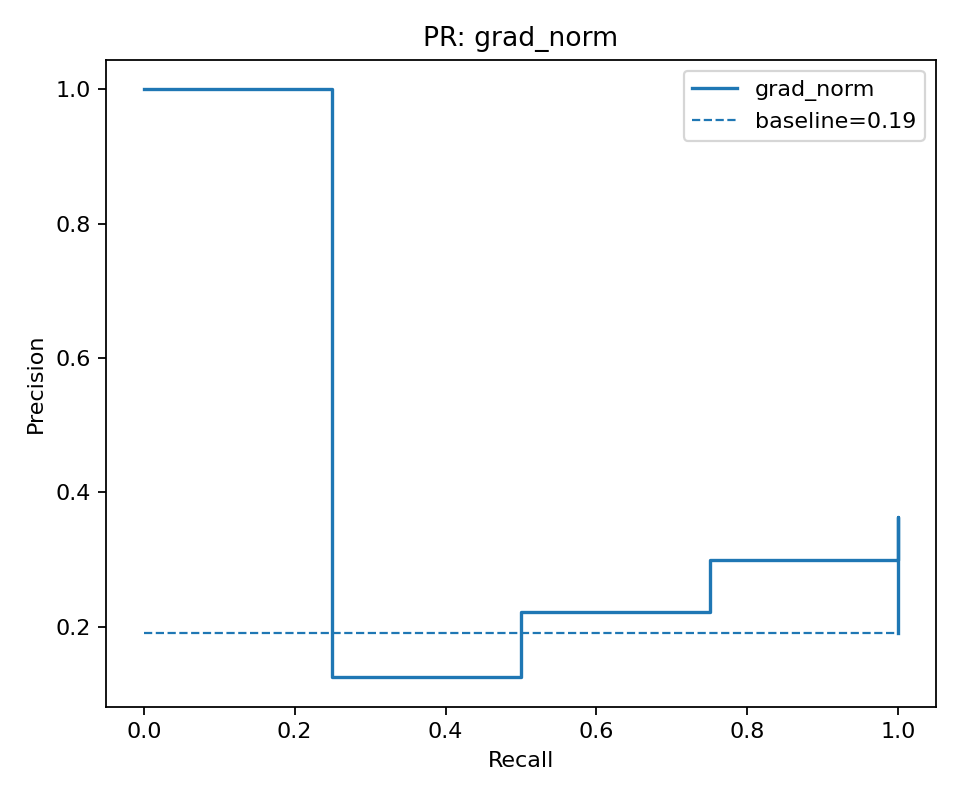}\hfill
  \includegraphics[width=0.32\textwidth]{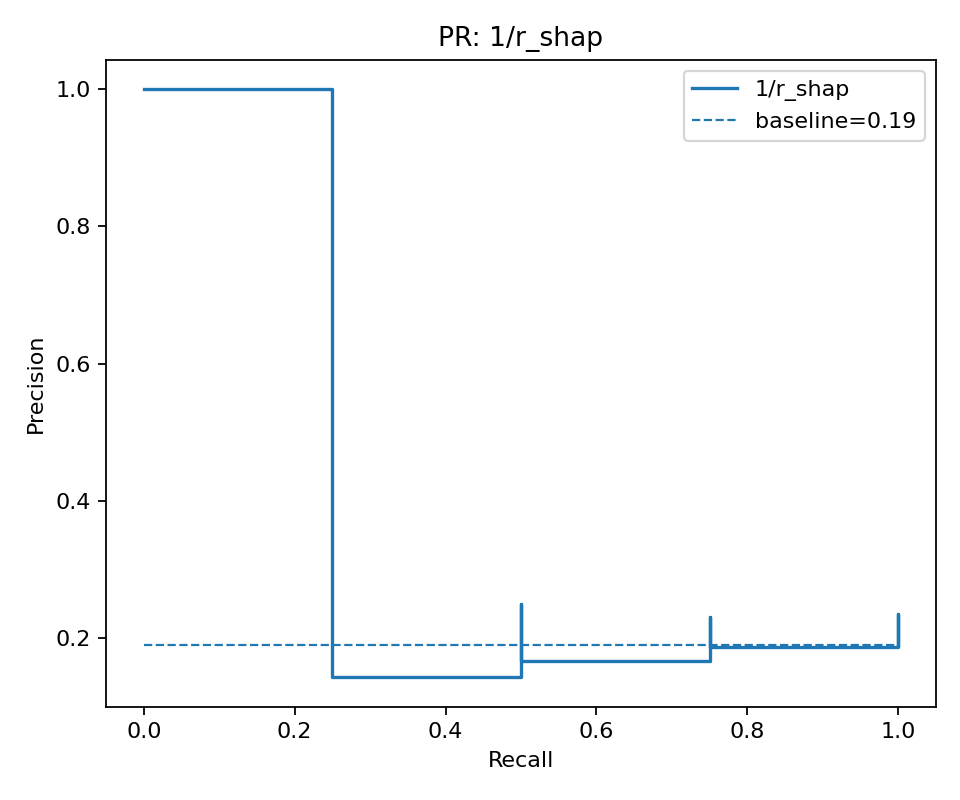}\hfill
  \includegraphics[width=0.32\textwidth]{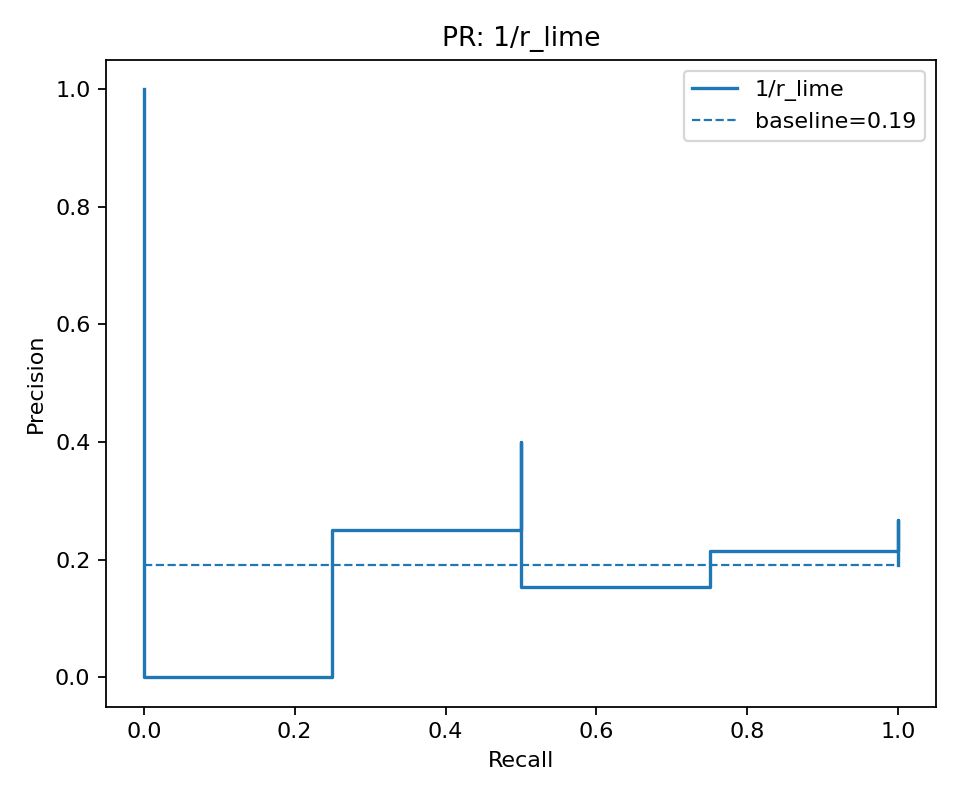}\\[0.5em]
  \includegraphics[width=0.32\textwidth]{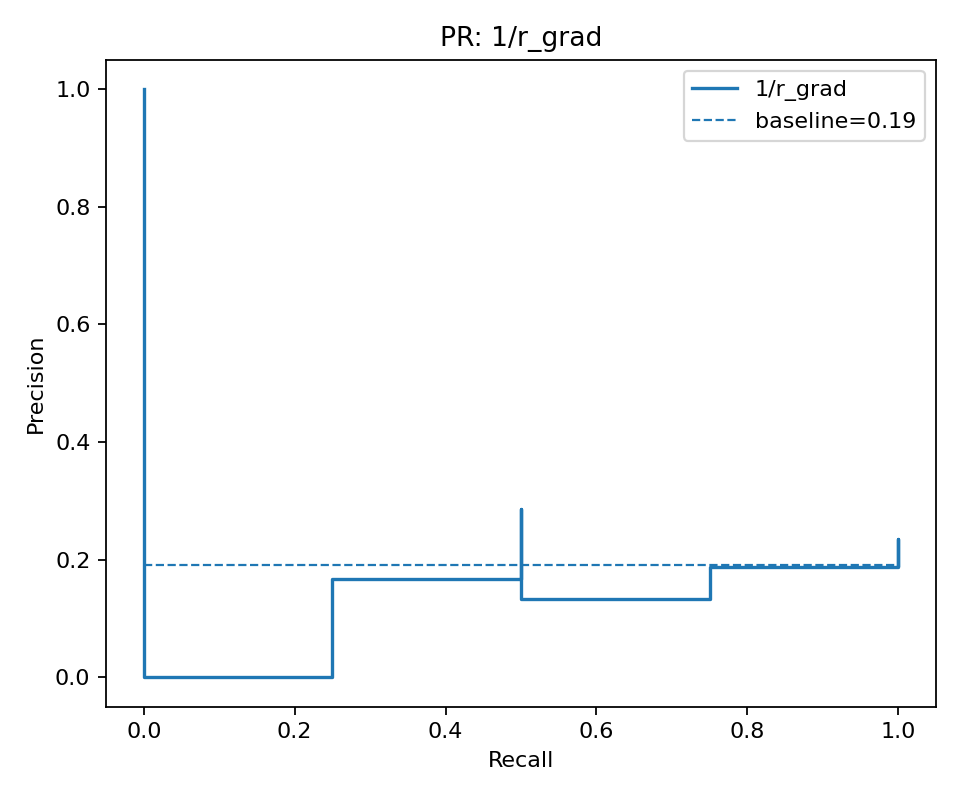}\hfill
  \includegraphics[width=0.32\textwidth]{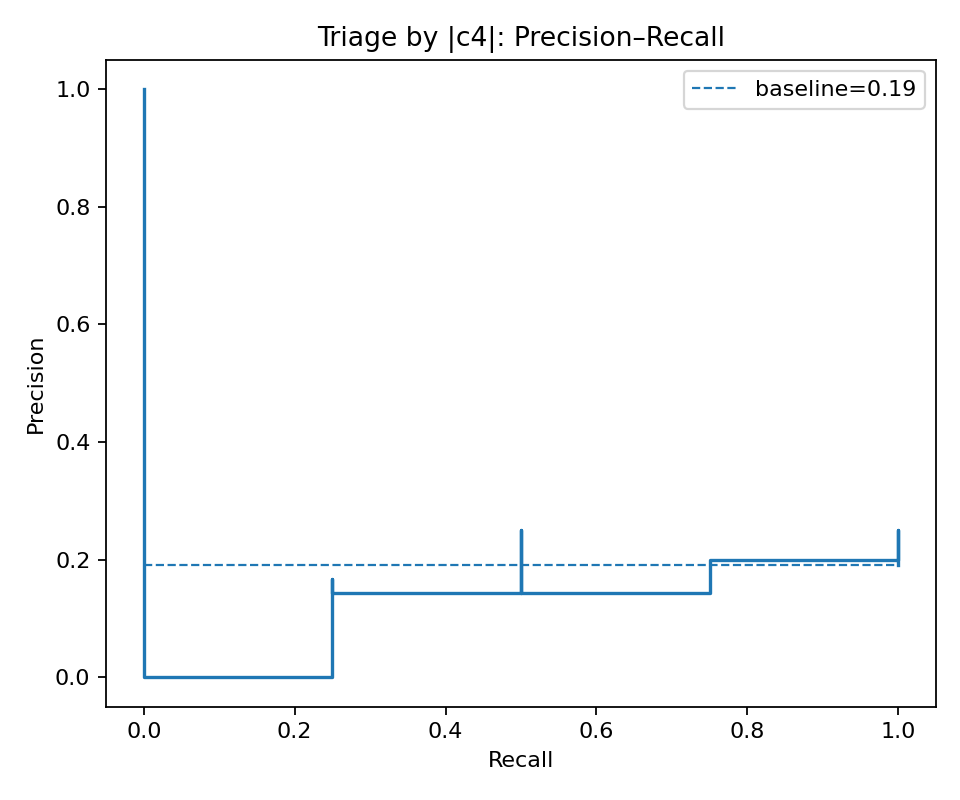}
  \caption{RadioML --- PR curves for triage scores. Dashed line marks the fragile prevalence (0.19).}
  \label{fig:pr_radioml}
\end{figure*}

\noindent
Positives are anchors that flip within the scan budget $r\!\le\!0.02$ (4/21; baseline AUPRC $\approx 0.19$).
Panels show precision--recall curves for the compared scores (Figure~\ref{fig:pr_radioml}). Consistent with Table~\ref{tab:triage_radioml_auprc},
the gradient norm yields the best ranking signal, SHAP-based rays are second, and $|c_4|$ is near the baseline.
On RadioML we thus use Puiseux descriptors for \emph{ranking} and for phase-aligned probing, not for absolute radii.

\FloatBarrier

%========================================================
\bibliographystyle{elsarticle-num}
\bibliography{references} % TODO: ensure updated, consistent formatting

\begin{thebibliography}{10}
\expandafter\ifx\csname url\endcsname\relax
  \def\url#1{\texttt{#1}}\fi
\expandafter\ifx\csname urlprefix\endcsname\relax\def\urlprefix{URL }\fi
\expandafter\ifx\csname href\endcsname\relax
  \def\href#1#2{#2} \def\path#1{#1}\fi

\bibitem{lecun2015deep}
Y.~LeCun, Y.~Bengio, G.~Hinton, Deep learning, nature 521~(7553) (2015)
  436--444.

\bibitem{trabelsi2017deep}
C.~Trabelsi, O.~Bilaniuk, Y.~Zhang, D.~Serdyuk, S.~Subramanian, J.~F. Santos,
  S.~Mehri, N.~Rostamzadeh, Y.~Bengio, C.~J. Pal, Deep complex networks, arXiv
  preprint arXiv:1705.09792 (2017).

\bibitem{barrachina2023theory}
J.~A. Barrachina, C.~Ren, G.~Vieillard, C.~Morisseau, J.-P. Ovarlez, Theory and
  implementation of complex-valued neural networks, arXiv preprint
  arXiv:2302.08286 (2023).

\bibitem{abdalla2023complex}
R.~Abdalla, Complex-valued neural networks--theory and analysis, arXiv preprint
  arXiv:2312.06087 (2023).

\bibitem{basterretxea2021survey}
J.~Bassey, L.~Qian, X.~Li, A survey of complex-valued neural networks, arXiv
  preprint arXiv:2101.12249 (2021).

\bibitem{montavon2018methods}
G.~Montavon, W.~Samek, K.-R. M{\"u}ller, Methods for interpreting and
  understanding deep neural networks, Digital signal processing 73 (2018)
  1--15.

\bibitem{samek2017explainable}
W.~Samek, T.~Wiegand, K.-R. M{\"u}ller, Explainable artificial intelligence:
  Understanding, visualizing and interpreting deep learning models, arXiv
  preprint arXiv:1708.08296 (2017).

\bibitem{naeini2015bayesian}
M.~P. Naeini, G.~Cooper, M.~Hauskrecht, Obtaining well calibrated probabilities
  using bayesian binning, Tech. Rep.~1 (2015).

\bibitem{wall2004singular}
C.~T.~C. Wall, Singular points of plane curves, no.~63, Cambridge University
  Press, 2004.

\bibitem{brieskorn1986plane}
E.~Brieskorn, H.~Kn{\"o}rrer, Plane Algebraic Curves, Birkh{\"a}user, 1986.

\bibitem{walker1950algebraic}
R.~J. Walker, Algebraic curves, Vol.~58, Springer, 1950.

\bibitem{watanabe2009book}
S.~Watanabe, Algebraic geometry and statistical learning theory, Vol.~25,
  Cambridge university press, 2009.

\bibitem{Maragos_2021}
P.~Maragos, V.~Charisopoulos, E.~Theodosis, Tropical geometry and machine
  learning, Proceedings of the IEEE 109~(5) (2021) 728–755.

\bibitem{suzuki2023waic}
J.~Suzuki, Waic and wbic with r stan: 100 exercises for building logic,
  Springer Nature, 2023.

\bibitem{o2016convolutional}
T.~J. O’Shea, J.~Corgan, T.~C. Clancy, Convolutional radio modulation
  recognition networks, in: International conference on engineering
  applications of neural networks, Springer, 2016, pp. 213--226.

\bibitem{hirose2006complex}
A.~Hirose, et~al., Complex-valued neural networks, Vol.~32, Wiley Online
  Library, 2006.

\bibitem{nitta2004orthogonality}
T.~Nitta, Orthogonality of decision boundaries in complex-valued neural
  networks, Neural computation 16~(1) (2004) 73--97.

\bibitem{kreutz2009complex}
K.~Kreutz-Delgado, The complex gradient operator and the cr-calculus, arXiv
  preprint arXiv:0906.4835 (2009).

\bibitem{guberman2016complex}
N.~Guberman, On complex valued convolutional neural networks, arXiv preprint
  arXiv:1602.09046 (2016).

\bibitem{arjovsky2016unitary}
M.~Arjovsky, A.~Shah, Y.~Bengio, Unitary evolution recurrent neural networks,
  in: International conference on machine learning, PMLR, 2016, pp. 1120--1128.

\bibitem{cvkan2025}
M.~Wolff, F.~Eilers, X.~Jiang, Cvkan: Complex-valued kolmogorov-arnold
  networks, arXiv preprint arXiv:2502.02417 (2025).

\bibitem{liu2024kan}
Z.~Liu, Y.~Wang, S.~Vaidya, F.~Ruehle, J.~Halverson, M.~Solja{\v{c}}i{\'c},
  T.~Y. Hou, M.~Tegmark, Kan: Kolmogorov-arnold networks, arXiv preprint
  arXiv:2404.19756 (2024).

\bibitem{dudek2025hkan}
G.~Dudek, T.~Rodak, Hkan: Hierarchical kolmogorov-arnold network without
  backpropagation, arXiv preprint arXiv:2501.18199 (2025).

\bibitem{cvnn_federated_2024}
H.~Yu, Y.~Liu, M.~Chen, Complex-valued neural-network-based federated learning
  for multiuser indoor positioning performance optimization, IEEE Internet of
  Things Journal 11~(21) (2024) 34065--34077.

\bibitem{cvyolo_2025}
D.~Zhao, Z.~Zhang, D.~Lu, X.~Qiu, W.~Li, H.~Li, Y.~Wu, Cv-yolo: A
  complex-valued convolutional neural network for oriented ship detection in
  single-polarization single-look complex sar images, Remote Sensing 17~(8)
  (2025) 1478.

\bibitem{zhang2018complex}
J.~Zhang, Y.~Wu, Complex-valued unsupervised convolutional neural networks for
  sleep stage classification, Computer methods and programs in biomedicine 164
  (2018) 181--191.

\bibitem{xu2023ultrasonic}
Y.~Xu, F.~Pourahmadian, J.~Song, C.~Wang, Deep learning for full-field
  ultrasonic characterization, Mechanical Systems and Signal Processing 201
  (2023) 110668.

\bibitem{Li_2025}
S.~Li, L.~Zhang, H.~Guo, J.~Li, J.~Yu, X.~He, Y.~Zhao, X.~He, Csa-fcn: Channel-
  and spatial-gated attention mechanism based fully complex-valued neural
  network for system matrix calibration in magnetic particle imaging, IEEE
  Transactions on Computational Imaging 11 (2025) 65–76.

\bibitem{ribeiro2016should}
M.~T. Ribeiro, S.~Singh, C.~Guestrin, " why should i trust you?" explaining the
  predictions of any classifier, in: Proceedings of the 22nd ACM SIGKDD
  international conference on knowledge discovery and data mining, 2016, pp.
  1135--1144.

\bibitem{lundberg2017unified}
S.~M. Lundberg, S.-I. Lee, A unified approach to interpreting model
  predictions, Advances in neural information processing systems 30 (2017).

\bibitem{jiang2019image}
L.~Jiang, Z.~Wang, M.~Xu, Z.~Wang, Image saliency prediction in transformed
  domain: A deep complex neural network method, in: Proceedings of the AAAI
  Conference on Artificial Intelligence, Vol.~33, 2019, pp. 8521--8528.

\bibitem{arrieta2020explainable}
A.~Barredo~Arrieta, N.~Díaz-Rodríguez, J.~Del~Ser, A.~Bennetot, S.~Tabik,
  A.~Barbado, S.~Garcia, S.~Gil-Lopez, D.~Molina, R.~Benjamins, R.~Chatila,
  F.~Herrera, Explainable artificial intelligence (xai): Concepts, taxonomies,
  opportunities and challenges toward responsible ai, Information Fusion 58
  (2020) 82–115.

\bibitem{guo2017calibration}
C.~Guo, G.~Pleiss, Y.~Sun, K.~Q. Weinberger, On calibration of modern neural
  networks, in: International conference on machine learning, PMLR, 2017, pp.
  1321--1330.

\bibitem{deepcshap_2024}
F.~Eilers, X.~Jiang, Deepcshap: Utilizing shapley values to explain deep
  complex-valued neural networks, arXiv preprint arXiv:2403.08428 (2024).

\bibitem{choi2018phase}
H.-S. Choi, J.-H. Kim, J.~Huh, A.~Kim, J.-W. Ha, K.~Lee, Phase-aware speech
  enhancement with deep complex u-net, in: International Conference on Learning
  Representations, 2018.

\bibitem{calibration_survey_2023}
C.~Wang, Calibration in deep learning: A survey of the state-of-the-art, arXiv
  preprint arXiv:2308.01222 (2023).

\bibitem{benchmark_calib_2023}
L.~Tao, Y.~Zhu, H.~Guo, M.~Dong, C.~Xu, A benchmark study on calibration, arXiv
  preprint arXiv:2308.11838 (2023).

\bibitem{manyclass_calibration_2024}
A.~Le~Coz, S.~Herbin, F.~Adjed, Confidence calibration of classifiers with many
  classes, Advances in Neural Information Processing Systems 37 (2024)
  77686--77725.

\bibitem{kull2019dirichlet}
M.~Kull, M.~Perello~Nieto, M.~K{\"a}ngsepp, T.~Silva~Filho, H.~Song, P.~Flach,
  Beyond temperature scaling: Obtaining well-calibrated multi-class
  probabilities with dirichlet calibration, Vol.~32, 2019.

\bibitem{minderer2021revisiting}
M.~Minderer, J.~Djolonga, R.~Romijnders, F.~Hubis, X.~Zhai, N.~Houlsby,
  D.~Tran, M.~Lucic, Revisiting the calibration of modern neural networks,
  Advances in neural information processing systems 34 (2021) 15682--15694.

\bibitem{lakshminarayanan2017simple}
B.~Lakshminarayanan, A.~Pritzel, C.~Blundell, Simple and scalable predictive
  uncertainty estimation using deep ensembles, Advances in neural information
  processing systems 30 (2017).

\bibitem{messoudi2022conformal}
S.~Messoudi, Conformal prediction methods for complex data: application to real
  estate management, Ph.D. thesis, Universit{\'e} de Technologie de
  Compi{\`e}gne (2022).

\bibitem{maclagan2015tropical}
D.~Maclagan, B.~Sturmfels, Introduction to tropical geometry, Vol. 161,
  American Mathematical Soc., 2015.

\bibitem{montufar2014number}
G.~F. Montufar, R.~Pascanu, K.~Cho, Y.~Bengio, On the number of linear regions
  of deep neural networks, Advances in neural information processing systems 27
  (2014).

\bibitem{tropical_tutorial_icassp_2024}
D.~Barnhill, R.~Yoshida, G.~Aliatimis, K.~Miura, Tropical geometric tools for
  machine learning: The tml package, Journal of Software for Algebra and
  Geometry 14~(1) (2024) 133--174.

\bibitem{douglas2021mumford}
M.~R. Douglas, From algebraic geometry to machine learning, Pure and Applied
  Mathematics Quarterly 17~(2) (2021) 605–617.

\bibitem{bao2022mlag}
J.~Bao, Y.-H. He, E.~Heyes, E.~Hirst, Machine learning algebraic geometry for
  physics, arXiv preprint arXiv:2204.10334 (2022).

\bibitem{coates2023terminal}
T.~Coates, A.~Kasprzyk, S.~Veneziale, Machine learning detects terminal
  singularities, Vol.~36, 2023, pp. 67183--67194.

\bibitem{slt_upper_bound_2024}
N.~Hayashi, Y.~Sawada, Upper bound of bayesian generalization error in partial
  concept bottleneck model (cbm): Partial cbm outperforms naive cbm, arXiv
  preprint arXiv:2403.09206 (2024).

\bibitem{degeneracy_mechinterp_iclr_2024}
L.~Bushnaq, J.~Mendel, S.~Heimersheim, D.~Braun, N.~Goldowsky-Dill,
  K.~H{\"a}nni, C.~Wu, M.~Hobbhahn, Using degeneracy in the loss landscape for
  mechanistic interpretability, arXiv preprint arXiv:2405.10927 (2024).

\bibitem{bishop2006pattern}
C.~M. Bishop, N.~M. Nasrabadi, Pattern recognition and machine learning,
  Vol.~4, Springer, 2006.

\bibitem{brandwood1983complex}
D.~H. Brandwood, A complex gradient operator and its application in adaptive
  array theory, in: IEE Proceedings F (Communications, Radar and Signal
  Processing), Vol. 130, IET, 1983, pp. 11--16.

\bibitem{golub2013matrix}
G.~H. Golub, C.~F. Van~Loan, Matrix computations, JHU press, 2013.

\bibitem{hansen1998rank}
P.~C. Hansen, Rank-deficient and discrete ill-posed problems: numerical aspects
  of linear inversion, SIAM, 1998.

\bibitem{demmel1997applied}
J.~W. Demmel, Applied numerical linear algebra, SIAM, 1997.

\bibitem{saritha2008ecg}
C.~Saritha, V.~Sukanya, Y.~N. Murthy, Ecg signal analysis using wavelet
  transforms, Bulg. J. Phys 35~(1) (2008) 68--77.

\bibitem{lu2018feature}
W.~Lu, H.~Hou, J.~Chu, Feature fusion for imbalanced ecg data analysis,
  Biomedical Signal Processing and Control 41 (2018) 152--160.

\bibitem{acharya2017automated}
U.~R. Acharya, H.~Fujita, O.~S. Lih, Y.~Hagiwara, J.~H. Tan, M.~Adam, Automated
  detection of arrhythmias using different intervals of tachycardia ecg
  segments with convolutional neural network, Information sciences 405 (2017)
  81--90.

\bibitem{fawcett2006introduction}
T.~Fawcett, An introduction to roc analysis, Pattern recognition letters 27~(8)
  (2006) 861--874.

\bibitem{pedregosa2011scikit}
F.~Pedregosa, G.~Varoquaux, A.~Gramfort, V.~Michel, B.~Thirion, O.~Grisel,
  M.~Blondel, P.~Prettenhofer, R.~Weiss, V.~Dubourg, et~al., Scikit-learn:
  Machine learning in python, the Journal of machine Learning research 12
  (2011) 2825--2830.

\bibitem{jolliffe2002principal}
I.~T. Jolliffe, Principal component analysis for special types of data,
  Springer, 2002.

\bibitem{mckay2000comparison}
M.~D. McKay, R.~J. Beckman, W.~J. Conover, A comparison of three methods for
  selecting values of input variables in the analysis of output from a computer
  code, Technometrics 42~(1) (2000) 55--61.

\bibitem{sobol1967distribution}
I.~M. Sobol, The distribution of points in a cube and the approximate
  evaluation of integrals, USSR Computational mathematics and mathematical
  physics 7 (1967) 86--112.

\bibitem{owen1997quasi}
A.~B. Owen, Monte carlo variance of scrambled net quadrature, SIAM Journal on
  Numerical Analysis 34~(5) (1997) 1884--1910.

\bibitem{dean1999design}
A.~Dean, D.~Voss, Design and analysis of experiments, Springer, 1999.

\bibitem{golub1979gcv}
G.~H. Golub, M.~Heath, G.~Wahba, Generalized cross-validation as a method for
  choosing a good ridge parameter, Technometrics 21~(2) (1979) 215--223.

\bibitem{wahba1990spline}
G.~Wahba, Spline models for observational data, SIAM, 1990.

\bibitem{benesty2009pearson}
I.~Cohen, Y.~Huang, J.~Chen, J.~Benesty, J.~Benesty, J.~Chen, Y.~Huang,
  I.~Cohen, Pearson correlation coefficient, Noise reduction in speech
  processing (2009) 1--4.

\bibitem{willmott2005advantages}
C.~J. Willmott, K.~Matsuura, Advantages of the mean absolute error (mae) over
  the root mean square error (rmse) in assessing average model performance,
  Climate research 30~(1) (2005) 79--82.

\bibitem{zassenhaus1969factor}
H.~Zassenhaus, On hensel factorization i, J. Number Theory 1~(1) (1969)
  291--311.

\bibitem{moody2001mitbih}
G.~B. Moody, R.~G. Mark, The impact of the mit-bih arrhythmia database, IEEE
  engineering in medicine and biology magazine 20~(3) (2001) 45--50.

\bibitem{goldberger2000physiobank}
A.~L. Goldberger, L.~A.~N. Amaral, L.~Glass, J.~M. Hausdorff, P.~C. Ivanov,
  R.~G. Mark, J.~E. Mietus, G.~B. Moody, C.-K. Peng, H.~E. Stanley, Physiobank,
  physiotoolkit, and physionet: Components of a new research resource for
  complex physiologic signals, Circulation 101~(23) (2000).

\bibitem{dibin2022hilbert}
N.~B. Nizam, S.~I. S.~K. Nuhash, T.~Hasan, Hilbert-envelope features for
  cardiac disease classification from noisy phonocardiograms, Biomedical Signal
  Processing and Control 78 (2022) 103864.

\bibitem{boashash1992estimating}
B.~Boashash, Estimating and interpreting the instantaneous frequency of a
  signal-part2: Algorithms and applications proceedings of the ieee, VOL80 4
  (1992).

\bibitem{benitez2001use}
D.~Benitez, P.~Gaydecki, A.~Zaidi, A.~Fitzpatrick, The use of the hilbert
  transform in ecg signal analysis, Computers in biology and medicine 31~(5)
  (2001) 399--406.

\bibitem{van2008visualizing}
L.~Van~der Maaten, G.~Hinton, Visualizing data using t-sne., Journal of machine
  learning research 9~(11) (2008).

\bibitem{smith2003digital}
S.~Smith, Digital signal processing: a practical guide for engineers and
  scientists, Newnes, 2003.

\end{thebibliography}

\end{document}